\pdfoutput=1
\documentclass[journal,12pt,onecolumn,draftclsnofoot]{IEEEtran}
\usepackage[margin=1in]{geometry} 
\usepackage{acronym}
\usepackage{amssymb}
\usepackage[table,xcdraw]{xcolor}
\usepackage{textcomp}
\usepackage{graphics}
\usepackage{epstopdf}
\usepackage{float}
\usepackage{color}
\usepackage[cmex10]{amsmath}
\usepackage{latexsym,amsfonts}
\usepackage{url}
\usepackage{longtable}
\usepackage[figuresright]{rotating}
\usepackage{listings}
\usepackage{etoolbox}
\usepackage{bm}
\usepackage{bbm}
\usepackage{euscript}
\usepackage{mathtools}
\usepackage{subcaption}
\usepackage{enumerate}
\usepackage{tikz}
\usepackage{relsize}
\usepackage{easyReview}
\usepackage{xcolor}
\usepackage{optidef}
\mathtoolsset{showonlyrefs,showmanualtags}

\begin{document}

\title{Weight Vector Tuning and Asymptotic Analysis of Binary Linear Classifiers}

\author{Lama~B.~Niyazi,
        Abla~Kammoun,~\IEEEmembership{Member,~IEEE},
        Hayssam~Dahrouj,~\IEEEmembership{Senior Member,~IEEE},
        Mohamed-Slim~Alouini,~\IEEEmembership{Fellow,~IEEE},
        and Tareq~Y.~Al-Naffouri,~\IEEEmembership{Senior Member,~IEEE}
 \thanks{L. B. Niyazi, A. Kammoun, M.-S. Alouni, and T. Y. Al-Naffouri  are with the Electrical and Computer Engineering Program, King Abdullah University of Science and Technology, Thuwal, Saudi Arabia; emails: \{lama.niyazi, abla.kammoun, slim.alouini, tareq.alnaffouri\}@kaust.edu.sa}
\thanks{H. Dahrouj is with the Center of Excellence for NEOM Research at KAUST, King Abdullah University of Science and Technology, Thuwal, Saudi Arabia; email: hayssam.dahrouj@gmail.com} }

\maketitle

\begin{abstract}
% A good discriminant function should be able to effectively extract discriminating information from the data point to which it is applied. Nevertheless, even good discriminants can be affected by non-ideal settings in which their discriminating capability is significantly reduced. %For example, under high estimation noise, the Linear Discriminant Analysis (LDA) classifier, although being the Bayes classifier under certain data assumptions, performs very poorly under those same assumptions. 
Unlike its intercept, a linear classifier's weight vector cannot be tuned by a simple grid search. Hence, this paper proposes weight vector tuning of a generic binary linear classifier through the parameterization of a decomposition of the discriminant by a scalar which controls the trade-off between conflicting informative and noisy terms. By varying this parameter, the original weight vector is modified in a meaningful way. Applying this method to a number of linear classifiers under a variety of data dimensionality and sample size settings reveals that the classification performance loss due to non-optimal native hyperparameters can be compensated for by weight vector tuning. This yields computational savings as the proposed tuning method reduces to tuning a scalar compared to tuning the native hyperparameter, which may involve repeated weight vector generation along with its burden of optimization, dimensionality reduction, etc., depending on the classifier. It is also found that weight vector tuning significantly improves the performance of Linear Discriminant Analysis (LDA) under high estimation noise. Proceeding from this second finding, an asymptotic study of the misclassification probability of the parameterized LDA classifier in the growth regime where the data dimensionality and sample size are comparable is conducted. Using random matrix theory, the misclassification probability is shown to converge to a quantity that is a function of the true statistics of the data. Additionally, an estimator of the misclassification probability is derived. Finally, computationally efficient tuning of the parameter using this estimator is demonstrated on real data. 
\end{abstract}

\section{Introduction}
% Linear classifiers, ubiquity, usefulness, advantages etc
% The many philosophies of linear classification
% Improvements to linear classifiers in the literature
% What we did applies to any and how it works and blah blah
% Talk about paper structure
% THE END
% GOD HELP ME
 A binary linear classifier classifies a data point to one class or the other by thresholding a discriminant that is a linear combination of the data features. The weights of the features make up a \textit{weight vector} and the constant term in the discriminant is the \textit{bias} of the classifier. 
 
Despite the availability of sophisticated non-linear methods for classification, linear classifiers are still widely used. In fact, new variants of standard linear methods catering to specific settings and applications are being developed all the time. A search of the recent literature reveals that linear classifiers are being employed in many tasks including clinical neuroimaging \cite{MARQUAND202083}, digital pulse shape discrimination \cite{wen2020performance}, predicting the genetic merit of beef cattle \cite{berry2019linear}, and in conjunction with other methods for applications such as pathogen identification \cite{randhawa2020machine}, strategy representation \cite{ashok2019strategy}, and cancer classification \cite{alanni2019novel}. Linear classifiers are especially suited to certain high-dimensional datasets on which they perform comparably with non-linear classifiers, with the advantage of much faster training times and quicker classification \cite{yuan2012recent}. Due to ease of computation, linear classifiers further make good trial classifiers during the initial exploratory phase, when the relationship between the data features and labels is yet unknown \cite{duda2001pattern}. 

% \alert{need to talk here about different ways of improving linear classifiers weight vector intercept feature manipulation and ours is DIFFERENT TEST POINT INFORMATION but also ours is different in the sense that it is not part of the architecture as dr abla says should check her email}

% \alert{-one way to improve is intercept
% tuning but this is a scalar? 
% -also expression for optimal in LDA case and cite references below
% -How can you improve upon the weight vector which can't be tuned in the same way
% -tuning of weight vector reduces to tuning of a scalar in a meaningful way}

One way of improving a given linear classifier's performance on a particular dataset is by tuning its bias so as to minimize training error on that dataset \cite{friedman2001elements}. Because the bias is a scalar, a grid search for the optimum is computationally undemanding. Even the need for a grid-search can be eliminated in many cases for which explicit representations of the optimal bias can be derived. For example, the authors of \cite{wang2018dimension} derive an explicit bias correction of the Linear Discriminant Analysis (LDA) classifier discriminant in order to improve classification in the high estimation noise regime. The authors of \cite{zollanvari2019asymptotically} similarly correct for the bias of this classifier in an explicit form, but in the context of cost-sensitive classification. Additionally, the references \cite{huang2010bias} and \cite{sifaou2020high} provide explicit bias corrections for certain high-dimensional variants of LDA. A related question has to do with improving upon a linear classifier's weight vector, which cannot be tuned or corrected in the same way. Relying on the intuition that a good weight vector should be able to extract the maximum discriminatory information content from the data point being classified, we show in this work that tuning the multidimensional weight vector can indeed be reduced to tuning a scalar. 

In the first half of this paper, it is shown that any binary linear classifier discriminant can be decomposed into terms containing discriminating information and non-discriminating noise. A linear form of this decomposition parameterized by a variable $\alpha$ controls the trade-off between conflicting noise and information terms. At the optimal setting of $\alpha$, the modified discriminant performs at least as good as the original classifier from which it was produced. Following this, the effect of the weight vector modification on the performance of an assortment of linear classifiers under different data dimensionality and sample size  scenarios is studied. The method specifically yields significant performance gains for the Linear Discriminant Analysis (LDA) classifier under high estimation noise. Interestingly, the parameterized LDA operates as a bridge between LDA and the nearest centroid classifier, and performs at least as good as either of these classifiers. Additionally, it is shown that tuning the weight vector according to the proposed method can significantly improve the performance of certain classifiers whose native hyperparameters are not optimally set. It is shown that with weight vector tuning, the Support Vector Machine (SVM) with non-optimally tuned penalty can achieve performance close to that of its tuned counterpart.  In this case, tuning the weight vector is fundamentally different from tuning the native hyperparameter of the classifier as it occurs post weight vector generation, while the native hyperparameter tuning occurs prior to weight vector generation. For SVM, generating the weight vector for each value of the native hyperparameter involves solving an optimization problem. Tuning the weight vector according to the proposed method, however, reduces to a simple grid search over a scalar parameter. This idea can be generalized to any classifier with hyperparameters that are set prior to weight vector generation.    %LDA is especially sensitive to estimation noise as it is an explicit function of sample estimates.

% To the best of our knowledge, this is the only work that modifies the LDA weight vector in order to improve classification performance in the high estimation noise regime.

% \alert{also mention what we show about improving untuned architectures and this is fundamentally different than tuning the native hyperparameters because LESS WORK give examples RP-LDA and SVM}

The second half of the paper consists of an asymptotic study of the parameterized LDA classifier under a growth regime in which the data dimensionality and sample size grow proportionally.
% This is commonly referred to in the classification context as the `small sample regime'.
We use random matrix theory to show that the probability of misclassification of this classifier converges to a limit that is a function of the true class statistics. We also derive a consistent estimator of the probability of misclassification by which the classifier parameter $\alpha$ can be tuned. This estimator is more computationally efficient than other tuning methods which rely on additional testing points  or recycling the training set, e.g. cross-validation, as it requires no additional testing points and no averaging. We demonstrate its performance on real data. 

An additional finding of this work is a new interpretation of the optimality of LDA. The LDA decision rule, derived by maximizing the posterior probability of a test point, assuming that it is drawn from a Gaussian distribution with classes having distinct means and common covariances, yields a weight vector which is the optimal Bayes direction. It can be shown that, under a common class covariance, the weight vector resulting from Fisher's linear discriminant, in which the ratio of the distance between the projected class means and the within class variance is maximized, is proportional to the Bayes direction \cite{friedman2001elements}. A proportional solution can also be arrived at via a least squares formulation of the fitted data from their labels \cite{friedman2001elements} in the binary case \cite{friedman2001elements}. This makes the Bayes direction optimal in the posterior probability sense, the Fisher's linear discriminant sense, and the least squares sense. Moreover, this paper shows that the Bayes direction is optimal in the sense that it achieves the minimum noise (in the mean square error sense) with respect to the test point when the classes are Gaussian with common covariance. 

To summarize, the main contributions of this paper are
\begin{itemize}
    \item A practical method for weight vector tuning which reduces to grid search over a scalar parameter
    \item A novel interpretation of the optimality of the LDA classifier in terms of minimizing test point noise
    \item Asymptotic expressions for the probability of misclassification of the parameterized LDA classifier 
    \item A consistent estimator of the probability of misclassification of the parameterized LDA classifier
\end{itemize}

Throughout the paper, scalars are denoted by plain lower-case letters, vectors by bold lower-case letters, and matrices by bold upper-case letters. The symbol $\textbf{I}_p$ is used to represent the $p\times p$ identity matrix, the symbol $\textbf{1}_p$ represents the all-ones $p\times 1$ vector, and the symbol $\textbf{0}_p$ represents the all-zeros $p\times 1$ vector. The notation $||\cdot||$ is used to symbolize the Euclidean norm when its argument is a vector and the spectral norm when its argument is a matrix. The operator $\lceil\cdot\rceil$ rounds its argument up to the nearest integer. Almost-sure convergence is denoted by $\xrightarrow{\text{a.s.}}$ or $a\asymp b$ which means $a-b\xrightarrow{\text{a.s.}}0$. As defined in \cite{benaych2016spectral}, for a sequence of random square matrices $\textbf{A}$ and $\textbf{B}$ of size $n$, $\textbf{A}\leftrightarrow \textbf{B}$ means that $\frac{1}{n}\text{tr}\left\{\textbf{D}\left(\textbf{A}-\textbf{B}\right)\right\}\xrightarrow[]{p}0$ and $\textbf{d}_1^T\left(\textbf{A}-\textbf{B}\right)\textbf{d}_2\xrightarrow[]{p}0$ for all sequences $\textbf{D}$ of deterministic $n\times n$ matrices of bounded norms and all deterministic sequences of vectors $\textbf{d}_1$, $\textbf{d}_2$ of bounded norms. The function $\Phi(\cdot)$ denotes the standard Gaussian cumulative distribution function and the $\sim$ symbol stands for `distributed as'.

\section{Weight Vector Tuning Procedure}
Consider a supervised classification problem in which a test point $\textbf{x}\in\mathbb{R}^p$ is to be labeled as belonging to one of two classes $\mathcal{C}_0$ and $\mathcal{C}_1$. A linear classification approach to this problem imposes a discriminant of the form
\begin{equation}
    \textbf{w}^T\textbf{x}+{w}_0, \label{discForm}
\end{equation}
characterized by a weight vector, $\textbf{w}\in\mathbb{R}^p$, and bias, $w_0\in\mathbb{R}$. The decision rule 
\begin{equation}
    C(\textbf{x})=\mathbbm{1}\left\{{\textbf{w}}^T\textbf{x}+{w}_0>0\right\}
    \label{decForm}
\end{equation}
based on \eqref{discForm} then classifies $\textbf{x}$ to one of the two classes, i.e., $C(\textbf{x})=i$ indicates that $\textbf{x}$ is classified to class $\mathcal{C}_i, \ i=0,1$. Examples of classifiers which fit this form include LDA, SVM and Least-Squares SVM (both using linear kernels), and Regularized LDA (R-LDA).

In this paper, we propose a method of tuning the weight vector $\textbf{w}$, which reduces the non-discriminative `noisy' components of the original discriminant \eqref{discForm}. As a result, the modified discriminant achieves a testing error rate at least as good as the original and, in certain cases, much better. 

Throughout this paper, let the means and covariances of classes $\mathcal{C}_0$ and $\mathcal{C}_1$ be denoted by $\bm{\mu}_0$, $\bm{\Sigma}_0$ and $\bm{\mu}_1$, $\bm{\Sigma}_1$ respectively. In Section \ref{knownMeans}, we explore an ideal case in which the discriminant neatly decomposes into separate information and noise terms and the noises cancel out optimally in a linear fashion under the assumption of perfectly known means and that $\mathcal{C}_0$ and $\mathcal{C}_1$ makeup a Gaussian mixture model with common class covariance $\bm{\Sigma}_0=\bm{\Sigma}_1=\bm{\Sigma}$. Inspired by the findings of Section \ref{knownMeans}, in Section \ref{parameter} we heuristically extend this result to a more practical scenario  which assumes unknown means and no restriction on the class distributions.

\subsection{Known Class Means}\label{knownMeans}
In this section, assume that the data distribution means $\bm{\mu}_0$ and $\bm{\mu}_1$ are known exactly and that  $\bm{\Sigma}_0=\bm{\Sigma}_1=\bm{\Sigma}$. We proceed to derive a noise-minimized version of \eqref{discForm}.

Consider the shifted test point $\tilde{\textbf{x}}=\textbf{x}-\frac{\bm{\mu}_0+\bm{\mu}_1}{2}$. For any given classifier with weight vector $\textbf{w}$, we show that the projection of $\tilde{\textbf{x}}$ onto $\textbf{w}$, i.e., $\textbf{w}^T\tilde{\textbf{x}}$, can be decomposed into `informative' components which aid in discriminating the class of $\textbf{x}$ and `noisy' components which interfere with discriminating the class of $\textbf{x}$. We then take advantage of this hidden structure for the purpose of reducing the overall noise and obtaining a better classifier.

Let $\bm{\mu}=\bm{\mu}_1-\bm{\mu}_0$. The expression $\tilde{\textbf{x}}$ can be expressed as the sum of its projection onto $\bm{\mu}$ and projection orthogonal to $\bm{\mu}$ as
\begin{equation}
   \tilde{\textbf{x}}=\frac{\bm{\mu}\bm{\mu}^T}{\bm{\mu}^T\bm{\mu}}\tilde{\textbf{x}}+\textbf{P}_{\bm{\mu}}\tilde{\textbf{x}}\label{deco}
\end{equation}
where $\textbf{P}_{\bm{\mu}}=\left(\textbf{I}-\frac{\bm{\mu}\bm{\mu}^T}{\bm{\mu}^T\bm{\mu}}\right)$ is the projection orthogonal to $\bm{\mu}$. Substituting \eqref{deco} into $\textbf{w}^T\tilde{\textbf{x}}$ results in the decomposition of $\textbf{w}^T\tilde{\textbf{x}}$ as
\begin{align}
  \frac{\textbf{w}^T\bm{\mu}}{\bm{\mu}^T\bm{\mu}}\bm{\mu}^T\tilde{\textbf{x}}+\textbf{w}^T\textbf{P}_{\bm{\mu}}\tilde{\textbf{x}}
     \label{disc}
\end{align}
We now show that the first term in \eqref{disc} is composed of an informative component and noisy component with respect to $\textbf{x}$, while the second term consists solely of noise. 
% The following argument is presented in general terms with respect to the class of the test point $\textbf{x}$.
Assume $\textbf{x}\in\mathcal{C}_i$, where $i$ is either 0 or 1. Then, assuming the Gaussian mixture model
\begin{equation}
    \textbf{x}|\textbf{x}\in\mathcal{C}_i\sim\mathcal{N}\left(\bm{\mu}_i,\bm{\Sigma}\right) \label{dist}
\end{equation}
with $i^{\text{th}}$ class prior $\pi_i=P[\textbf{x}\in\mathcal{C}_i]$, we have
$\textbf{x}\sim\bm{\mu}_i+\bm{\Sigma}^{1/2}\textbf{z}$ where $\textbf{z}\sim\mathcal{N}\left(\textbf{0},\textbf{I}\right)$. The first term in \eqref{disc} is then distributed as follows
\begin{align}
   \frac{\textbf{w}^T\bm{\mu}}{\bm{\mu}^T\bm{\mu}}\bm{\mu}^T\tilde{\textbf{x}}&\sim\frac{\textbf{w}^T\bm{\mu}}{\bm{\mu}^T\bm{\mu}}\bm{\mu}^T\left({(-1)^{i+1}\frac{\bm{\mu}}{2}}+{\bm{\Sigma}^{1/2}\textbf{z}}\right)\\
    &=\underbrace{(-1)^{i+1}\frac{\textbf{w}^T\bm{\mu}}{2}}_{I_1 (\text{information})}+\overbrace{\frac{\textbf{w}^T\bm{\mu}}{\bm{\mu}^T\bm{\mu}}\bm{\mu}^T\bm{\Sigma}^{1/2}\textbf{z}}^{N_1 (\text{noise})}\label{one}
    \end{align}
    The first term in \eqref{one} carries information about the class of $\textbf{x}$ through its sign. The second term is the same regardless of the class of $\textbf{x}$ and therefore carries no discriminating information. This is a direct result of assuming a common covariance between $\mathcal{C}_0$ and $\mathcal{C}_1$. The informative component is denoted by $I_1$ while the noisy component with respect to $\textbf{x}$ is denoted by $N_1$. The second term of \eqref{disc} is
\begin{align}
  \textbf{w}^T\textbf{P}_{\bm{\mu}}\tilde{\textbf{x}}&\sim\textbf{w}^T\textbf{P}_{\bm{\mu}}\left({(-1)^{i+1}\frac{\bm{\mu}}{2}}+{\bm{\Sigma}^{1/2}\textbf{z}}\right)\\
    &=\underbrace{\textbf{w}^T\textbf{P}_{\bm{\mu}}\bm{\Sigma}^{1/2}\textbf{z}}_{N_2 (\text{noise})}\label{two}
\end{align}
The discriminatory component of this term is lost in the orthogonal projection, and therefore this term consists solely of noise with respect to the testing point, denoted by $N_2$.

In the interest of achieving better classification performance, we wish to reduce the overall noise content in the discriminant. To this end, consider the following modification of the discriminant \eqref{disc},
\begin{align}
\frac{\textbf{w}^T\bm{\mu}}{\bm{\mu}^T\bm{\mu}}\bm{\mu}^T\tilde{\textbf{x}}+g\left(\textbf{w}^T\textbf{P}_{\bm{\mu}}\tilde{\textbf{x}}\right)\label{general}
\end{align}
for any function $g(\cdot)$, and which, by the above analysis, is equivalent to
\begin{equation}
    I_1+N_1+g\left(N_2\right)\label{first}
\end{equation}
The optimal $g(\cdot)$ such that
$$\mathbb{E}\left[({N_1}+g\left(N_2\right))^2\right]$$ is minimized is the MMSE estimator $\mathbb{E}[-N_1|N_2]$. This choice of $g(\cdot)$ has the effect of minimizing the total noise in the discriminant in the mean square error sense. In the following Lemma 1, we derive the exact form of $g(\cdot)$ for a given $\textbf{w}$ based on the class distribution assumptions \eqref{dist}. 

% Note throughout the derivation that the resulting expression for $g(\cdot)$ holds regardless of the test point's actual class (which, in practice, is not known in advance), since the quantities involved are the same in either case. 

\textbf{Lemma 1} \textit{The optimal $g(N_2)$ is the linear function of $N_2$ given by}
    $g^*(N_2)=\alpha_{\text{MMSE}}(\textbf{w})N_2 $, where 
\begin{equation}
    \alpha_{\text{MMSE}}(\textbf{w})=-\frac{\textbf{w}^T\bm{\mu}}{\bm{\mu}^T\bm{\mu}}\frac{\bm{\mu}^T\bm{\Sigma}\textbf{P}_{\bm{\mu}}\textbf{w}}{\textbf{w}^T\textbf{P}_{\bm{\mu}}\bm{\Sigma}\textbf{P}_{\bm{\mu}}\textbf{w}}.\label{alpha}
\end{equation}
    
\textbf{Proof:}
Given $\textbf{w}$, 
\begin{equation}
    -N_1=-\frac{\textbf{w}^T\bm{\mu}}{\bm{\mu}^T\bm{\mu}}\bm{\mu}^T\bm{\Sigma}^{1/2}\textbf{z}\sim\mathcal{N}\left(0,\left(\frac{\textbf{w}^T\bm{\mu}}{\bm{\mu}^T\bm{\mu}}\right)^2\bm{\mu}^T\bm{\Sigma}\bm{\mu}\right)
\end{equation}
and 
\begin{equation}
    N_2=\textbf{w}^T\textbf{P}_{\bm{\mu}}\bm{\Sigma}^{1/2}\textbf{z}\sim\mathcal{N}\left(0,\textbf{w}^T\textbf{P}_{\bm{\mu}}\bm{\Sigma}\textbf{P}_{\bm{\mu}}\textbf{w}\right)
\end{equation}
are jointly Gaussian random variables. Thus, the optimal $g^*(N_2)=\mathbb{E}[-N_1|N_2]$ reduces to a linear function of $N_2$ given by
\begin{align}
    g^*(N_2)
    &=\frac{\text{Cov}[-N_1,N_2]}{\text{Var}[N_2]}(N_2-\mathbb{E}[N_2])+\mathbb{E}[-N_1]\label{MMSE}\\
    &=-\frac{\textbf{w}^T\bm{\mu}}{\bm{\mu}^T\bm{\mu}}\frac{\bm{\mu}^T\bm{\Sigma}\textbf{P}_{\bm{\mu}}\textbf{w}}{\textbf{w}^T\textbf{P}_{\bm{\mu}}\bm{\Sigma}\textbf{P}_{\bm{\mu}}\textbf{w}}N_2.
\end{align}
Note that $N_2$ is observable only through the expression $\textbf{w}^T\textbf{P}_{\bm{\mu}}\tilde{\textbf{x}}$ and so when using this result we replace $N_2$ by its observable counterpart. 

 Based on this result, we have the following theorem.

\textbf{Theorem 1} \textit{The discriminant that minimizes the noise with respect to the test point in the MSE sense for a given $\textbf{w}$, known means, and under the data distribution assumptions of \eqref{dist}, is} 
\begin{align}
    \frac{\textbf{w}^T\bm{\mu}}{\bm{\mu}^T\bm{\mu}}\bm{\mu}^T\tilde{\textbf{x}}+\alpha_{\text{MMSE}}(\textbf{w})\textbf{w}^T\textbf{P}_{\bm{\mu}}\tilde{\textbf{x}},\label{mod}
\end{align}
or, equivalently, 
\begin{equation}
    \textbf{w}'^T\textbf{x}+w'_0,
\end{equation}
\textit{where}
\begin{equation}
    \textbf{w}'=\frac{\textbf{w}^T\bm{\mu}}{\bm{\mu}^T\bm{\mu}}\bm{\mu}+\alpha_{\text{MMSE}}(\textbf{w})\textbf{P}_{\bm{\mu}}\textbf{w}
\end{equation}
and
\begin{equation}
    w'_0={-\frac{1}{2}\left(\frac{\textbf{w}^T\bm{\mu}}{\bm{\mu}^T\bm{\mu}}\bm{\mu}+\alpha_{\text{MMSE}}(\textbf{w})\textbf{P}_{\bm{\mu}}\textbf{w}\right)^T{\left(\bm{\mu}_0+\bm{\mu}_1\right)}}.
\end{equation}

This result is obtained by simply evaluating \eqref{general} using $g^*(\cdot)$. We make several remarks concerning this result. Firstly, the modified discriminant is linear. This is a direct result of the Gaussian assumption \eqref{dist}, which, while not technically necessary, is desirable, as it produces a simple linear form which inspires the parameterized formulation presented in the next section.  Secondly, the original weight vector $\textbf{w}$ is modified to $\textbf{w}'$ and a bias $w_0'$ is generated. This bias is the optimal bias in the sense of minimizing the probability of misclassification under the class distribution assumptions of \eqref{dist} and equal class priors when fixing the weight vector to $\textbf{w}'$  (see \cite{mai2012direct} Proposition 2). Finally, viewing the modified discriminant \eqref{mod} as a function of a parameter $\alpha$ as follows
\begin{equation}
    \frac{\textbf{w}^T\bm{\mu}}{\bm{\mu}^T\bm{\mu}}\bm{\mu}^T\tilde{\textbf{x}}+\alpha\textbf{w}^T\textbf{P}_{\bm{\mu}}\tilde{\textbf{x}},\label{para}
\end{equation}
$\alpha=\alpha_{\text{MMSE}}(\textbf{w})$ yields a stationary point of its probability of misclassification and achieves the minimum probability of misclassification when $\textbf{w}^T\bm{\mu}>0$. This is demonstrated in Section \ref{exp1}.

The following corollary of Theorem 1 lends intuition as well as credibility to this technique by showing that it recovers the Bayes optimal classifier discriminant for the assumed class distributions from its weight vector. The Bayes classifier in this case is linear. It is the LDA classifier, with decision rule
 \begin{equation}
   \mathbbm{1}\left\{\bm{\mu}^T\bm{\Sigma}^{-1}\tilde{\textbf{x}}+\text{ln}\frac{\pi_1}{\pi_0}>0\right\}.\label{Bayes}
\end{equation}
The LDA weight vector is $\textbf{w}=\bm{\Sigma}^{-1}\bm{\mu}$. 

\textbf{Corollary 1} \textit{Computing the parameter \eqref{alpha} corresponding to the LDA classifier \eqref{Bayes} yields}
\begin{align}
    \alpha_{\text{MMSE}}\left(\bm{\Sigma}^{-1}\bm{\mu}\right)=1
\end{align}
\textit{and the resulting discriminant \eqref{mod} recovers the LDA discriminant in \eqref{Bayes} when the class priors are equal.}

Since there is no modification of the weight vector, we conclude that the LDA weight vector (in the case of known statistics) is optimal relative to itself in that it achieves the minimum noise (in the mean square error sense) with respect to the test point under the assumed class distributions. 

\subsubsection{Experiments with Known Means}\label{exp1}
For the following simulation and any simulations involving synthetic data in the remainder of this paper, the exact expected testing error/probability of misclassification of a linear classifier learned on a given training set is computed using knowledge of the data distribution from which the testing data is generated. All synthetic data in this paper is generated from a two-class Gaussian mixture model. The expected testing error under these data distribution assumptions of a generic binary linear classifier 
\begin{equation}
    \mathbbm{1}\left\{{\bm{\beta}}^T\textbf{x}+{\beta}_0>0\right\},
\end{equation}
with weight vector $\bm{\beta}$ and intercept ${\beta}_0$, can easily be derived as (see Lemma 1 in \cite{niyazi2020asymptoticARXIV})
\begin{equation}
\pi_0\Phi\left(\frac{\bm{\beta}^T\bm{\mu}_0+\beta_0}{\sqrt{\bm{\beta}^T\bm{\Sigma}_0\bm{\beta}}}\right)+\pi_1\Phi\left(-\frac{\bm{\beta}^T\bm{\mu}_1+\beta_0}{\sqrt{\bm{\beta}^T\bm{\Sigma}_1\bm{\beta}}}\right).\label{herewego}
\end{equation}

Now consider the parameterized version \eqref{para} of \eqref{mod}. The objective of the following simulation is to show that  $\alpha_{\text{MMSE}}(\textbf{w})$ given by \eqref{alpha} coincides with the $\alpha$ yielding a stationary point of the expected testing error of \eqref{para}. The stationary point is a minimum when $\textbf{w}^T\bm{\mu}>0$ and is otherwise a maximum, as in that case, the orientation of $\textbf{w}$ flips the class labels.

To demonstrate this, a weight vector $\textbf{w}$ is uniformly sampled from all $\textbf{w}$ such that $\lVert\textbf{w}\rVert_2=1$ using the method in \cite{weisstein2017hypersphere}. It is then fed to \eqref{para} and the exact expected testing error with varying $\alpha$ is plotted using \eqref{herewego}. The quantity $\alpha_{\text{MMSE}}\left(\textbf{w}\right)$ is then computed from \eqref{alpha} for comparison. The class statistics used for this simulation are 
\begin{equation}
   \bm{\mu}_0=\frac{1}{p^{1/4}}\left[\textbf{1}^T_{\lceil{\sqrt{p}}\rceil} \ \textbf{0}^T_{p-\lceil{\sqrt{p}}\rceil-2} \ 2 \ 2\right]^T, \ \bm{\mu}_1=\textbf{0}_p, \label{setting1}
\end{equation}
and 
\begin{equation}
    \bm{\Sigma}_0=\bm{\Sigma}_1=\frac{10}{p}\textbf{1}_p\textbf{1}_p^T+0.1\textbf{I}_p \label{setting2}
\end{equation}
where $p=200$. Here, $\pi_0=\pi_1=0.5$.

\begin{figure*}[t!]
    \centering
    \begin{subfigure}[t]{0.5\textwidth}
        \centering
        \includegraphics[width=0.9\linewidth]{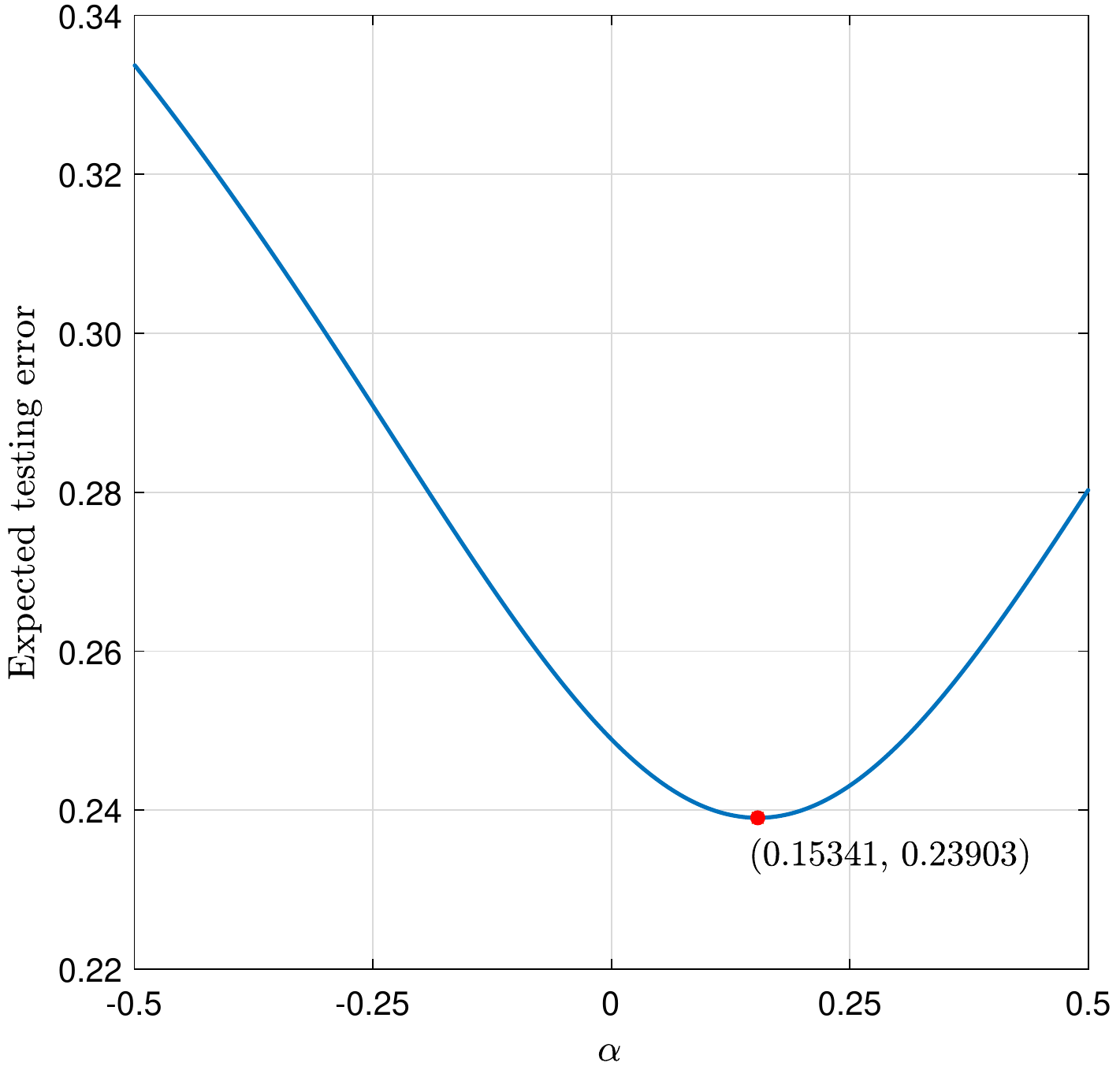}
        \caption{Here $\textbf{w}^T\bm{\mu}>0$ and $\alpha_{\text{MMSE}}\left(\textbf{w}\right)=0.1534$ coincides with the $\alpha$ yielding the minimum expected testing error}
        \label{fig1a}
    \end{subfigure}%
    ~ 
    \begin{subfigure}[t]{0.5\textwidth}
        \centering
        \includegraphics[width=0.9\linewidth]{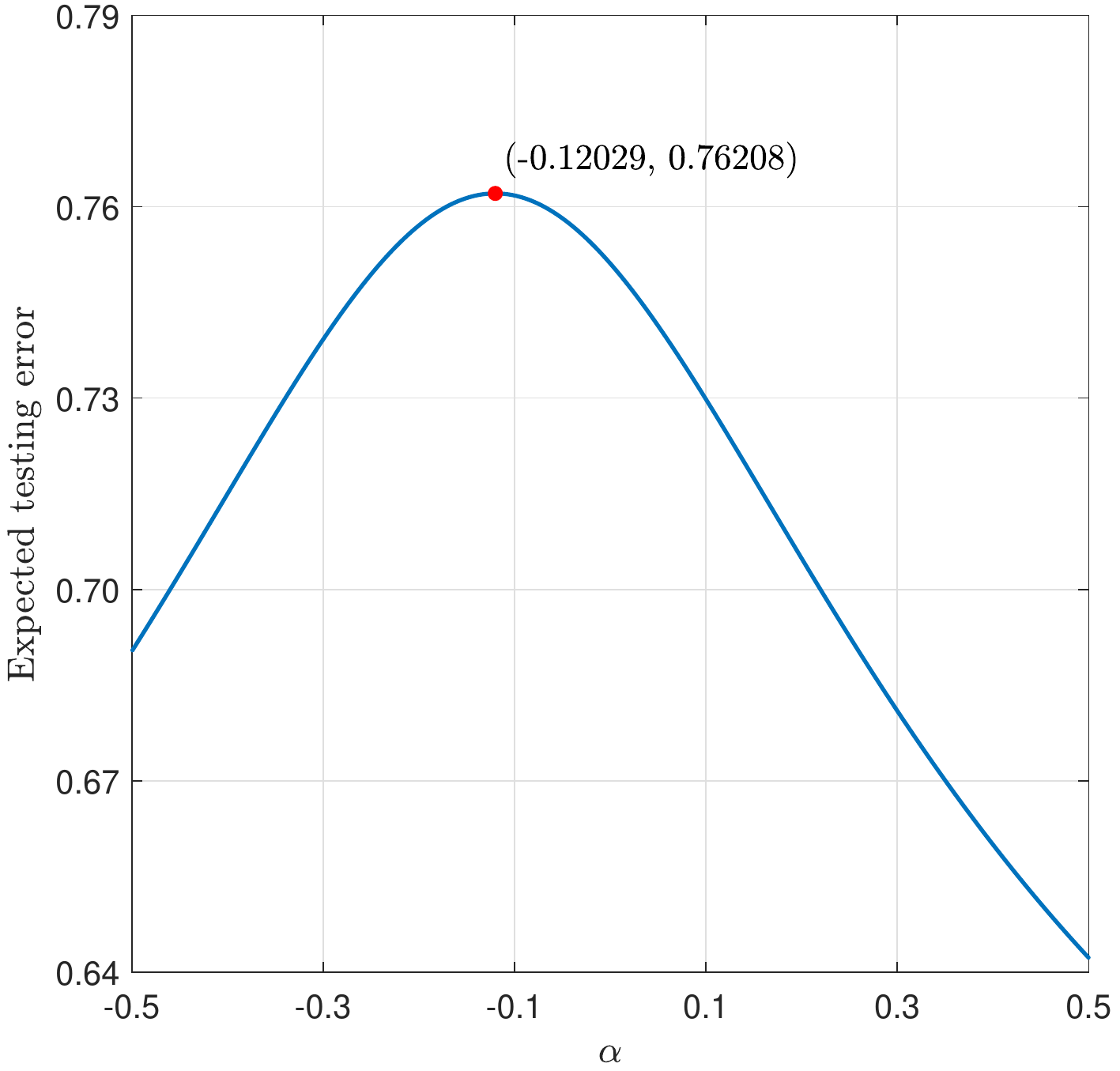}
        \caption{Here $\textbf{w}^T\bm{\mu}<0$ and $\alpha_{\text{MMSE}}\left(\textbf{w}\right)=-0.1203$ coincides with the $\alpha$ yielding the maximum expected testing error}
        \label{fig1b}
    \end{subfigure}
    \caption{Plot of the expected testing error of \eqref{para} against $\alpha$ for a randomly generated weight vector $\textbf{w}$}
    \label{fig1}
\end{figure*}
Figure \ref{fig1a} and Figure \ref{fig1b} show the results when $\textbf{w}^T\bm{\mu}>0$ and $\textbf{w}^T\bm{\mu}<0$, respectively. In Figure \ref{fig1a}, the minimum expected testing error occurs at $\alpha=0.15341$. This exactly coincides with  $\alpha_{\text{MMSE}}\left(\textbf{w}\right)$ of Theorem 1 that minimizes the noise in the discriminant. In Figure \ref{fig1b}, the \textit{maximum} expected testing error occurs at $\alpha=-0.12029$, which, again, exactly coincides with  $\alpha_{\text{MMSE}}\left(\textbf{w}\right)$ that minimizes the noise in the discriminant. The latter discriminant's behavior can be explained by the fact that the orientation of the randomly generated $\textbf{w}$ flips the class labels. Simply taking the negative of $\textbf{w}$ yields a classifier having the \textit{minimum} expected testing error at $\alpha_{\text{MMSE}}\left(\textbf{w}\right)$. In conclusion, minimizing the noise in the discriminant in the MSE sense is equivalent to minimizing the expected testing error, as long as $\textbf{w}$ is sensibly oriented. This motivates using this criteria as the basis for designing a better classifier in the next section.

\subsection{Unknown Class Means}\label{parameter}
The previous section derives the discriminant with minimum noise with respect to the test point for a general binary linear classifier with weight vector $\textbf{w}$ under the assumption of Gaussian classes with known means and a common covariance. A more practical scenario is when all class statistics are unknown and sample statistics are used instead. Using the sample mean estimates introduces an additional estimation noise into the discriminant.

Let the $n$ individual training vectors corresponding to classes $\mathcal{C}_0$ and $\mathcal{C}_1$ make up the columns of the matrices $\textbf{X}_0\in\mathbb{R}^{p\times n_0}$ and $\textbf{X}_1\in\mathbb{R}^{p\times n_1}$, respectively ($n=n_0+n_1$). The maximum likelihood estimates of the class means are given by the sample means $\hat{\bm{\mu}}_0=\frac{1}{n_0}\textbf{X}_0\textbf{1}_{n_0}$ and $\hat{\bm{\mu}}_1=\frac{1}{n_1}\textbf{X}_1\textbf{1}_{n_1}$. Let $\hat{\bm{\mu}}=\hat{\bm{\mu}}_1-\hat{\bm{\mu}}_0$ and $\hat{\tilde{\bm{x}}}=\textbf{x}-\frac{\hat{\bm{\mu}}_0+\hat{\bm{\mu}}_1}{2}$. Given a weight vector, $\textbf{w}$, $\textbf{w}^T\hat{\tilde{\bm{x}}}$ can be expressed as
\begin{align}
    \textbf{w}^T\hat{\tilde{\bm{x}}}&=\frac{\textbf{w}^T\hat{\bm{\mu}}}{\hat{\bm{\mu}}^T\hat{\bm{\mu}}}\hat{\bm{\mu}}^T\hat{\tilde{\bm{x}}}+\textbf{w}^T\textbf{P}_{\hat{\bm{\mu}}}\hat{\tilde{\bm{x}}}\label{genDecoUnknown}
\end{align}
where $\textbf{P}_{\hat{\bm{\mu}}}=\left(\textbf{I}-\frac{\hat{\bm{\mu}}\hat{\bm{\mu}}^T}{\hat{\bm{\mu}}^T\hat{\bm{\mu}}}\right)$. Regardless of the class distributions and whether assuming distinct covariances $\bm{\Sigma}_0$ and $\bm{\Sigma}_1$ or common class covariances $\bm{\Sigma}_0=\bm{\Sigma}_1=\bm{\Sigma}$, following a similar line of logic to the analysis in Section \ref{knownMeans} reveals that, while the first term in \eqref{genDecoUnknown} is similarly composed of both information and noise (whether that be estimation noise, noise from the test point, or both), the second term is not purely noise. In fact, it is informative. This is shown in detail in Appendix \ref{sampleProof}.

% and the overall projection has the form \footnote{Note that the information and noise component variables $I_1$, $I_2$, $N_1$, and $N_2$ are repeatedly redefined throughout this section. This should not pose any issues, as unlike in the previous section, what concerns us here is whether these quantities are information or noise, not their actual values.}
% \begin{equation}
%   \underbrace{\frac{\textbf{w}^T\hat{\bm{\mu}}}{\hat{\bm{\mu}}^T\hat{\bm{\mu}}}\hat{\bm{\mu}}^T\hat{\tilde{\bm{x}}}}_{I_1+N_1}+\underbrace{\textbf{w}^T\left(\textbf{I}-\frac{\hat{\bm{\mu}}\hat{\bm{\mu}}^T}{\hat{\bm{\mu}}^T\hat{\bm{\mu}}}\right)\hat{\tilde{\bm{x}}}}_{I_2+N_2}.\label{common}
% \end{equation}
% Furthermore, when the class covariances are distinct, the second term is entirely information and the overall projection has the form 
% \begin{equation}
%       \underbrace{\frac{\textbf{w}^T\hat{\bm{\mu}}}{\hat{\bm{\mu}}^T\hat{\bm{\mu}}}\hat{\bm{\mu}}^T\hat{\tilde{\bm{x}}}}_{I_1+N_1}+\underbrace{\textbf{w}^T\left(\textbf{I}-\frac{\hat{\bm{\mu}}\hat{\bm{\mu}}^T}{\hat{\bm{\mu}}^T\hat{\bm{\mu}}}\right)\hat{\tilde{\bm{x}}}}_{I_2}\label{distinct}.
% \end{equation}
% This is because in the case of common covariances, the noise presents itself in the form of the common covariance between classes and as estimation noise from the sample means, while in the case of distinct covariances, the covariance itself aids in discriminating between the classes, and the only noise is due to the estimation of the means. More detail can be found in Appendix \ref{sampleProof}.

Thus, when the means are unknown, the approach taken in Section \ref{knownMeans} of minimizing the squared sum of `noise 1' with the second term no longer applies, as the second term is informative. Nonetheless, the interaction of this term with the noise in the first term can potentially yield performance gains and so motivated by Section \ref{knownMeans}, the following parameterized version of the sample statistic equivalent of \eqref{mod} is proposed
 \begin{align}
\frac{\textbf{w}^T\hat{\bm{\mu}}}{\hat{\bm{\mu}}^T\hat{\bm{\mu}}}\hat{\bm{\mu}}^T\hat{\tilde{\bm{x}}}+\alpha\textbf{w}^T\textbf{P}_{\hat{\bm{\mu}}}\hat{\tilde{\bm{x}}}\label{herewego2}
\end{align}
where $\alpha$ is a parameter to be tuned. 
% This is equivalent to
% \begin{equation}
%       I_1+N_1+\alpha I_2+\alpha N_2

% \end{equation}
% when the covariances are equal, and to 
% \begin{equation}
%       I_1+N_1+\alpha I_2
% \end{equation}
% when the covariances are distinct.
 The following Section \ref{simSample} demonstrates that a better misclassification rate may be achieved by setting $\alpha$ to a value that is not equal to one (where $\alpha=1$ recovers the original projection with optimal bias assuming equal priors and the class distribution in \eqref{dist}). A significant improvement is observed when the estimation noise is high.
 
%  The optimal $\alpha$ can be explained as a trade-off between retaining the entirety of $I_2$ at $\alpha=1$ and forfeiting all of or a proportion of $I_2$, at other values of $\alpha$, in return for minimizing either $N_1+N_2$ in the case of common covariances, or just $N_1$ in the case of distinct covariances.  This logic holds regardless of the distributions of $\mathcal{C}_0$ and $\mathcal{C}_1$. (Note, however, that if the data under study is not distributed as in \eqref{dist}, the residual bias is not optimal and might need to be tuned separately.)

%   Not only does this formulation allow for generalizing the analysis in the previous section to unknown means, but it also allows adapting the discriminant to metrics other than the probability of misclassification; The parameter $\alpha$ can be tuned according to the desired metric to obtain a weight vector suited to this metric and the bias can be tuned independent of $\alpha$ according to the same metric.
\subsubsection{Experiments with Unknown Means}\label{simSample}
In this section we explore the behavior of \eqref{herewego2} under a variety of settings and for an assortment of starting weight vectors. We first list and briefly describe the discriminants from which these weight vectors are extracted, namely, LDA, logistic regression, linear support vector machine (SVM), regularized LDA (R-LDA), and randomly-projected LDA ensemble (RP-LDA). 
\begin{itemize}
    \item \textbf{LDA} (see \cite{friedman2001elements}) in the form \eqref{Bayes} is the Bayes classifier for data distributed as \eqref{dist}. In practice, the class statistics are unknown and sample estimates are used instead. The sample means $\hat{\bm{\mu}}_0$ and $\hat{\bm{\mu}}_1$ are defined at the beginning of Section \ref{parameter}. The maximum likelihood estimates of the common covariance matrix and class priors are the pooled sample covariance matrix  
    \begin{equation}
        \hat{\bm{\Sigma}}=\frac{(n_0-1)\hat{\bm{\Sigma}}_0+(n_1-1)\hat{\bm{\Sigma}}_1}{n_0+n_1-2},
    \end{equation} 
  where $\hat{\bm{\Sigma}}_0=\frac{1}{n_0-1}\left(\textbf{X}_0-\hat{\bm{\mu}}_0\textbf{1}^T\right)\left(\textbf{X}_0-\hat{\bm{\mu}}_0\textbf{1}^T\right)^T$ and $\hat{\bm{\Sigma}}_1=\frac{1}{n_1-1}\left(\textbf{X}_1-\hat{\bm{\mu}}_1\textbf{1}^T\right)\left(\textbf{X}_1-\hat{\bm{\mu}}_1\textbf{1}^T\right)^T$, and the prior estimates $\hat{\pi}_i=\frac{n_i}{n}, \ i=0,1$, respectively. The LDA discriminant is then
 \begin{equation}
 \hat{\bm{\mu}}^T\hat{\bm{\Sigma}}^{-1}\hat{\tilde{\bm{x}}}+\text{ln}\frac{\hat{\pi}_1}{\hat{\pi}_0}.\label{LDA}
\end{equation}
Its weight vector is $\textbf{w}=\hat{\bm{\Sigma}}^{-1}\hat{\bm{\mu}}$. 
    \item For linearly separable training data, \textbf{SVM with linear kernel}  (see \cite{friedman2001elements}) finds a hyperplane that maximizes the margin between one class and the other subject to constraints of perfect classification on the training points. When the training data is linearly inseparable, the constraints are relaxed by penalizing each (possibly) misclassified point. The penalty is a parameter that must be tuned. This variant is called the soft-margin SVM with linear kernel, and it is what we use in this paper.
    \item \textbf{Logistic regression} (see \cite{friedman2001elements}) models the log-odds `$\text{ln}\left( \frac{P[\textbf{x}\in\mathcal{C}_1|\textbf{x}]}{1-P[\textbf{x}\in\mathcal{C}_1|\textbf{x}]}\right)$' as a linear function of the test point. The decision boundary corresponds to the set of points at which the log-odds equals zero. The weight vector and bias of the decision boundary are learned by maximizing the likelihood of the training data.
    \item \textbf{R-LDA} counters the small sample issue in LDA by regularizing the pooled sample covariance estimate before inverting it. There are several possibilities for the form of the regularization (see \cite{guo2007regularized}). In this paper we opt for
    \begin{equation}
 \hat{\bm{\mu}}^T\left(\hat{\bm{\Sigma}}+\gamma\textbf{I}_p\right)^{-1}\hat{\tilde{\bm{x}}}+\text{ln}\frac{\hat{\pi}_1}{\hat{\pi}_0},\label{RLDA}
 \end{equation}
    where $\gamma$ is the regularization parameter that must be tuned. The weight vector here is $\textbf{w}=\left(\hat{\bm{\Sigma}}+\gamma\textbf{I}_p\right)^{-1}\hat{\bm{\mu}}$. 
    \item \textbf{RP-LDA ensemble} (see \cite{durrant2013random}) counters the small sample issue in LDA by reducing the dimensionality of the training samples (and test point) using random matrices. Each projection $\textbf{R}_i\in\mathbb{R}^{d\times p}$ yields a discriminant. These are averaged over all $M$ projections so that the final discriminant has the form
    \begin{equation}
    \frac{1}{M}\sum_{i=1}^M\hat{\bm{\mu}}^T\textbf{R}_i^T(\textbf{R}_i\hat{\Sigma}\textbf{R}_i^T)^{-1}\textbf{R}_i\hat{\tilde{\bm{x}}}+\text{ln}\frac{\hat{\pi}_1}{\hat{\pi}_0}
    \end{equation}
    The weight vector is $\textbf{w}=\frac{1}{M}\sum_{i=1}^M\textbf{R}_i^T(\textbf{R}_i\hat{\Sigma}\textbf{R}_i^T)^{-1}\textbf{R}_i\hat{\bm{\mu}}$. The reduced dimension $d$ is a parameter that must be tuned.
\end{itemize}

For these simulations, we consider two data distributions: data generated from classes having a common covariance and data generated from classes having distinct covariance matrices. We also consider three regimes of $n$ versus $p$: $n$ on the order of $p$ ($p=400$, $n=450$), $n>p$ ($p=10$, $n=500$), and $n<p$ ($p=300$, $n=100$). We apply the appropriate classifiers to each regime. LDA requires $n>p$, soft-margin SVM is applicable in any regime, logistic regression requires $n$ be much greater than $p$ to ensure convergence of the maximum likelihood estimates of the weight vector and bias, and finally, R-LDA and RP-LDA are designed for the regime $n<p$. 

Each classifier is trained on a generated training set. Additionally, for SVM, R-LDA, and RP-LDA, the penalty, $\gamma$, and $d$ parameters are chosen to minimize the expected testing error given that training set. The SVM penalty is tuned within the set $\{10^{-4}, 10^{-3}, 10^{-2}, 10^{-1}, 1, 10, 100, 1000\}$, $\gamma$ within the set $[10^{-4},2]$, in increments of $0.1$, and $d$ from $1$ to the maximum allowable setting of $d=\text{rank}\left(\hat{\bm{\Sigma}}\right)-2$, in increments of $2$. After this is done, we have a weight vector $\textbf{w}$ for each classifier. Each weight vector is fed into \eqref{herewego2} to obtain an $\alpha$-parameterized version of the discriminant. Let us refer to these new classifiers as $\alpha$-LDA, $\alpha$-SVM, $\alpha$-log, $\alpha$-RLDA, and $\alpha$-RPLDA for short. For each $\alpha$-parameterized discriminant, we vary $\alpha$ and compute the expected testing error using \eqref{herewego}. These errors are averaged over $100$ independently generated training sets. Error bars depicting the standard errors are plotted alongside this average. 

Recall that setting $\alpha=1$ in \eqref{herewego2} produces a discriminant having the original weight vector $\textbf{w}$ and a bias with minimum probability of misclassification (under the Gaussian mixture model and equal priors assumption) for that weight vector. In what follows, we use $\alpha=1$ as a reference point for determining whether or not there is a significant improvement in classifier performance at the $\alpha$ achieving the minimum error rate. To quantify the improvement, we report percentage changes relative to the average expected testing error at $\alpha=1$ computed as $\frac{\text{error at } \alpha \text{ achieving the minimum}-\text{error at } \alpha=1}{\text{error at } \alpha=1}\times 100$. This quantity reflects the fact that a given error improvement starting at an already low error rate at the baseline $\alpha=1$ is more significant than when the error is high to start with.

\begin{figure*}[t!]
    \centering
    \begin{subfigure}[t]{0.5\textwidth}
        \centering
        \includegraphics[width=0.9\linewidth]{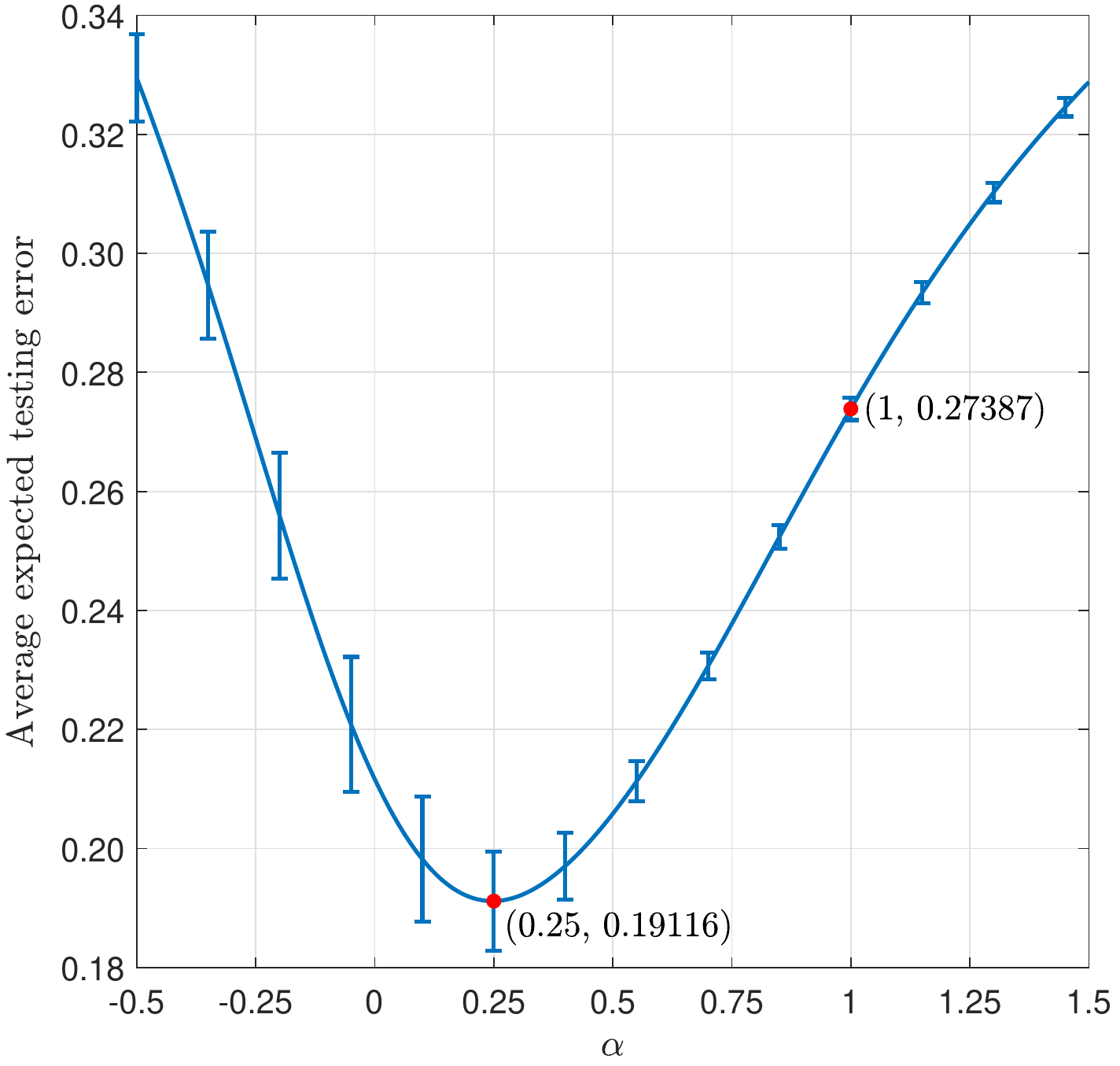}
        \caption{$\alpha$-LDA}
        \label{fig2a}
    \end{subfigure}%
    ~ 
    \begin{subfigure}[t]{0.5\textwidth}
        \centering
        \includegraphics[width=0.9\linewidth]{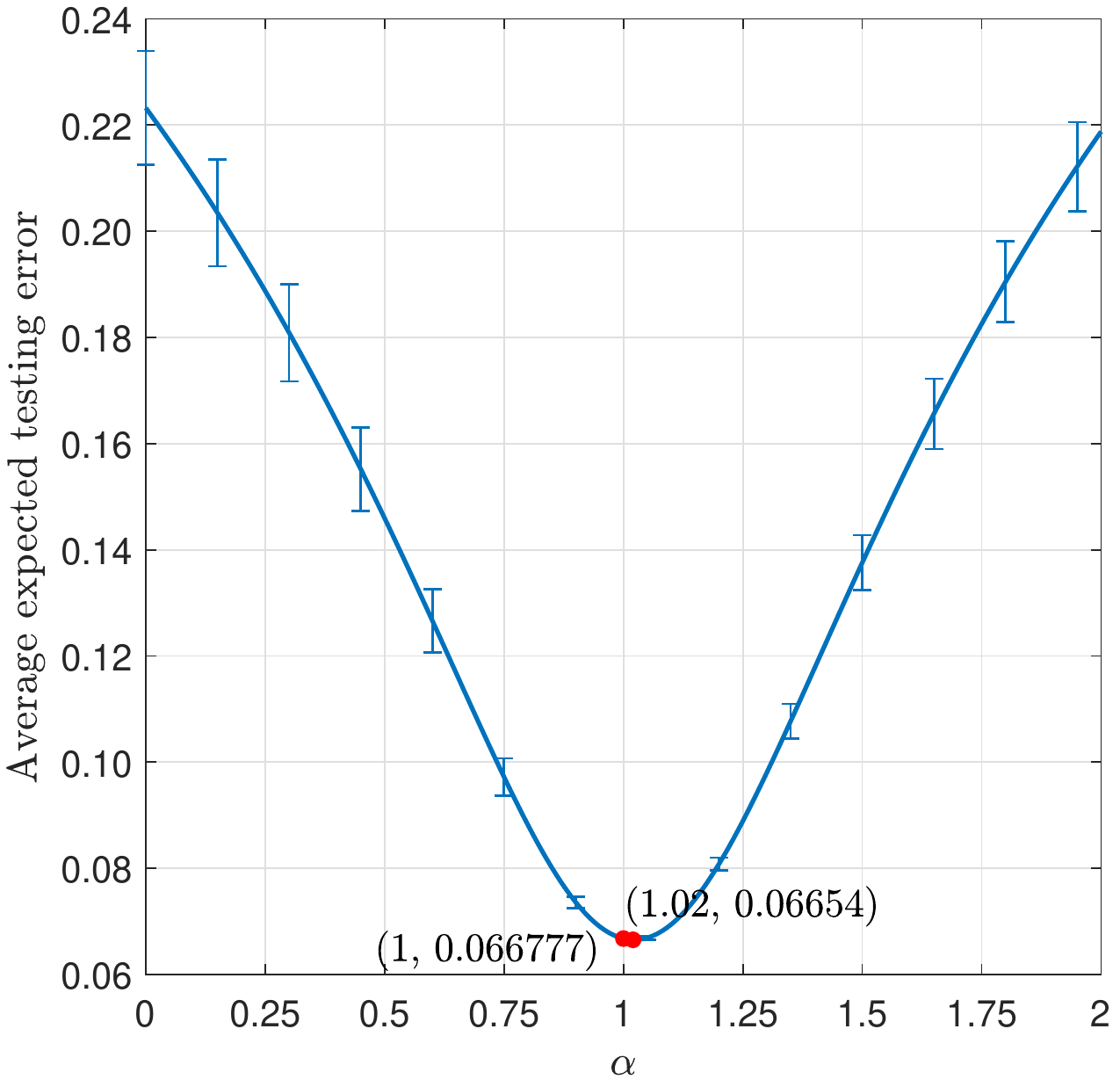}
        \caption{$\alpha$-SVM}
        \label{fig2b}
    \end{subfigure}
    \caption{Plots of expected testing error averaged over $100$ training sets for data generated from classes with a common $\bm{\Sigma}$. Here, $p=400$ and $n=450$.}
    \label{fig2}
\end{figure*}

\begin{figure*}[t!]
    \centering
    \begin{subfigure}[t]{0.3\textwidth}
        \centering
        \includegraphics[width=1\linewidth]{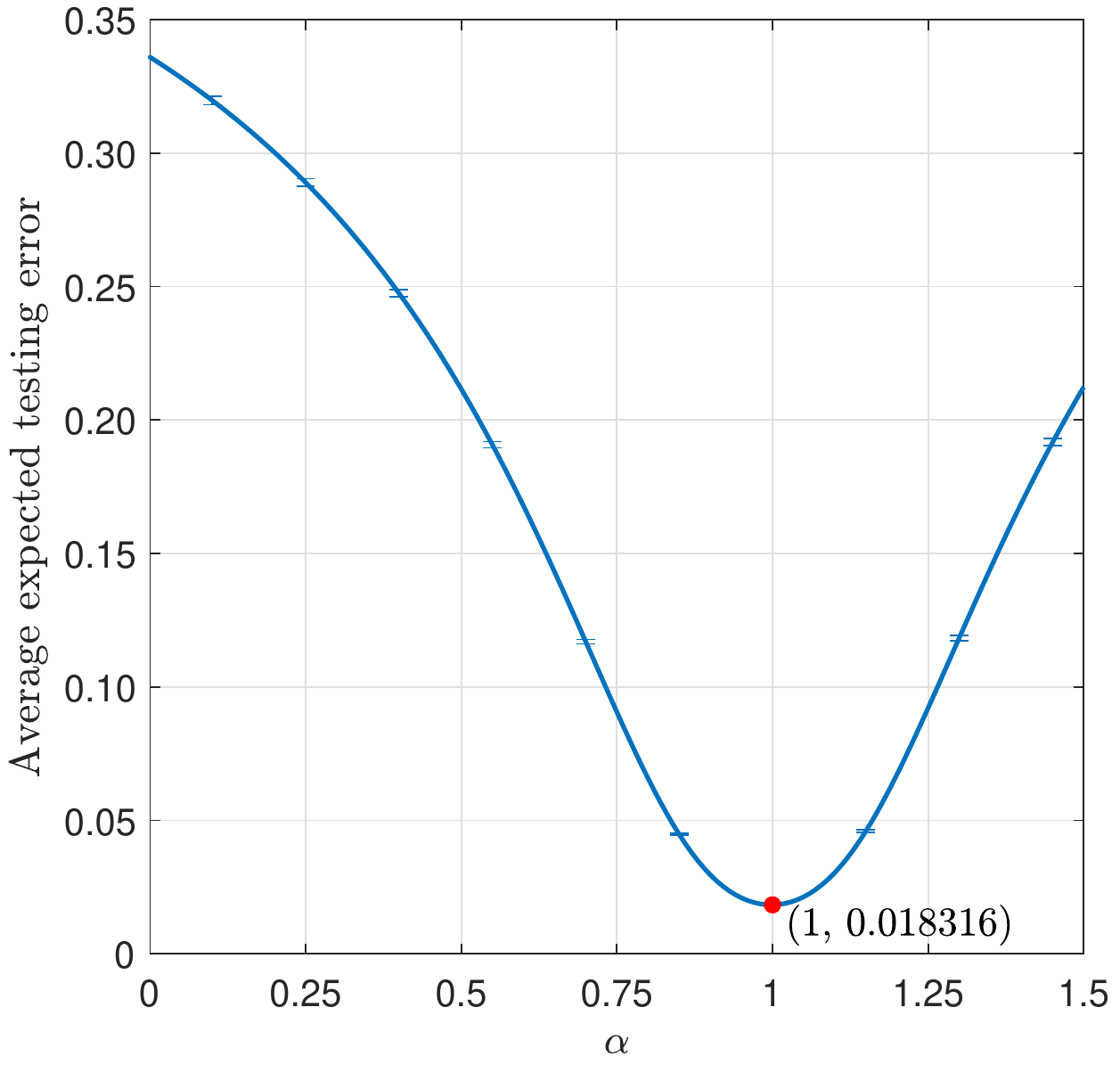}
        \caption{$\alpha$-LDA}
        \label{fig3a}
    \end{subfigure}%
    ~ 
    \begin{subfigure}[t]{0.3\textwidth}
        \centering
        \includegraphics[width=1\linewidth]{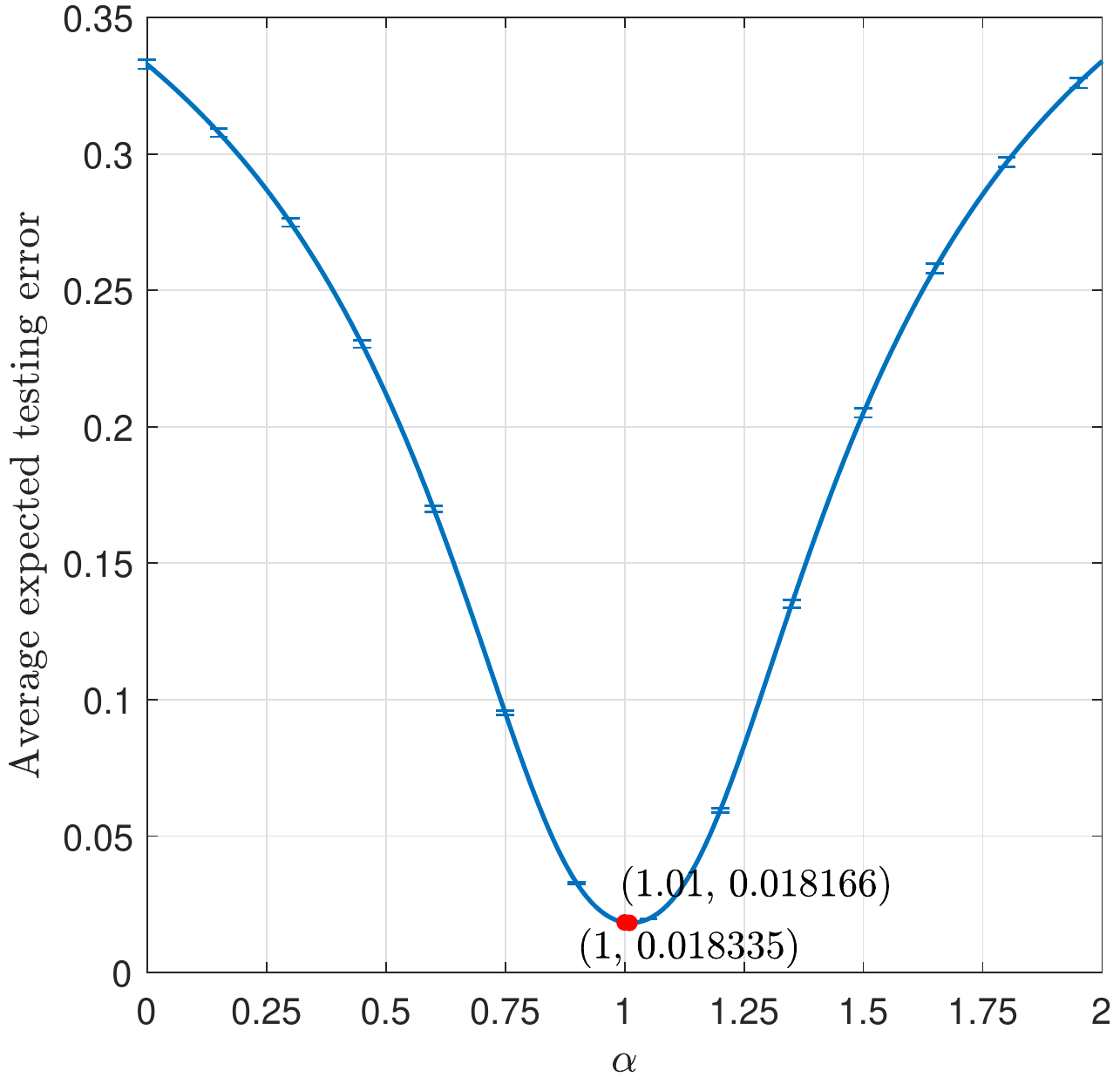}
        \caption{$\alpha$-SVM}
        \label{fig3b}
    \end{subfigure}
    ~
      \begin{subfigure}[t]{0.3\textwidth}
        \centering
        \includegraphics[width=1\linewidth]{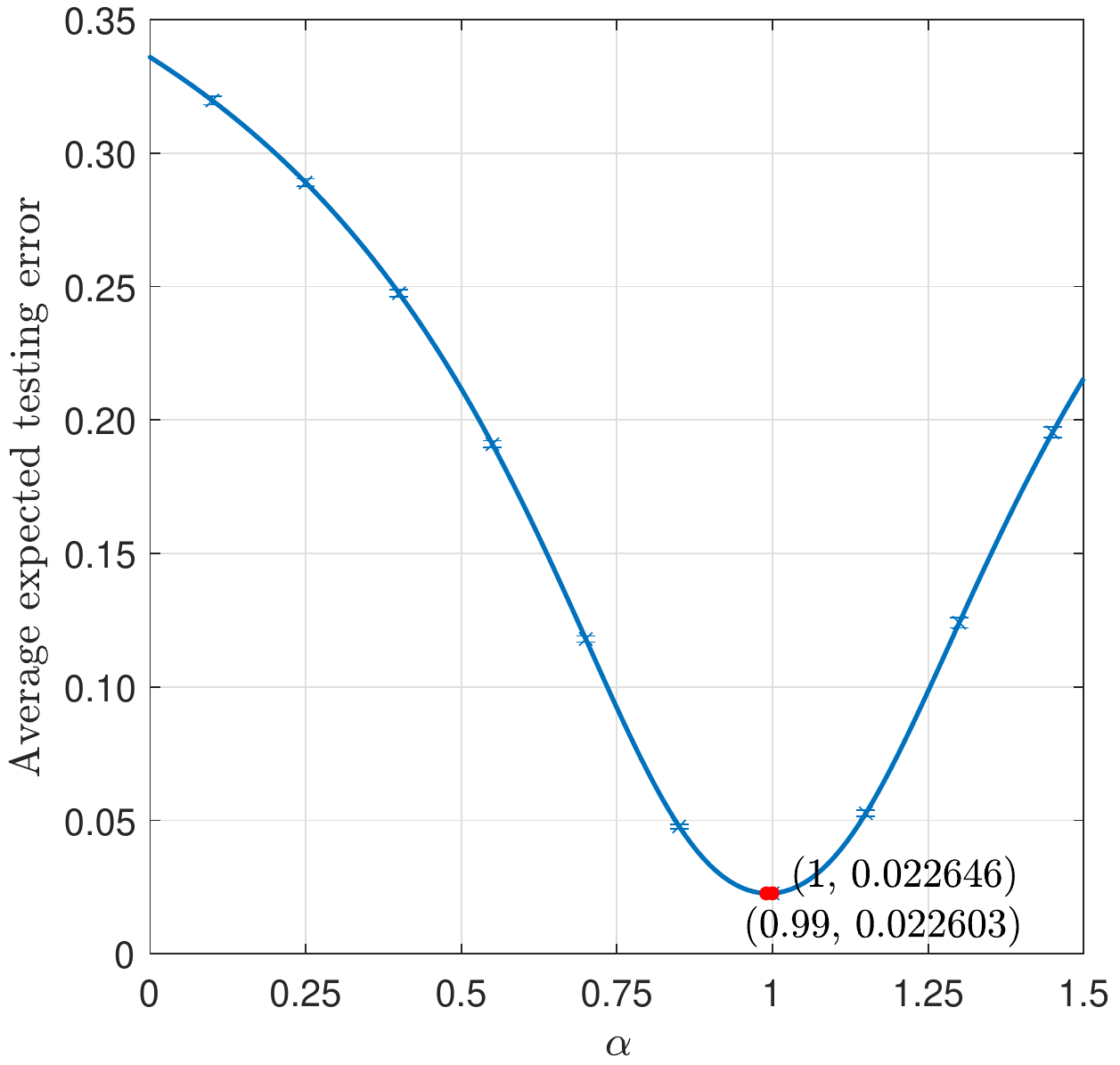}
        \caption{$\alpha$-log}
        \label{fig3c}
    \end{subfigure}
    \caption{Plots of expected testing error averaged over $100$ training sets for data generated from classes with a common $\bm{\Sigma}$. Here, $p=10$ and $n=500$.}
    \label{fig3}
\end{figure*}

\begin{figure*}[t!]
    \centering
    \begin{subfigure}[t]{0.3\textwidth}
        \centering
        \includegraphics[width=1\linewidth]{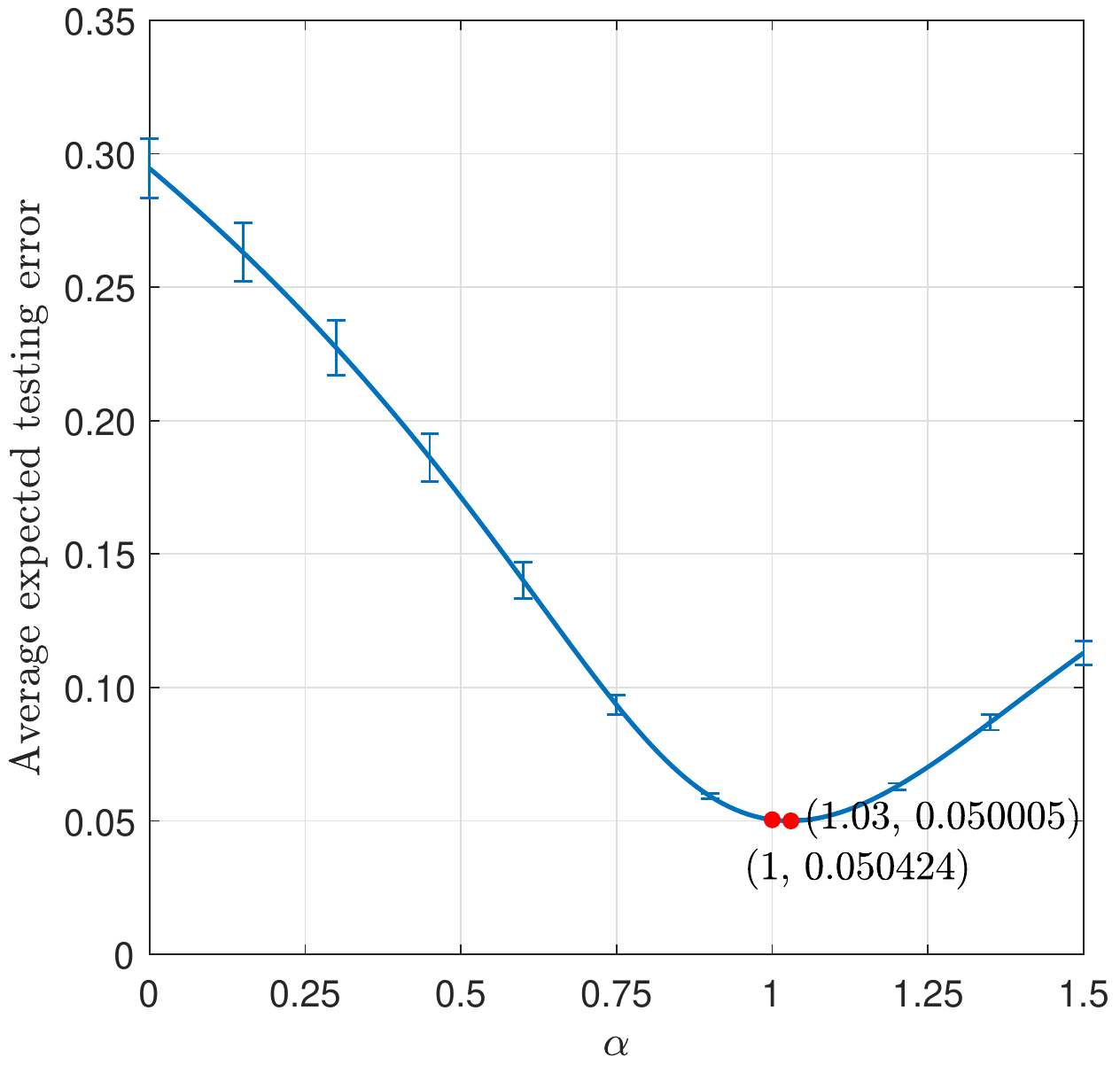}
        \caption{$\alpha$-RLDA}
        \label{fig4a}
    \end{subfigure}%
    ~ 
    \begin{subfigure}[t]{0.3\textwidth}
        \centering
        \includegraphics[width=1\linewidth]{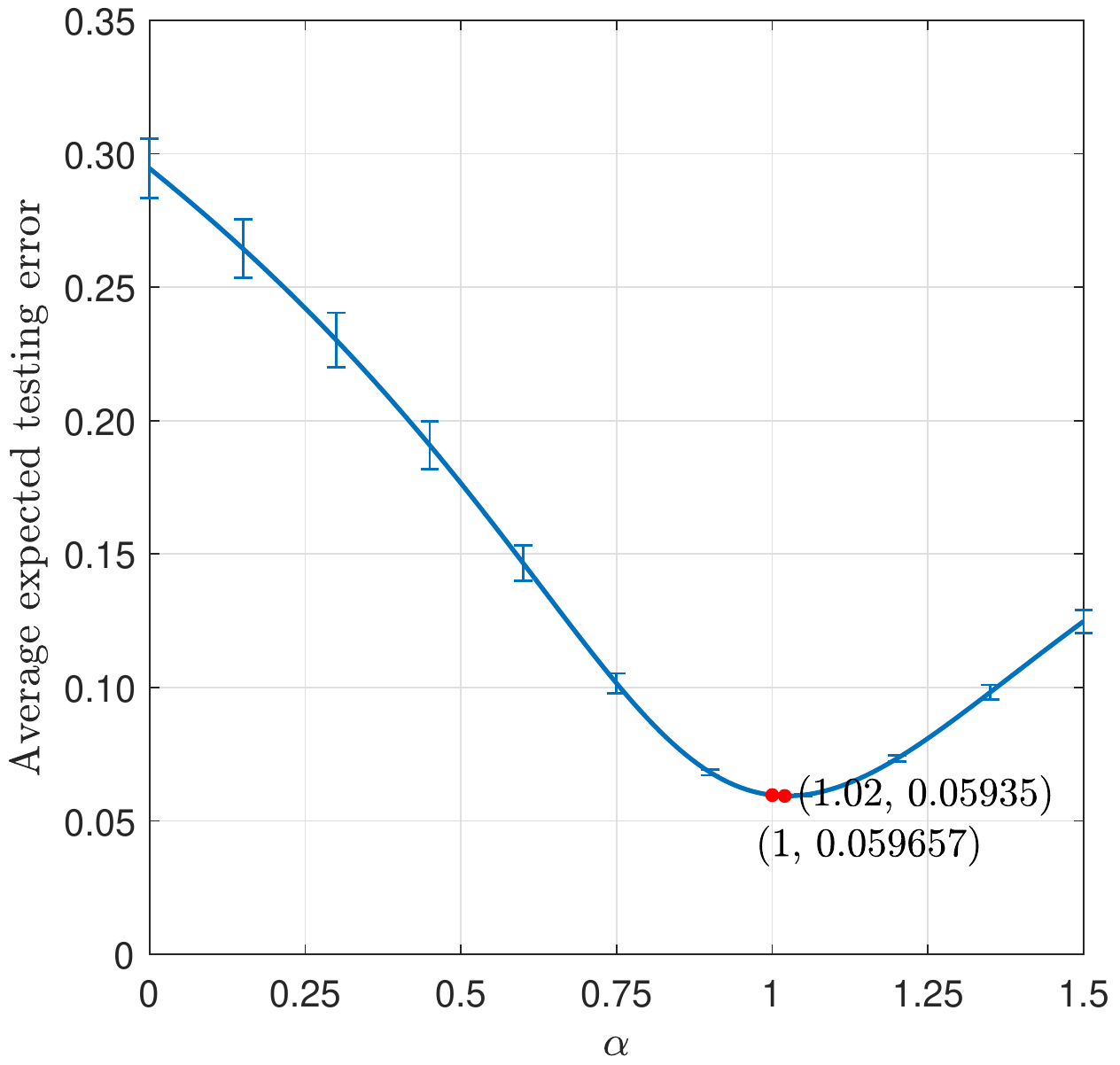}
        \caption{$\alpha$-RPLDA}
        \label{fig4b}
    \end{subfigure}
    ~
      \begin{subfigure}[t]{0.3\textwidth}
        \centering
        \includegraphics[width=1\linewidth]{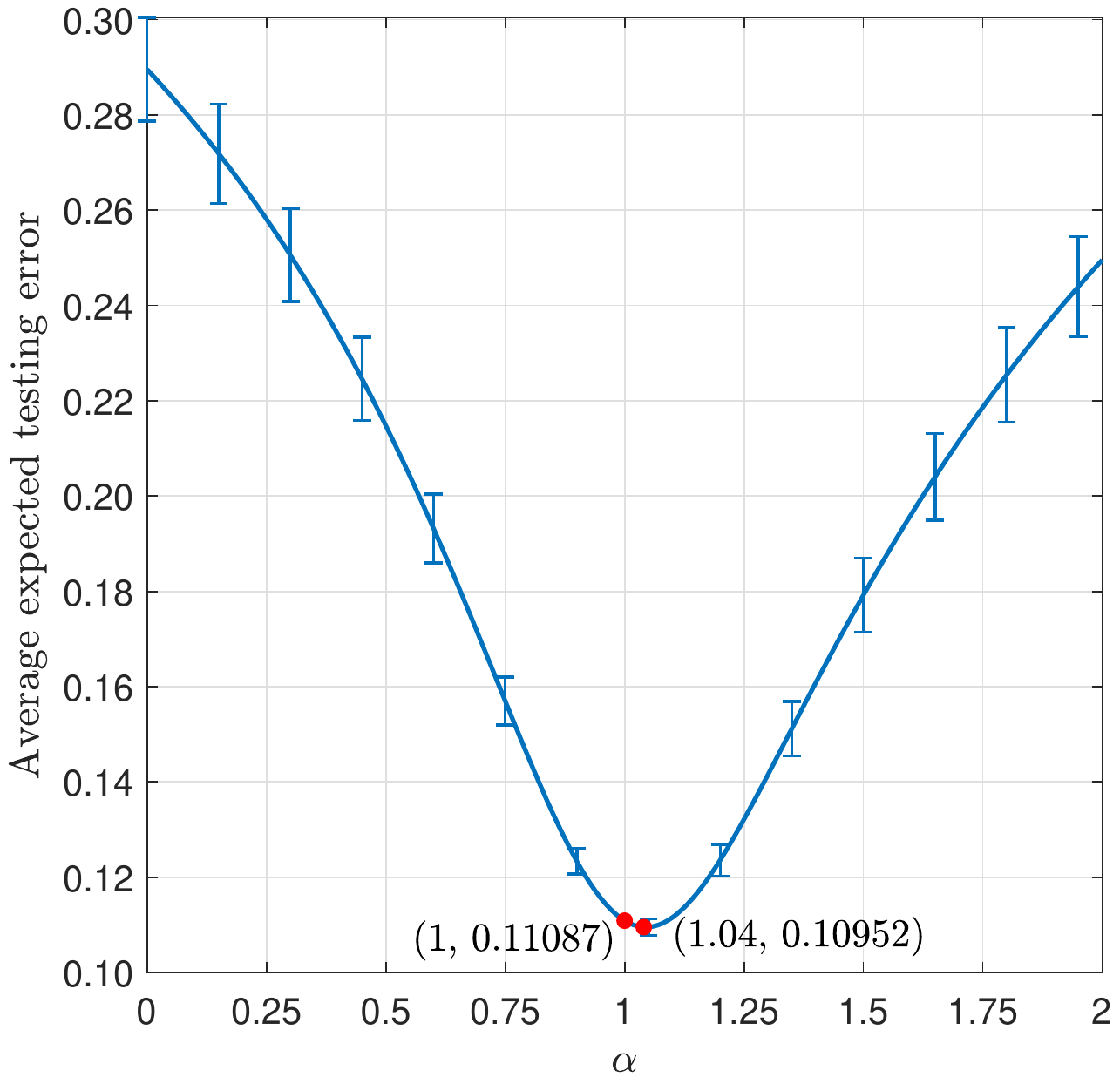}
        \caption{$\alpha$-SVM}
        \label{fig4c}
    \end{subfigure}
    \caption{Plots of expected testing error averaged over $100$ training sets for data generated from classes with a common $\bm{\Sigma}$. Here, $p=300$ and $n=100$.}
    \label{fig4}
\end{figure*}

The first set of class statistics we consider are \eqref{setting1}, \eqref{setting2}, and $\pi_0=\pi_1=0.5$. Corresponding to this data distribution are Figures \ref{fig2}, \ref{fig3}, and \ref{fig4}.


Figures \ref{fig2a} and \ref{fig2b} plot the average expected testing errors of $\alpha$-LDA and $\alpha$-SVM respectively against varying $\alpha$ when $p=400$ and $n=450$. At $\alpha=0.25$, the $\alpha$-LDA classifier achieves a $30.2\%$ relative decrease in the average expected testing error. Note that ordinary LDA ($\alpha=1$) is nowhere near optimal. On the other hand, $\alpha$-SVM achieves a $0.355\%$ decrease in average expected testing error at $\alpha=1.02$. These results suggest that there is a lot to be gained performance-wise by LDA in this regime but not so much by linear SVM. This can be attributed to the fact that LDA relies on sample estimation and that the noise due to estimation is high when $p=400$ and $n=450$. This is further supported by the results of Figures \ref{fig3a}, \ref{fig3b}  and \ref{fig3c}, which plot the average expected testing errors of $\alpha$-LDA, $\alpha$-SVM, and $\alpha$-log, respectively against varying $\alpha$ when $p=10$ and $n=500$. The minimum average expected occurs at exactly $\alpha=1$ for $\alpha$-LDA, $\alpha=1.01$ for $\alpha$-SVM and at $\alpha=0.99$ for $\alpha$-log, with the latter two classifiers achieving a relative decrease of no more than $1\%$ and $0.2\%$ respectively. The extreme behavior in all three figures can be explained by the fact that there is very little estimation noise for this choice of dimensions. What is notable is the difference between Figure \ref{fig2a} and Figure \ref{fig3a} whch suggests that the weight vector tuning method is most effective under high estimation noise and for methods which are most sensitive to it. This idea is again reinforced in Figures \ref{fig4a}, \ref{fig4b}, and \ref{fig4c}, in which the average expected testing errors of $\alpha$-RLDA, $\alpha$-RPLDA, and $\alpha$-SVM respectively are plotted against varying $\alpha$ when $p=300$ and $n=100$. The relative decrease in errors for each of the three classifiers does not exceed $1.3\%$. It must be that R-LDA and RP-LDA are able to reduce much of the estimation noise on their own, and so the $\alpha$ parameterization does not bring much improvement.

\begin{figure*}[t!]
    \centering
    \begin{subfigure}[t]{0.5\textwidth}
        \centering
        \includegraphics[width=0.9\linewidth]{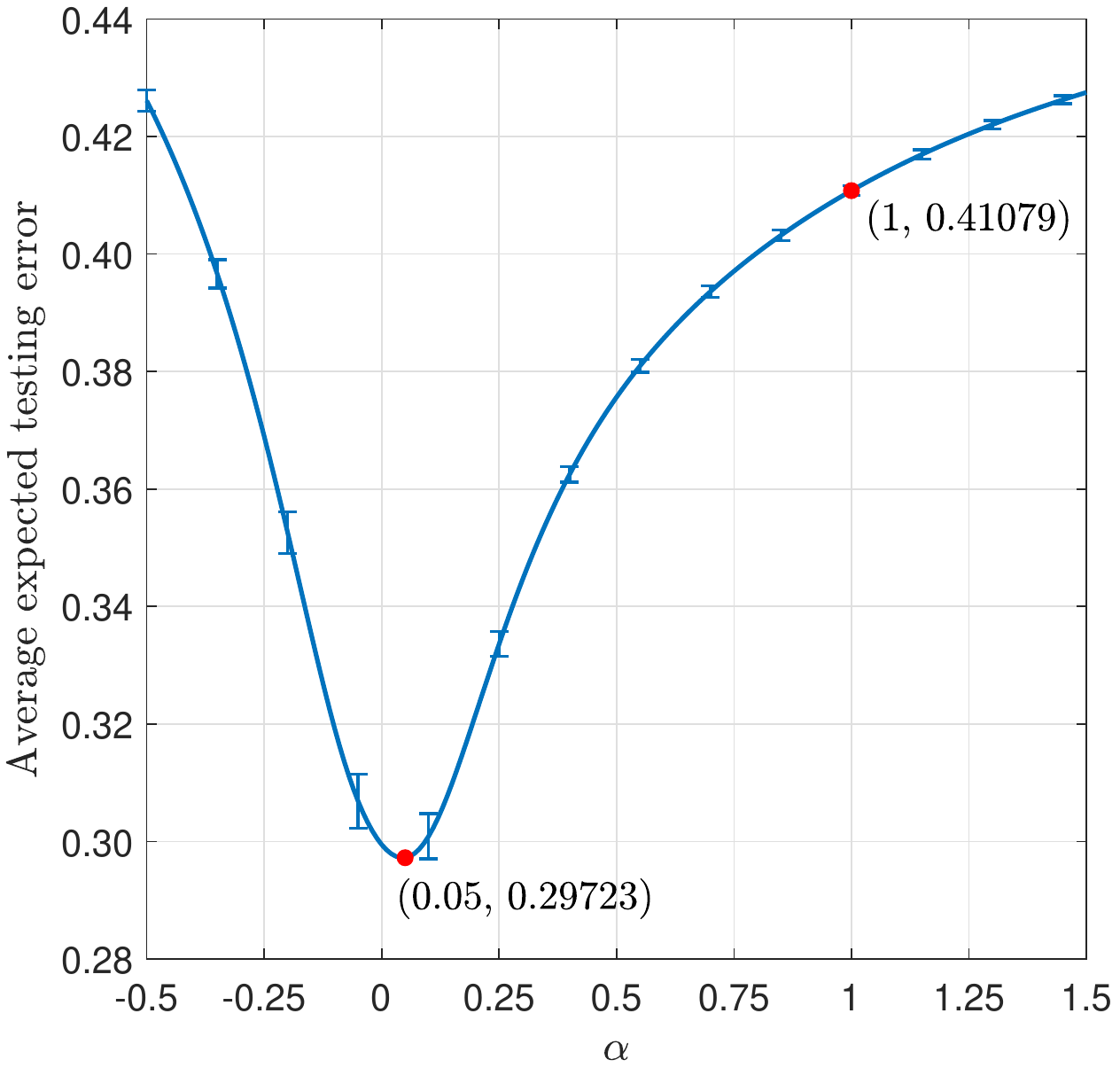}
        \caption{$\alpha$-LDA}
        \label{fig5a}
    \end{subfigure}%
    ~ 
    \begin{subfigure}[t]{0.5\textwidth}
        \centering
        \includegraphics[width=0.9\linewidth]{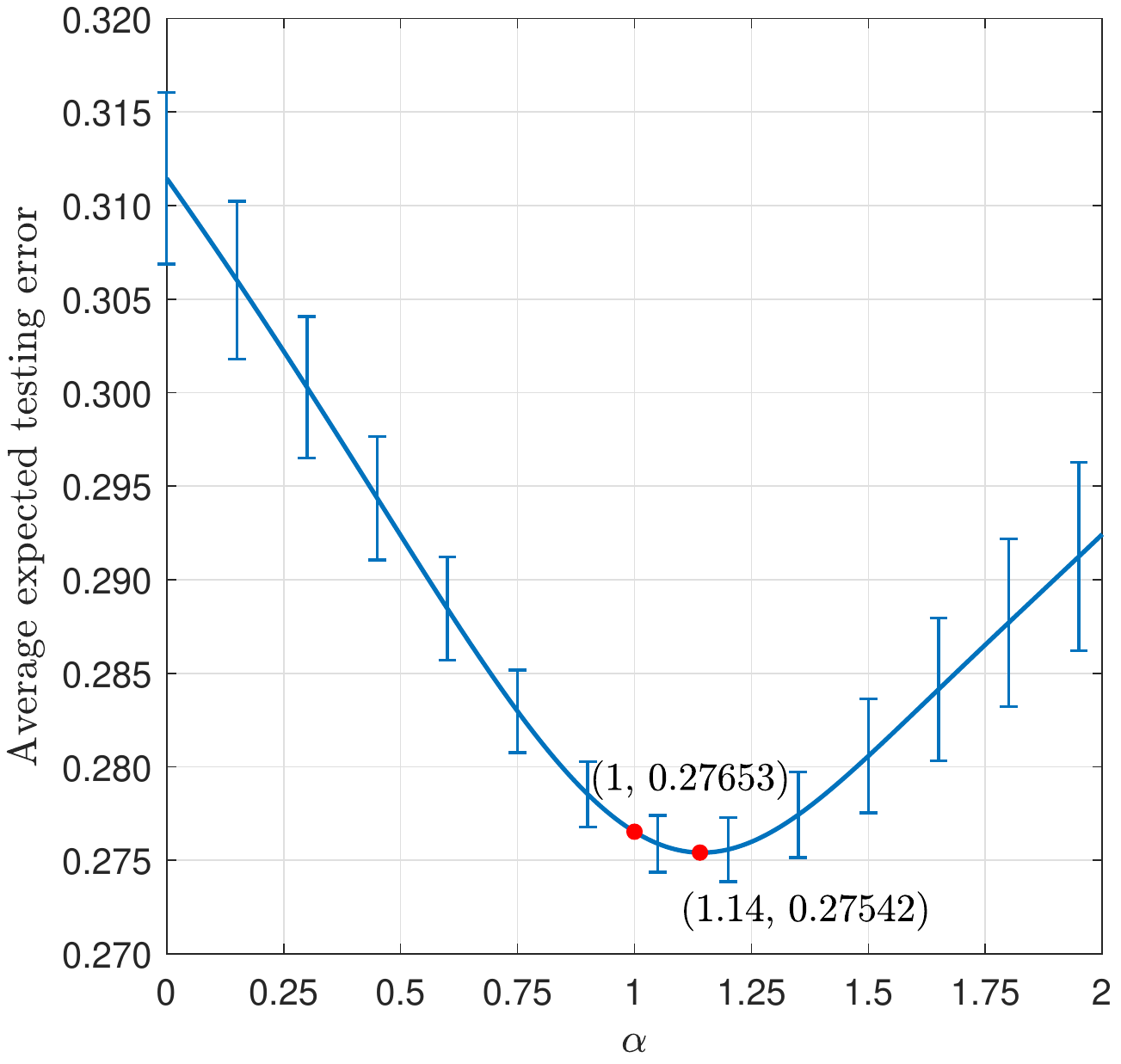}
        \caption{$\alpha$-SVM}
        \label{fig5b}
    \end{subfigure}
    \caption{Plots of expected testing error averaged over $100$ training sets for data generated from classes with distinct $\bm{\Sigma}_0$ and $\bm{\Sigma}_1$. Here, $p=400$ and $n=450$.}
    \label{fig5}
\end{figure*}

\begin{figure*}[t!]
    \centering
    \begin{subfigure}[t]{0.3\textwidth}
        \centering
        \includegraphics[width=1\linewidth]{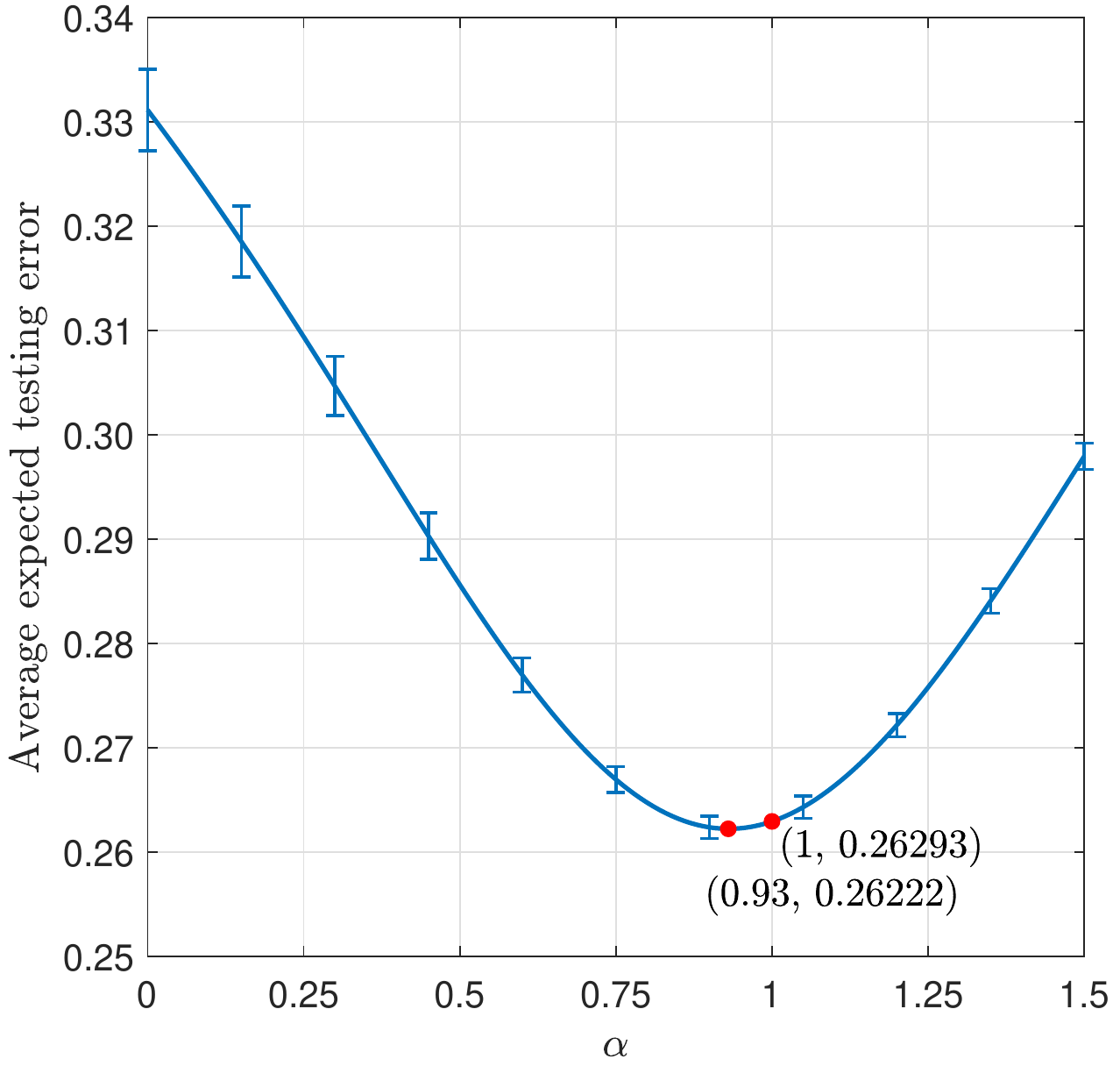}
        \caption{$\alpha$-RLDA}
        \label{fig6a}
    \end{subfigure}%
    ~ 
    \begin{subfigure}[t]{0.3\textwidth}
        \centering
        \includegraphics[width=1\linewidth]{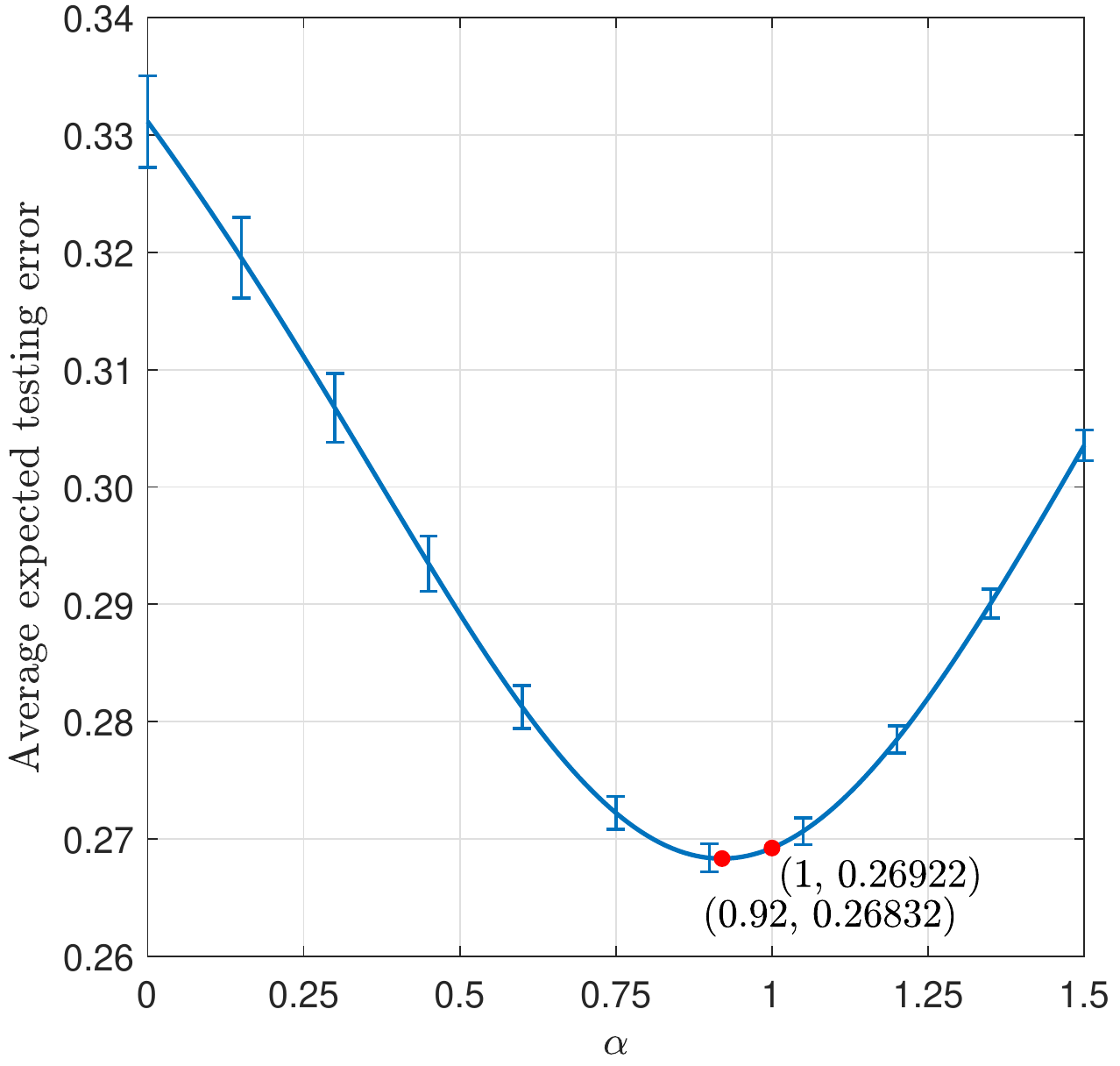}
        \caption{$\alpha$-RPLDA}
        \label{fig6b}
    \end{subfigure}
    ~
      \begin{subfigure}[t]{0.3\textwidth}
        \centering
        \includegraphics[width=1\linewidth]{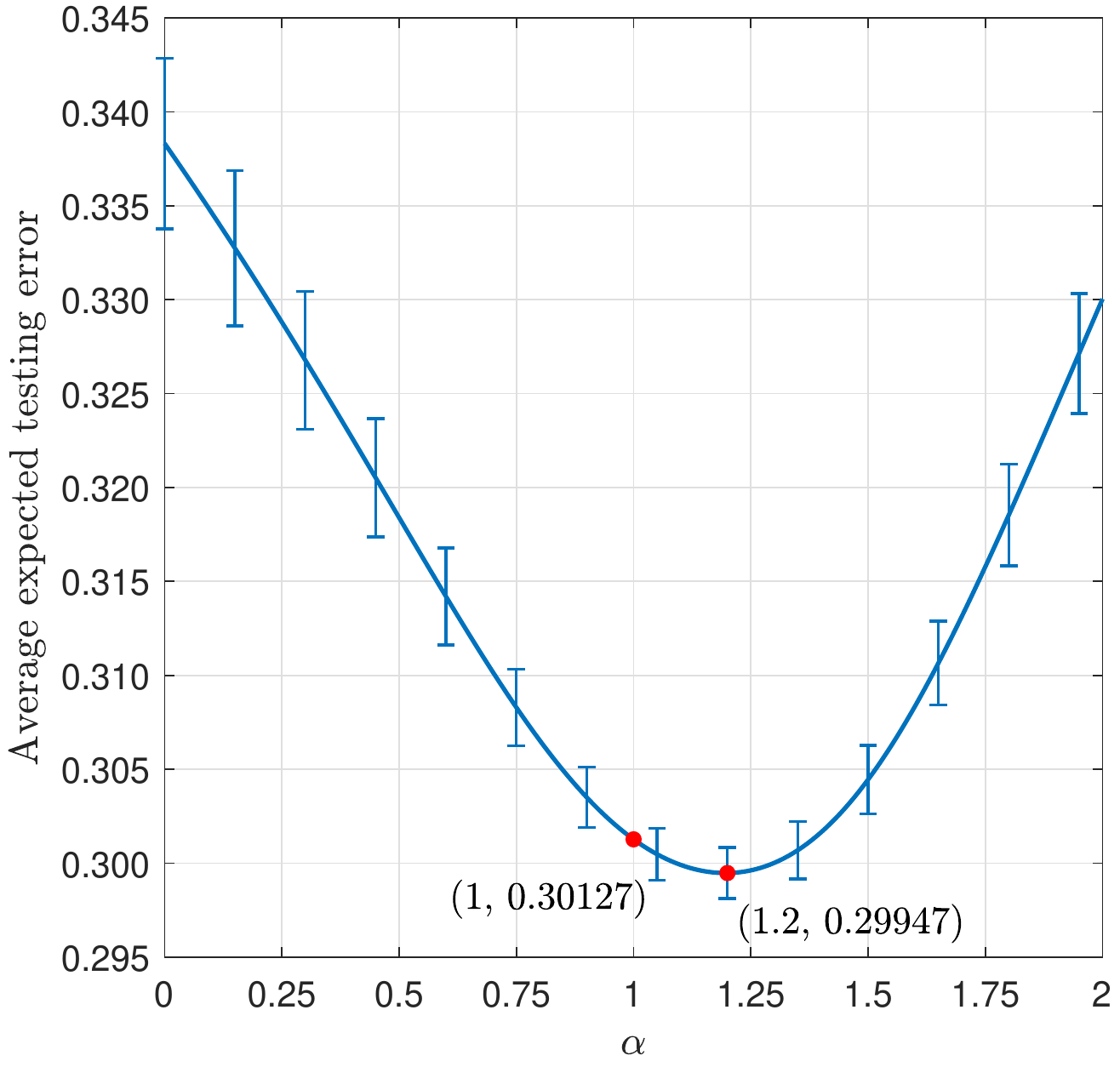}
        \caption{$\alpha$-SVM}
        \label{fig6c}
    \end{subfigure}
    \caption{Plots of expected testing error averaged over $100$ training sets for data generated from classes with distinct $\bm{\Sigma}_0$ and $\bm{\Sigma}_1$. Here, $p=300$ and $n=100$.}
    \label{fig6}
\end{figure*}
Figures \ref{fig5} and \ref{fig6} are based on data with the class statistics 
\begin{equation}
  \mu_0=\frac{1}{p^{1/4}}\left[\textbf{1}^T_{\lceil{\sqrt{p}}\rceil} \ \textbf{0}^T_{p-\lceil{\sqrt{p}}\rceil-2} \ 2 \ 2\right]^T, \ \mu_1=\textbf{0}_p, 
\end{equation}
\begin{equation}
    \left[\bm{\Sigma}_0\right]_{ij}=0.9^{|i-j|}, \  i,j=1,\ldots,p,
\end{equation}
\begin{equation}
    \bm{\Sigma}_1=\frac{10}{p}\textbf{1}_p\textbf{1}_p^T+0.1\textbf{I}_p,
\end{equation}
and $\pi_0=\pi_1=0.5$. The difference here is that the class covariances are distinct. Figures \ref{fig5a} and \ref{fig5b} again plot the average expected testing errors of $\alpha$-LDA and $\alpha$-SVM, respectively, against varying $\alpha$ when $p=400$ and $n=450$. In this case, $\alpha$-LDA significantly improves in performance when $\alpha$ is set to a non-unit value. It achieves a relative decrease in error of $27.6\%$ at $\alpha=0.05$, while $\alpha-$SVM achieves a relative decrease in error of $0.4\%$ at $\alpha=1.14$. Finally, Figures \ref{fig6a}, \ref{fig6b}, and \ref{fig6c} plot the average expected testing errors of $\alpha$-RLDA, $\alpha$-RPLDA and $\alpha$-SVM against varying $\alpha$ when $p=300$ and $n=100$. Here, the relative decreases in error do not exceed $0.6\%$.

\begin{figure}
    \centering
    \includegraphics[width=0.5\linewidth]{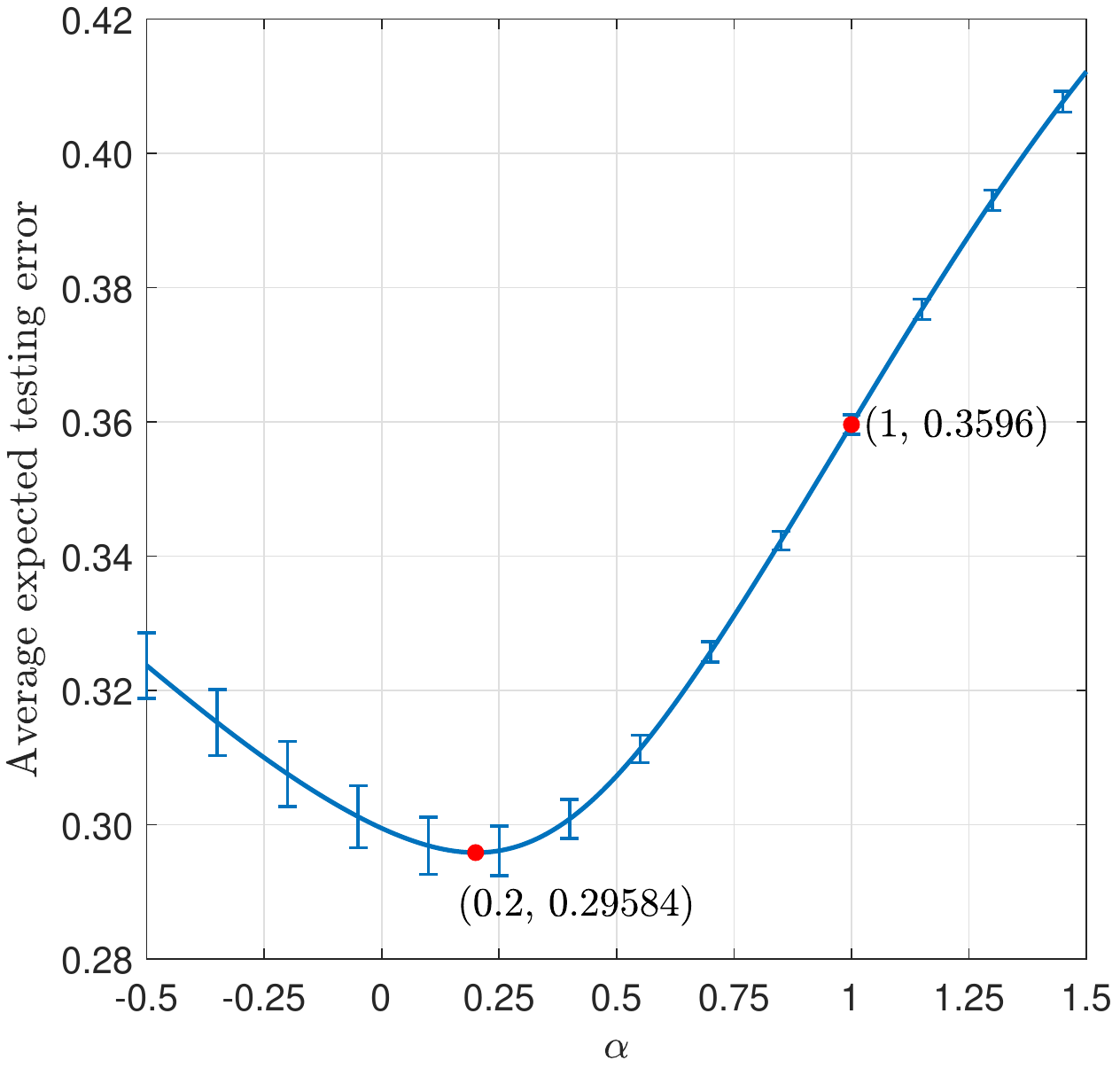}
    \caption{Plot of expected testing error of $\alpha$-SVM with penalty set to $1$ averaged over $100$ training sets for data generated from classes with distinct $\bm{\Sigma}_0$ and $\bm{\Sigma}_1$. Here, $p=400$ and $n=450$.}
    \label{bonus}
\end{figure}

As described at the beginning of this section, for each training set, the SVM penalty is tuned to the value yielding the lowest expected testing error. We found that SVM does not show much improvement when it is $\alpha$ parameterized. It is interesting to observe what happens when the penalty is not tuned beforehand. Instead we set the penalty to $1$ (its default setting in the MATLAB R2019b `fitcsvm' function) uniformly across all training sets. Figure \ref{bonus} shows the resulting average expected testing error of $\alpha$-SVM plotted against vary $\alpha$ in the same setting as in Figure \ref{fig5b}, i.e. $p=450$, $n=400$, and distinct $\bm{\Sigma}_0$ and $\bm{\Sigma}_1$.
In this case, $\alpha$-SVM achieves a relative decrease in error of $17.7\%$ at $\alpha=0.2$. Clearly, the method improves performance when $\textbf{w}$ itself is not at its optimal.

\begin{figure*}[t!]
    \centering
    \begin{subfigure}[t]{0.5\textwidth}
        \centering
        \includegraphics[width=0.9\linewidth]{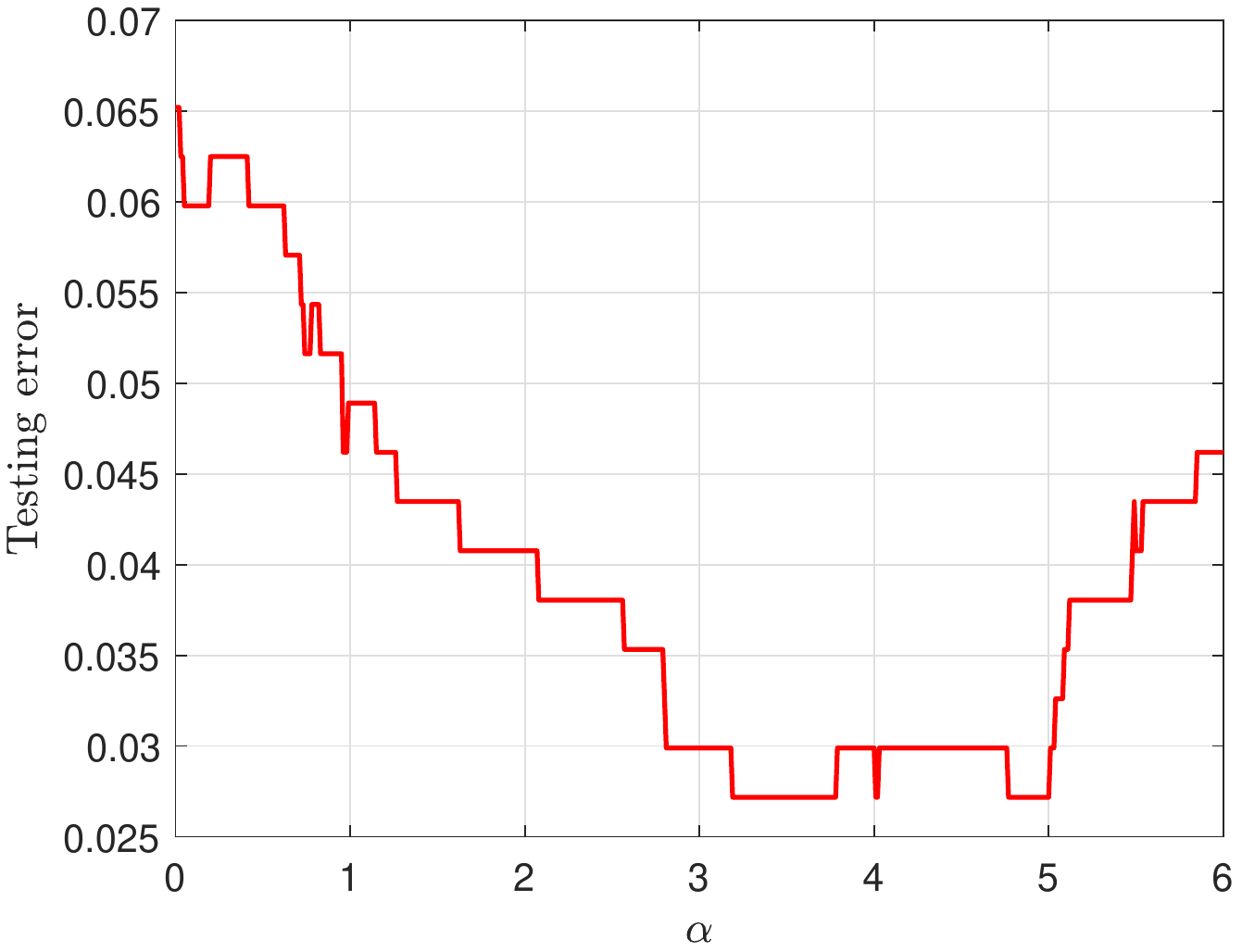}
        \caption{USPS 2 and 6}
        \label{extraSVM1}
    \end{subfigure}%
    ~ 
    \begin{subfigure}[t]{0.5\textwidth}
        \centering
        \includegraphics[width=0.9\linewidth]{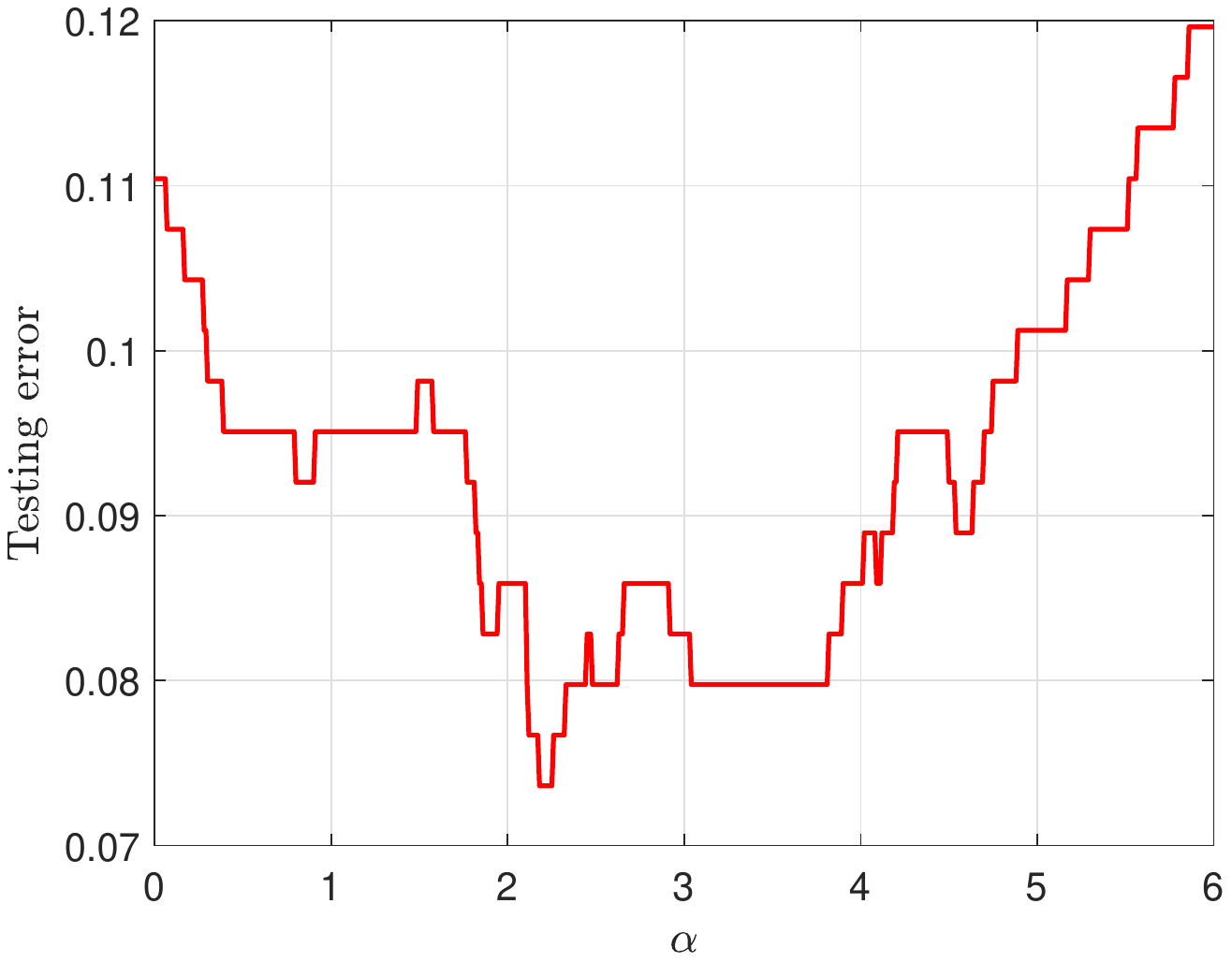}
        \caption{USPS 3 and 5}
        \label{extraSVM2}
    \end{subfigure}
    \caption{Plots of testing error on USPS digit pairs of $\alpha$-SVM with penalty set non-optimally}
    \label{extraSVM}
\end{figure*}

Taking this idea further, we show that tuning the weight vector of a SVM classifier with a poorly chosen penalty can compensate for the resulting loss in performance. Figure \ref{extraSVM} is based on the USPS dataset consisting of separate training and testing sets of grayscale images of handwritten digits $0-9$. Pairs of digits are used to form a binary classification problem. For each pair of digits, a poorly tuned SVM classifier is $\alpha$ parameterized and the testing error plotted against $\alpha$ to illustrate the effect of weight vector tuning.

For the digit pair `2' and `6', an optimized SVM classifier can achieve a testing error of $0.0217$. Figure \ref{extraSVM1} shows the testing error of $\alpha$-SVM starting with a poorly tuned SVM classifer whose testing error on this digit pair is $0.0489$. By weight vector tuning, the testing error can be brought down to $0.0272$. This is comparable to the performance of the original optimized SVM classifer. Similarly, for the digit pair `3' and `5', an optimized SVM classifier can achieve a testing error of $0.0675$. Figure \ref{extraSVM1} shows the testing error of $\alpha$-SVM starting with a poorly tuned SVM classifer whose testing error on this digit pair is $0.0951$. By weight vector tuning, the testing error can be brought down to $0.0736$.

The significance of this finding is the potential savings in computation that can be made by weight vector tuning versus penalty tuning. The reason for this is that weight vector tuning is an afterthought; it occurs post weight vector generation. On the other hand, setting the penalty is done prior to weight vector generation. An optimization problem must be solved to generate the weight vector with each setting of the penalty. 

This idea generalizes to any linear classifier whose native hyperparameters are set prior to weight vector generation. The tuning of the hyperparameters will then involve repeatedly generating the weight vector. If this process is costly, weight vector tuning can provide a more computationally efficient method of improving performance than tuning the native hyperparameters. Another example that is not demonstrated here is the RP-LDA ensemble classifier whose projection dimension $d$ is a native hyperparameter. Tuning this is computationally inefficient as it means projecting all the data with each setting of $d$. A simple alternative is weight vector tuning.

Overall, we conclude from this section that $\alpha$-LDA in the `$n$ on the order of $p$' scenario shows the most promise in terms of improved performance. For this reason, we proceed to study this classifier in the RMT asymptotic regime in the next section.

\section{Asymptotic Analysis of the Parameterized LDA Classifier}
%need to put what it means, need to pput motivating simulations for the asymptotic regime
In this section, we extend our study of $\alpha$-LDA, the modified weight discriminant \eqref{mod} corresponding to the plugin LDA weight vector. The $\alpha$-LDA discriminant
  \begin{equation}
     \frac{\hat{\bm{\mu}}^T\hat{\bm{\Sigma}}^{-1}\hat{\bm{\mu}}}{\hat{\bm{\mu}}^T\hat{\bm{\mu}}}\hat{\bm{\mu}}^T\hat{\tilde{\bm{x}}}+\alpha\hat{\bm{\mu}}^T\hat{\bm{\Sigma}}^{-1}\textbf{P}_{\hat{\bm{\mu}}}\hat{\tilde{\bm{x}}}\label{alphaLDA}
\end{equation}
is a bridging between LDA (when $\alpha=1$) and the nearest centroid classifier (when $\alpha=0$) with decision rule
  \begin{align}
    \mathbbm{1}\left\{||\hat{\bm{\mu}}_0-\textbf{x}||_2^2-||\hat{\bm{\mu}}_1-\textbf{x}||_2^2>0\right\}=\mathbbm{1}\left\{\hat{\bm{\mu}}^T\hat{\tilde{\bm{x}}}>0\right\}.\label{sampleCentroid}
\end{align}
As suggested by the name, the nearest centroid classifier classifies $\textbf{x}$ to the class with nearest sample mean. It is the Bayes classifier for data distributed as \eqref{dist} when $\bm{\Sigma}=\textbf{I}_p$.

As the previous section has shown, $\alpha$-LDA exhibits the greatest improvement in performance among the sampled classifiers, particularly when the data dimensionality $p$ is on the order of the number of samples $n$. This can be attributed to the fact that the LDA weight vector is an explicit function of the sample statistics. Due to estimation noise, there is much to be gained in this regime. We thus pursue an asymptotic study of $\alpha$-LDA in growth regime where $n$ and $p$ grow at constant rates to each other. Under this growth regime, we derive an asymptotic expression and an estimator for the probability of misclassification of $\alpha$-LDA. 

\subsection{Asymptotic Analysis}
In this section we first show that under the following growth regime assumptions
\begin{enumerate}[{\hspace{20px}(a)}]
    \item $0<\liminf\frac{p}{n}<\limsup\frac{p}{n}<1$
    \item $\frac{n_i}{n}\rightarrow c_i\in(0,1), \ i=0,1$
    \item  $\limsup\limits_{p} \Vert\bm{\mu}_0-\bm{\mu}_1\Vert_2<\infty$
    % \alert{\item  $\liminf\limits_{p} \Vert\bm{\mu}_0-\bm{\mu}_1\Vert_2>0$}
    \item  $\limsup\limits_{p} \Vert\bm{\Sigma}_i\Vert_2<\infty, \ i=0,1$
    \item $\liminf\limits_{p} \lambda_{\text{min}}\left(\bm{\Sigma}_i\right)>0, \ i=0,1$
\end{enumerate}
and considering the training set to be random, the probability of misclassification of the $\alpha$-LDA classifier converges to a quantity that is a function of only true statistics.
 This quantity is referred to as the \textit{deterministic equivalent} of the probability of misclassification. The deterministic equivalent approximates the random realization of the probability of misclassification, and can be useful for understanding the behavior of the classifier with synthetic data, for which the statistics are perfectly known. In practice, however, the statistics are unknown. For this reason, we also derive an estimator of the probability of misclassification which is consistent under the same growth assumptions. This is referred to as a \textit{G-estimator} of the probability of misclassification and can be used to tune $\alpha$. To proceed with these derivations, we first require an expression for the expected probability of misclassification.

Assuming the classes $\mathcal{C}_0$ and $\mathcal{C}_1$ are Gaussian with means and covariances $\bm{\mu}_0$, $\bm{\Sigma}_0$ and $\bm{\mu}_1$, $\bm{\Sigma}_1$ respectively, the probability of misclassification of a test point $\textbf{x}$ by the $\alpha$-LDA classifier has the form 
\begin{equation}
\varepsilon=\pi_0\Phi\left(\frac{m_0}{\sqrt{\sigma_0^2}}\right)+\pi_1\Phi\left(-\frac{m_1}{\sqrt{\sigma_1^2}}\right)
\end{equation}
where $m_0$, $m_1$, $\sigma_0^2$, and $\sigma_1^2$ are the discriminant means and variances conditioned on $\textbf{x}\in\mathcal{C}_0$ and $\textbf{x}\in\mathcal{C}_1$ respectively. Define $\rho=\frac{\hat{\bm{\mu}}^T\hat{\bm{\Sigma}}^{-1}\hat{\bm{\mu}}}{\hat{\bm{\mu}}^T\hat{\bm{\mu}}}$. Then   
\begin{align}
    m_i&=\left(\rho\hat{\bm{\mu}}^T+\alpha\hat{\bm{\mu}}^T\hat{\bm{\Sigma}}^{-1}\textbf{P}_{\hat{\bm{\mu}}}\right)\left(\bm{\mu}_i-\frac{\hat{\bm{\mu}}_0+\hat{\bm{\mu}}_1}{2}\right), \ i=0,1,
\end{align}
and
\begin{align}
     \sigma^2_i&=\left(\rho\hat{\bm{\mu}}^T+\alpha\hat{\bm{\mu}}^T\hat{\bm{\Sigma}}^{-1}\textbf{P}_{\hat{\bm{\mu}}}\right)\bm{\Sigma}_i\left(\rho\hat{\bm{\mu}}^T+\alpha\hat{\bm{\mu}}^T\hat{\bm{\Sigma}}^{-1}\textbf{P}_{\hat{\bm{\mu}}}\right)^T, \ i=0,1
\end{align}

In the following sections, we present the DEs and G-estimators for both the general case of distinct covariances and the special case of common covariances.
\subsubsection{Deterministic Equivalent of the Probability of Misclassification}
Formally, the deterministic equivalent of $\varepsilon$, denoted by $\bar{\varepsilon}$, is a sequence of $p$ and $n$ satisfying
\begin{equation}
    \varepsilon-\bar{\varepsilon}\xrightarrow[]{\text{a.s.}}0\label{error}
\end{equation}
under the growth regime assumptions (a)-(f). For sequences $\bar{m}_0$, $\bar{m}_1$, $\bar{\sigma}_0^2$, and $\bar{\sigma}_1^2$ such that
\begin{align}
   &{m}_i-\bar{m}_i\xrightarrow[]{\text{a.s.}}0, \ i=0,1\\
   &{\sigma}_i^2-\bar{\sigma}_i^2\xrightarrow[]{\text{a.s.}}0, \ i=0,1
   \label{all}
\end{align}
under the growth regime assumptions (a)-(e), it is 
\begin{equation}
\bar{\varepsilon}=\pi_0\Phi\left(\frac{\bar{m}_0}{\sqrt{\bar{\sigma}_0^2}}\right)+\pi_1\Phi\left(-\frac{\bar{m}_1}{\sqrt{\bar{\sigma}_1^2}}\right)
\end{equation}
(see Lemma 2 in \cite{niyazi2020asymptotic} for proof).
Thus, the deterministic equivalent $\bar{\varepsilon}$ is itself a function of deterministic equivalents $\bar{m}_0$, $\bar{m}_1$, $\bar{\sigma}_0^2$, and $\bar{\sigma}_1^2$ which are also functions of only true statistics.

In the following theorem, we state the expressions of $\bar{m}_0$, $\bar{m}_1$, $\bar{\sigma}_0^2$, and $\bar{\sigma}_1^2$ which are used to compute $\bar{\varepsilon}$. This is followed by a corollary which corresponds to the special case when $\bm{\Sigma}_0=\bm{\Sigma}_1=\bm{\Sigma}$.\footnote{Note in these statements that while technically $n-2$ is equivalent to $n$ asymptotically, we retain the $n-2$ in these expressions for increased accuracy in finite dimensions.} First, define
\begin{align}
    \bar{\textbf{Q}}&=\left(\frac{n_0-1}{n-2}\frac{1}{1+\tilde{\delta}}\bm{\Sigma}_0+\frac{n_1-1}{n-2}\frac{1}{1+\tilde{\nu}}\bm{\Sigma}_1\right)^{-1},
\end{align}

\begin{equation}
    R_{ij}=\frac{n_{i-1}-1}{n_{j-1}-1}\left[\left(\textbf{I}_2-\bm{\Omega}\right)^{-1}\bm{\Omega}\right]_{i,j}, \ i,j=1,2,
\end{equation}
\begin{align}
    \left[\bm{\Omega}\right]_{1j}&=\frac{n_{j-1}-1}{n-2}\left(\frac{1}{1+\tilde{\delta}}\right)^2\frac{1}{n-2}\text{tr}\left\{\bm{\Sigma}_0\bar{\textbf{Q}}\bm{\Sigma}_{j-1}\bar{\textbf{Q}}\right\}, \ j=1,2,\\
    \left[\bm{\Omega}\right]_{2j}&=\frac{n_{j-1}-1}{n-2}\left(\frac{1}{1+\tilde{\nu}}\right)^2\frac{1}{n-2}\text{tr}\left\{\bm{\Sigma}_1\bar{\textbf{Q}}\bm{\Sigma}_{j-1}\bar{\textbf{Q}}\right\},\ j=1,2,
\end{align}
\begin{equation}
  \textbf{A}_i=\bm{\Sigma}_i\bar{\textbf{Q}}, \ i=0,1,
\end{equation}
\begin{equation}
\tilde{\textbf{Q}}_i=\bar{\textbf{Q}}\left(\textbf{A}_i+R_{1(i+1)}\textbf{A}_0+R_{2(i+1)}\textbf{A}_1\right), \ i=0,1,
\end{equation}
\begin{equation}
   \kappa=\frac{{\bm{\mu}}^T\bar{\textbf{Q}}{\bm{\mu}}+\frac{1}{n_0}\text{tr}\left\{\textbf{A}_0\right\}+\frac{1}{n_1}\text{tr}\left\{\textbf{A}_1\right\}}{\bm{\mu}^T\bm{\mu}+\frac{1}{n_0}\text{tr}\left\{\bm{\Sigma}_0\right\}+\frac{1}{n_1}\text{tr}\left\{\bm{\Sigma}_1\right\}}, 
\end{equation}

\begin{equation}
\eta=\frac{    \left(\frac{1}{1-\frac{p}{n-2}}\right)\left[\bm{\mu}^T\bm{\Sigma}^{-1}\bm{\mu}+\frac{p}{n_0}+\frac{p}{n_1}\right]}{   \bm{\mu}^T\bm{\mu}+\left(\frac{1}{n_0}+\frac{1}{n_1}\right)\text{tr}\left\{\bm{\Sigma}\right\}},
\end{equation}
\begin{equation}
    \tau=\frac{1}{1-\frac{p}{n-2}},
\end{equation}
 and $\tilde{\delta}$ and $\tilde{\nu}$ are the results of the fixed point iteration
\begin{align}
    \tilde{\delta}^{(k)}&=\frac{1}{n-2}\text{tr}\left\{\bm{\Sigma}_0\left(\frac{n_0-1}{n-2}\frac{1}{1+\tilde{\delta}^{(k-1)}}\bm{\Sigma}_0+\frac{n_1-1}{n-2}\frac{1}{1+\tilde{\nu}^{(k-1)}}\bm{\Sigma}_1\right)^{-1}\right\}\\
       \tilde{\nu}^{(k)}&=\frac{1}{n-2}\text{tr}\left\{\bm{\Sigma}_1\left(\frac{n_0-1}{n-2}\frac{1}{1+\tilde{\delta}^{(k-1)}}\bm{\Sigma}_0+\frac{n_1-1}{n-2}\frac{1}{1+\tilde{\nu}^{(k-1)}}\bm{\Sigma}_1\right)^{-1}\right\}, \ k=1,2,3,\ldots\label{fixedPOINT}
\end{align}
for any positive initialization $\tilde{\delta}^{(0)}$ and $\tilde{\nu}^{(0)}$.

\textbf{Theorem 2}\textit{ (Distinct covariance DEs)
The deterministic equivalents $\bar{m}_0$, $\bar{m}_1$, $\bar{\sigma}_0^2$, and $\bar{\sigma}_1^2$, satisfying \eqref{all} under the growth regime assumptions (a)-(e) are given by} 
\begin{align}
   {\bar{m}_i }&=(1-\alpha)\kappa\left[\frac{(-1)^{i+1}}{2}\bm{\mu}^T\bm{\mu}+\frac{1}{2}\left(\frac{1}{n_0}\text{tr}\left\{\bm{\Sigma}_0\right\}-\frac{1}{n_1}\text{tr}\left\{\bm{\Sigma}_1    \right\}\right)\right]\\
   &\hspace{50px}+\alpha \left[\frac{(-1)^{i+1}}{2}{\bm{\mu}}^T\bar{\textbf{Q}}{\bm{\mu}}+\frac{1}{2}\left(\frac{1}{n_0}\text{tr}\left\{\textbf{A}_0\right\}-\frac{1}{n_1}\text{tr}\left\{\textbf{A}_1\right\}\right)\right], \ i=0,1
\end{align}
and
\begin{align}
   \bar{\sigma}_i^2&=(1-\alpha)^2\kappa^2\left[\bm{\mu}^T\bm{\Sigma}_i\bm{\mu}+\frac{1}{n_0}\text{tr}\left\{\bm{\Sigma}_0\bm{\Sigma}_i\right\}+\frac{1}{n_1}\text{tr}\left\{\bm{\Sigma}_1\bm{\Sigma}_i\right\}\right]\\
   &\hspace{20px}+2\alpha(1-\alpha)\kappa\left[\bm{\mu}^T\textbf{A}_i\bm{\mu}+\frac{1}{n_0}\text{tr}\left\{\bm{\Sigma}_i\textbf{A}_0\right\}+\frac{1}{n_1}\text{tr}\left\{\bm{\Sigma}_i\textbf{A}_1\right\}\right]\\
   &\hspace{20px}+\alpha^2\left[\bm{\mu}^T\tilde{\textbf{Q}}_i\bm{\mu}+\frac{1}{n_0}\text{tr}\left\{\bm{\Sigma}_0\tilde{\textbf{Q}}_i\right\}+\frac{1}{n_1}\text{tr}\left\{\bm{\Sigma}_1\tilde{\textbf{Q}}_i\right\}\right], \ i=0,1.
\end{align}

\textbf{Proof:} See Appendix \ref{DE_distinct}.

\textbf{Corollary 2}\textit{ (Common covariance DEs) The deterministic equivalents $\bar{m}_0$, $\bar{m}_1$, $\bar{\sigma}_0^2$, and $\bar{\sigma}_1^2$ satisfying \eqref{all} under the growth regime assumptions (a)-(e) are given by} 

\begin{align}
   {\bar{m}_i }&{=(1-\alpha)\eta\left(\frac{(-1)^{i+1}}{2}\bm{\mu}^T\bm{\mu}+\frac{1}{2}\left(\frac{1}{n_0}-\frac{1}{n_1}\right)\text{tr}\left\{\bm{\Sigma}\right\}\right)}\\
    & {\hspace{10px}+\alpha\left[\frac{\tau}{2}\left[(-1)^{i+1}\bm{\mu}^T\bm{\Sigma}^{-1}\bm{\mu}+\frac{p}{n_0}-\frac{p}{n_1}\right]\right]}, \ i=0,1 
\end{align}
and
\begin{align}
   \bar{\sigma}_0^2&=\bar{\sigma}_1^2=(1-\alpha)^2\eta^2\left[\bm{\mu}^T\bm{\Sigma}\bm{\mu}+\left(\frac{1}{n_0}+\frac{1}{n_1}\right)\text{tr}\left\{\bm{\Sigma}^2\right\}\right]\\
    & \hspace{10px}+\alpha^2\tau^3\left[\bm{\mu}^T\bm{\Sigma}^{-1}\bm{\mu}+\frac{p}{n_0}+\frac{p}{n_1}\right]+2\alpha(1-\alpha)\tau\eta\left[\bm{\mu}^T\bm{\mu}+\left(\frac{1}{n_0}+\frac{1}{n_1}\right)\text{tr}\left\{\bm{\Sigma}\right\}\right]
\end{align}

\textbf{Proof:} See Appendix \ref{DE_common}.

\subsubsection{G-estimator of the Probability of Misclassification}
The G-estimator $\hat{\varepsilon}$ of the probability of misclassification $\varepsilon$ is a function of sample statistics $\hat{\bm{\mu}}_0$, $\hat{\bm{\mu}}_1$, $\hat{\bm{\Sigma}}_0$, and $\hat{\bm{\Sigma}}_1$ such that 
\begin{equation}
    \hat{\varepsilon}-\varepsilon\xrightarrow[]{\text{a.s.}}0\label{error2}
\end{equation}
under the growth regime assumptions (a)-(f). For sequences $\hat{m}_0$, $\hat{m}_1$, $\hat{\sigma}_0^2$, and $\hat{\sigma}_1^2$, which are also functions of only sample statistics, such that
\begin{align}
   &\hat{m}_i-{m}_i\xrightarrow[]{\text{a.s.}}, \ i=0,1,\\
   &\hat{\sigma}_i^2-{\sigma}i^2\xrightarrow[]{\text{a.s.}}0, \ i=0,1\label{all2}
\end{align}
under the growth regime assumptions (a)-(e), it is 
\begin{equation}
\hat{\varepsilon}=\hat{\pi}_0\Phi\left(\frac{\hat{m}_0}{\sqrt{\hat{\sigma}_0^2}}\right)+\hat{\pi}_1\Phi\left(-\frac{\hat{m}_1}{\sqrt{\hat{\sigma}_1^2}}\right).
\end{equation}

The following theorem states the expressions of $\hat{m}_0$, $\hat{m}_1$, $\hat{\sigma}_0^2$, and $\hat{\sigma}_1^2$ which are used to compute $\hat{\varepsilon}$. This is followed by a corollary which is specific to the case when $\bm{\Sigma}_0=\bm{\Sigma}_1=\bm{\Sigma}$ is assumed. First, define
\begin{equation}
    \lambda_i=\frac{\frac{1}{n-2}\text{tr}\left\{\hat{\bm{\Sigma}}_i\hat{\bm{\Sigma}}^{-1}\right\}}{1-\frac{1}{n-2}\text{tr}\left\{\hat{\bm{\Sigma}}_i\hat{\bm{\Sigma}}^{-1}\right\}}.
\end{equation}

\textbf{Theorem 3}\textit{ (Distinct covariance G-estimators) The G-estimators $\hat{m}_0$, $\hat{m}_1$, $\hat{\sigma}_0^2$, and $\hat{\sigma}_1^2$, satisfying \eqref{all2} under the growth regime assumptions (a)-(e) are given by} 
\begin{align}
   \hat{m}_i&=(-1)^{i+1}\left[(1-\alpha)\rho\left(\frac{1}{2} \hat{\bm{\mu}}^T\hat{\bm{\mu}}-\frac{1}{n_i}\text{tr}\left\{\hat{\bm{\Sigma}}_i\right\}\right)+\alpha\left(\hat{\bm{\mu}}^T\hat{\bm{\Sigma}}^{-1}\hat{\bm{\mu}}-\frac{n-2}{n_i}\lambda_i\right)\right], \ i=0,1
\end{align}
\textit{and}
\begin{align}
    \hat{\sigma}_i^2 &=(1-\alpha)^2\rho^2\hat{\bm{\mu}}^T\hat{\bm{\Sigma}}_i\hat{\bm{\mu}}+2\alpha(1-\alpha)\rho \left(1+\lambda_i\right)\hat{\bm{\mu}}^T\hat{\bm{\Sigma}}_i\hat{\bm{\Sigma}}^{-1}\hat{\bm{\mu}}\\
    &\hspace{50px}+\alpha^2 \left(1+\lambda_i\right)^2\hat{\bm{\mu}}^T\hat{\bm{\Sigma}}^{-1}\hat{\bm{\Sigma}}_i\hat{\bm{\Sigma}}^{-1}\hat{\bm{\mu}}, \ i=0,1
\end{align}

\textbf{Proof:} See Appendix \ref{GE_distinct}.

\textbf{Corollary 3}\textit{ (Common covariance G-estimators) The G-estimators $\hat{m}_0$, $\hat{m}_1$, $\hat{\sigma}_0^2$, and $\hat{\sigma}_1^2$, satisfying \eqref{all2} under the growth regime assumptions (a)-(e) are given by} 

\begin{align}
   \hat{m}_i&=\frac{(-1)^{i+1}}{2}\left(\rho\hat{\bm{\mu}}^T+\alpha\hat{\bm{\mu}}^T\hat{\bm{\Sigma}}^{-1}\textbf{P}_{\hat{\bm{\mu}}}\right)\hat{\bm{\mu}}\\
    & \hspace{10px}+(-1)^{i+1}\left[\rho(\alpha-1) \frac{1}{n_i}\text{tr}\left\{\hat{\bm{\Sigma}}\right\}-\alpha\frac{\frac{p}{n_i}}{1-\frac{p}{n-2}}\right], \ i=0,1
\end{align}
\textit{and}
\begin{align}
    \hat{\sigma}_0^2=\hat{\sigma}_1^2 &=\rho^2(1-\alpha)^2\hat{\bm{\mu}}^T{\hat{\bm{\Sigma}}}\hat{\bm{\mu}}+\alpha^2  \tau^2 \hat{\bm{\mu}}^T\hat{\bm{\Sigma}}^{-1}\hat{\bm{\mu}}+2\alpha\rho(1-\alpha) \tau\hat{\bm{\mu}}^T\hat{\bm{\mu}}
\end{align}

\textbf{Proof:} See Appendix \ref{GE_common}.

Notice that $\hat{\varepsilon}$ is a function of the sample statistics. It estimates the probability of misclassification without the need for additional testing data and it is much more computationally efficient than the cross-validation procedure. In the next section, we show how to use $\hat{\varepsilon}$ for the purpose of tuning the $\alpha$ parameter.

\subsection{Tuning the $\alpha$-LDA Parameter}
In this section, $\alpha$-LDA is applied to real data. The objective is to show how $\alpha$-LDA performs as compared to LDA and the nearest centroid classifier on real data, as well as to demonstrate the use of the G-estimator $\hat{\varepsilon}$ in tuning the $\alpha$ parameter. We consider binary classification of digit pairs from the USPS dataset \cite{le1990handwritten} and phoneme pairs from the dataset \cite{hastie1995penalized}. For each problem, we train and test LDA, nearest centroid, and $\alpha$-LDA on the relevant dataset. The empirical errors are plotted against varying $\alpha$. Also plotted is the G-estimator $\hat{\varepsilon}$ of the error of $\alpha$-LDA. \footnote{Note that for these particular datasets, the two G-estimators almost match. Out of the two, the G-estimator which assumes common covariances is plotted.}

\begin{figure*}[t!]
    \centering
    \begin{subfigure}[t]{0.5\textwidth}
        \centering
        \includegraphics[width=0.9\linewidth]{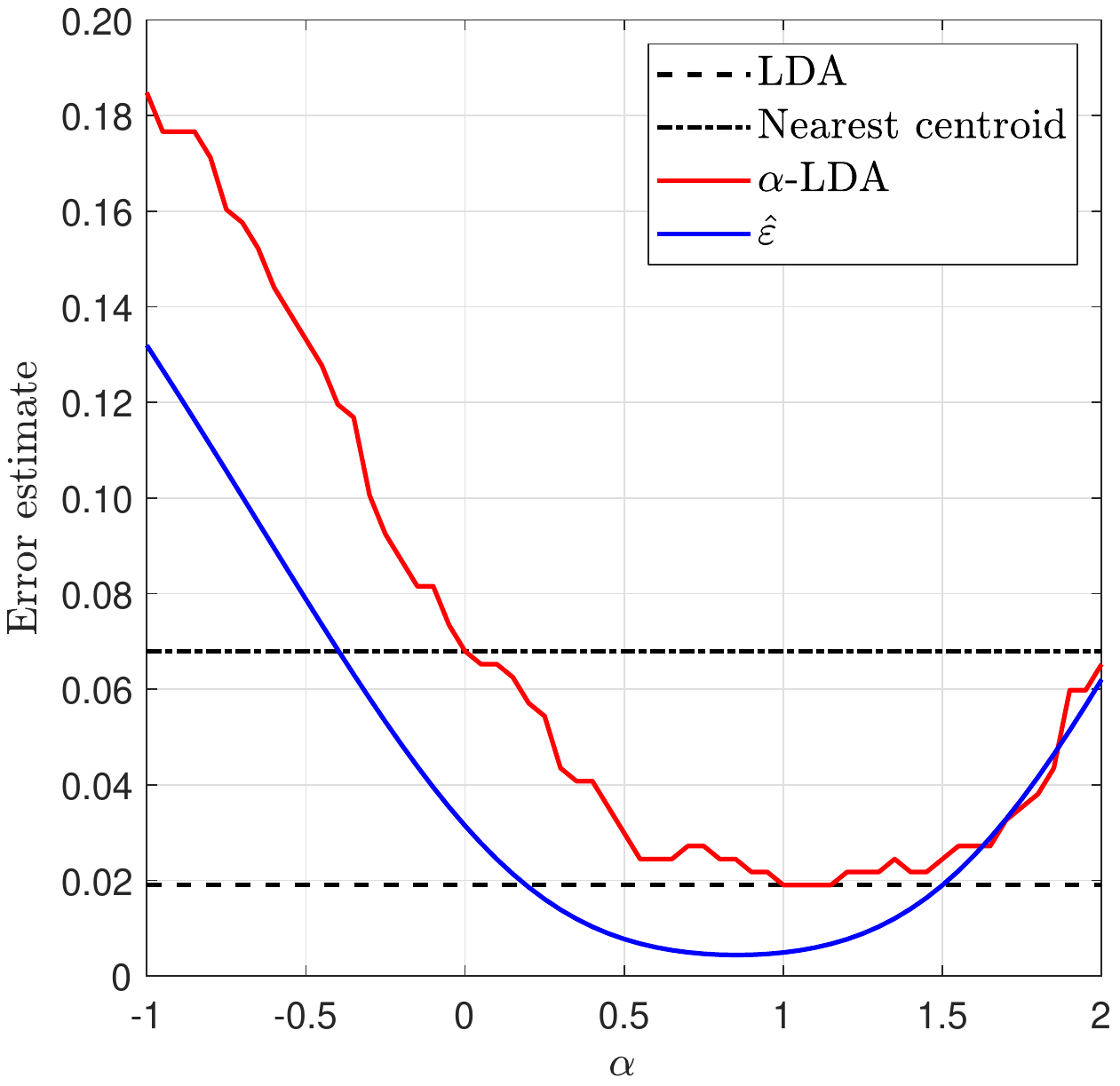}
        \caption{USPS 2 and 6}
        \label{fig7a}
    \end{subfigure}%
    ~ 
    \begin{subfigure}[t]{0.5\textwidth}
        \centering
        \includegraphics[width=0.9\linewidth]{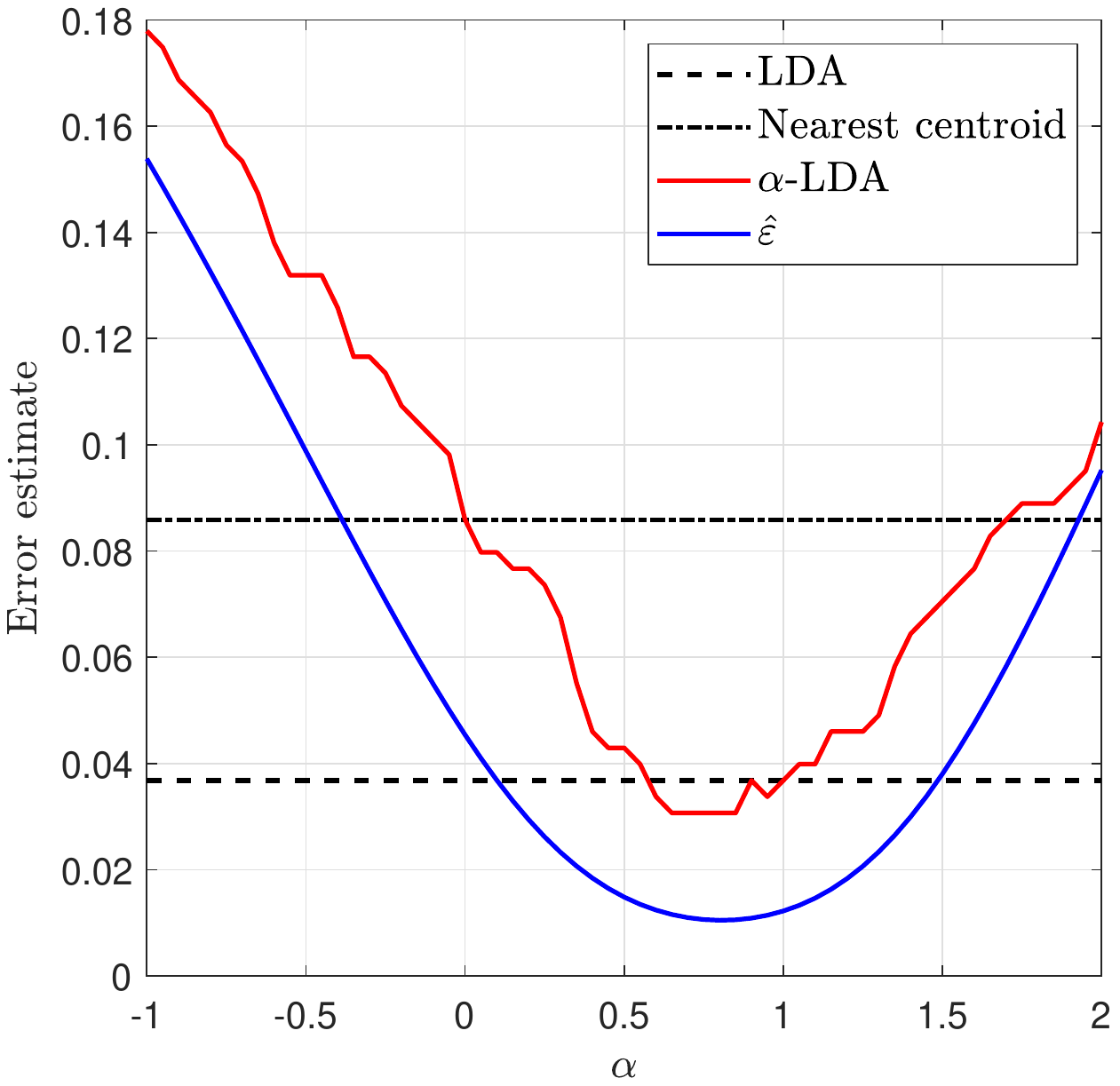}
        \caption{USPS 5 and 8}
        \label{fig7b}
    \end{subfigure}
    \caption{Plots of testing error estimates of classifying USPS digit pairs for LDA, the nearest centroid, and $\alpha$-LDA as well as the G-estimator $\hat{\varepsilon}$ of the $\alpha$-LDA expected testing error.}
    \label{fig7}
\end{figure*}

Figure \ref{fig7} shows the results on two digit pairs from the USPS dataset. As mentioned in Section \ref{simSample}, this dataset consists of grayscale images of handwritten digits $0-9$ encoded as $256$-dimensional vectors. 

% There are $7291$ training images and $2007$ testing images in this dataset.

For  Figure \ref{fig7a}, we use the digit pair `2' and `6'. Overall, there are $n=1395$ total training vectors and $368$ total testing vectors corresponding to this digit pair. The figure shows that LDA achieves the lowest empirical error on this digit pair. This performance is matched by $\alpha$-LDA at $\alpha=1$. Although $\hat{\varepsilon}$ does not exactly match the empirical error, for parameter tuning it suffices that it follows the same trend. In this case, if we had directly used $\hat{\varepsilon}$ to tune the $\alpha$ parameter, we would have set it to $\alpha=0.85$. This setting results in an increase of merely $0.0054$ in error compared to the optimal setting. For more sensitive applications, the parameter setting suggested by the G-estimator may be used as a starting point from which to search for the optimal $\alpha$ using a more accurate (but computationally-intensive) method.

For  Figure \ref{fig7b}, we use the digit pair `5' and `8'. Overall, there are $n=1098$ total training vectors and $326$ total testing vectors corresponding to this digit pair. In this case, $\alpha$-LDA achieves the lowest error of $0.0307$ at $\alpha=0.65$. This is a $16.6\%$ decrease in error relative to LDA which has an error rate of $0.0368$. If we had directly used $\hat{\varepsilon}$ to tune the $\alpha$ parameter, we would have set it to $\alpha=0.8$. This setting incurs no loss in accuracy. Notice this dataset has less training samples than the last one. The increased estimation noise explains why $\alpha$-LDA is able to provide a performance advantage over LDA.

\begin{figure}
    \centering
    \includegraphics[width=0.5\linewidth]{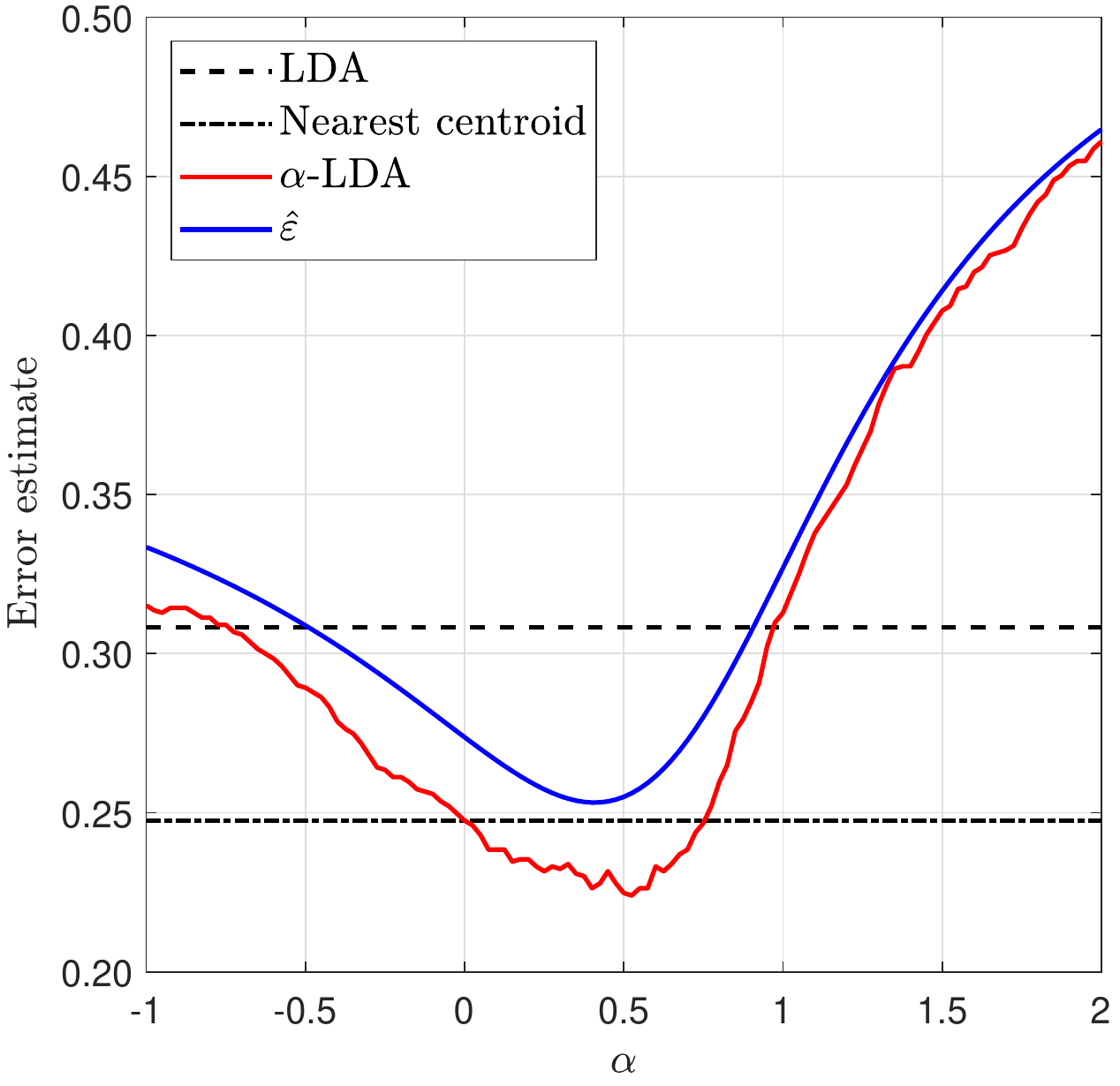}
    \caption{Plots of testing error estimates of classifying phonemes `aa' and `ao' for LDA, the nearest centroid, and $\alpha$-LDA as well as the G-estimator $\hat{\varepsilon}$ of the $\alpha$-LDA expected testing error.}
    \label{fig8}
\end{figure}

Figure \ref{fig8} considers a phoneme pair. The phoneme dataset consists of a total of $4509$ instances of digitized speech vectors of the five phonemes `aa', `ao', `dcl', `iy', and `sh', having $256$ features each. All $1717$ instances of the phonemes `ao' and `aa' (which are the closest in pronunciation) were extracted in order to construct this binary classification problem. As the dataset is not pre-divided into training and testing sets, the splitting was performed randomly. We take advantage of this to construct a classification problem in which $n$ is not much greater than $p$. A training set consisting of $400$ samples is randomly extracted from the full set of `aa' and `ao' phonemes according to the same proportions. This leaves $1317$ samples for testing. Based on the simulations from the previous section, we expect to observe a much greater performance gain in this scenario compared to Figure \ref{fig7}.

 Figure \ref{fig8} shows that, as expected, $\alpha$-LDA significantly outperforms LDA with an error of $0.224$ corresponding to the former compared to $0.3083$ corresponding to the latter. It achieves a $27.3\%$ decrease in error at $\alpha=0.525$. In this case, it seems that the data leans more towards an isotropic covariance structure, as nearest centroid performs better than LDA. Even so, $\alpha=0$ is not optimal. Thus, $\alpha$-LDA provides the best balance between both of these classifiers. Lastly, the G-estimator points towards an $\alpha$ setting of $0.4$. Using this setting incurs an increase in error of just $0.0023$ relative to the optimal setting. 

\section{Conclusion}
In this work, we design a method of weight vector tuning for binary linear classifiers based on the decomposition of the discriminant into informative and noisy components. The tuning takes the form of a linear parameterization of the decomposition. Deriving this method reveals a novel interpretation of the classic LDA classifier weight vector as minimizing the noise from the test point to which it is applied.

We simulate the performance gain of this method for a variety of linear classifiers: LDA, SVM, logistic regression, R-LDA, and RP-LDA ensemble, and under different data dimensionality and sample size settings. Firstly, we find that weight vector tuning can compensate performance loss due to poorly chosen native classifier hyperparameters. It thus eliminates the need for native hyperparameter tuning. As weight vector tuning occurs post weight vector generation, this can be advantageous in terms of computational efficiency when the native hyperparameters need to be set prior to weight vector generation. Secondly, we find that the parameterization significantly improves the performance of LDA under high estimation noise. We proceed to derive the parameterized LDA classifier misclassification probability in the RMT growth regime corresponding to these settings, in which the data dimensionality and sample size grow at comparable rates to each other. We also provide an estimator of the probability of misclassification which neither relies on additional data samples nor requires intensive computations, and thus can be used to tune the parameter of this classifier in a computationally efficient manner.

\appendices
\section{Analysis of the Projected Test Point in the case of Unknown Means}\label{sampleProof}

\subsection{Common Covariance Matrices}
As in the derivation of Section \ref{knownMeans}, assume $\textbf{x}\in\mathcal{C}_i$, where $i$ is either 0 or 1 and assume the two classes have a common covariance matrix $\bm{\Sigma}$. Then $\textbf{x}\sim\bm{\mu}_i+\bm{\Sigma}^{1/2}\textbf{z}$ where $\textbf{z}\sim\mathcal{N}\left(\textbf{0},\textbf{I}\right)$.

Using the fact that $\hat{\bm{\mu}}_i=\bm{\mu}_i+\frac{\bm{\Sigma}^{1/2}\textbf{Z}_i\textbf{1}}{n_i}$ for some $\textbf{Z}_i$ with $\mathcal{N}(\textbf{0},\textbf{I})$ columns, $i=0,1$, $\hat{\tilde{\bm{x}}}$ can be expressed as 
\begin{align}
 \hat{\tilde{\bm{x}}}&=\frac{(-1)^{i+1}}{2}\bm{\mu}+\bm{\Sigma}^{1/2}\textbf{z}-\frac{1}{2}\frac{\bm{\Sigma}^{1/2}\textbf{Z}_0\textbf{1}}{n_0}-\frac{1}{2}\frac{\bm{\Sigma}^{1/2}\textbf{Z}_1\textbf{1}}{n_1}.
 \label{eq1}
\end{align}
The first term in \eqref{genDecoUnknown} can then be rewritten as
\begin{align}
    \frac{\textbf{w}^T\hat{\bm{\mu}}}{\hat{\bm{\mu}}^T\hat{\bm{\mu}}}\hat{\bm{\mu}}^T\hat{\tilde{\bm{x}}}&=\underbrace{(-1)^{i+1}\frac{1}{2}\frac{\textbf{w}^T\hat{\bm{\mu}}}{\hat{\bm{\mu}}^T\hat{\bm{\mu}}}\hat{\bm{\mu}}^T\bm{\mu}}_{I_1 (\text{information})}+\overbrace{ \frac{\textbf{w}^T\hat{\bm{\mu}}}{\hat{\bm{\mu}}^T\hat{\bm{\mu}}}\hat{\bm{\mu}}^T\left(\bm{\Sigma}^{1/2}\textbf{z}-\frac{1}{2}\frac{\bm{\Sigma}^{1/2}\textbf{Z}_0\textbf{1}}{n_0}-\frac{1}{2}\frac{\bm{\Sigma}^{1/2}\textbf{Z}_1\textbf{1}}{n_1}\right)}^{N_1 (\text{noise})}
\end{align} 
Note that the noise here is due to both the common covariance between the classes (this is the test point noise) as well as estimation noise from the sample means.
Similarly, the second term can be expressed using \eqref{eq1} as 
\begin{align}
   \textbf{w}^T\textbf{P}_{\hat{\bm{\mu}}}\hat{\tilde{\bm{x}}}&=\underbrace{\frac{(-1)^{i+1}}{2}\textbf{w}^T\textbf{P}_{\hat{\bm{\mu}}}{\bm{\mu}}}_{I_2 (\text{information})}+\underbrace{\textbf{w}^T\textbf{P}_{\hat{\bm{\mu}}}\left(\bm{\Sigma}^{1/2}\textbf{z}-\frac{1}{2}\frac{\bm{\Sigma}^{1/2}\textbf{Z}_0\textbf{1}}{n_0}-\frac{1}{2}\frac{\bm{\Sigma}^{1/2}\textbf{Z}_1\textbf{1}}{n_1}\right)}_{N_2 (\text{noise})}
\end{align}

Alternatively, expressing \eqref{eq1} in terms of $\hat{\bm{\mu}}$ rather than $\bm{\mu}$ so that the orthogonal projection can be put to use yields a similar result. By using the fact that $\bm{\mu}=\hat{\bm{\mu}}+\frac{\bm{\Sigma}^{1/2}\textbf{Z}_0\textbf{1}}{n_0}-\frac{\bm{\Sigma}^{1/2}\textbf{Z}_1\textbf{1}}{n_1}$, \eqref{eq1} can be expressed as
\begin{equation}
     \hat{\tilde{\bm{x}}}=\frac{(-1)^{i+1}}{2}\hat{\bm{\mu}}+\bm{\Sigma}^{1/2}\textbf{z}-\frac{\bm{\Sigma}^{1/2}\textbf{Z}_i\textbf{1}}{n_i}\label{metoo_b}
\end{equation}
and the second term in \eqref{genDecoUnknown} as
\begin{align}
   \textbf{w}^T\textbf{P}_{\hat{\bm{\mu}}}\hat{\tilde{\bm{x}}}&=   \textbf{w}^T\textbf{P}_{\hat{\bm{\mu}}}\left(\frac{(-1)^{i+1}}{2}\hat{\bm{\mu}}+\bm{\Sigma}^{1/2}\textbf{z}-\frac{\bm{\Sigma}^{1/2}\textbf{Z}_i\textbf{1}}{n_i}\right)\\
   &=\underbrace{\textbf{w}^T\textbf{P}_{\hat{\bm{\mu}}}\bm{\Sigma}^{1/2}\textbf{z}}_{N_2 (\text{noise })}-\overbrace{\textbf{w}^T\textbf{P}_{\hat{\bm{\mu}}}\left(\frac{\bm{\Sigma}^{1/2}\textbf{Z}_i\textbf{1}}{n_i}\right)}^{I_2 (\text{information})}
\end{align}
From this perspective, the information in the test point combines with the sample estimation noise in the term $\frac{\bm{\Sigma}^{1/2}\textbf{Z}_i\textbf{1}}{n_i}$. Note that even if we were to have equal samples so that $n_0=n_1$, we would still be able to discriminate the class of the test point through $\textbf{Z}_i$. This is not immediately obvious as $\textbf{Z}_i, \ i=0,1$, both have the same distribution, but can be observed asymptotically by computing the deterministic equivalents.

% The probability that $\textbf{Z}_0=\textbf{Z}_1$ (the only case for which this term would not provide any discriminatory information when $n_0=n_1$) is zero. DIDN't I change this? cuz DEs are the same no?

\subsection{Distinct Covariance Matrices}
An analogous result to that of the previous section can be derived in the case when the class covariance matrices are distinct. In this case, $\textbf{x}\sim\bm{\mu}_i+\bm{\Sigma}_i^{1/2}\textbf{z}$ where $\textbf{z}\sim\mathcal{N}\left(\textbf{0},\textbf{I}\right)$. Using the fact that $\hat{\bm{\mu}}_i=\bm{\mu}_i+\frac{\bm{\Sigma}_i^{1/2}\textbf{Z}_i\textbf{1}}{n_i}$ for some $\textbf{Z}_i$ with $\mathcal{N}(\textbf{0},\textbf{I})$ columns, $i=0,1$,
\begin{align}
 \hat{\tilde{\bm{x}}}&=\frac{(-1)^{i+1}}{2}\bm{\mu}+\bm{\Sigma}_i^{1/2}\textbf{z}-\frac{1}{2}\frac{\bm{\Sigma}_0^{1/2}\textbf{Z}_0\textbf{1}}{n_0}-\frac{1}{2}\frac{\bm{\Sigma}_1^{1/2}\textbf{Z}_1\textbf{1}}{n_1}.
 \label{eq1_b}
\end{align}
The first term in \eqref{genDecoUnknown} can then be rewritten as
\begin{align}
    \frac{\textbf{w}^T\hat{\bm{\mu}}}{\hat{\bm{\mu}}^T\hat{\bm{\mu}}}\hat{\bm{\mu}}^T\hat{\tilde{\bm{x}}}&=\underbrace{(-1)^{i+1}\frac{1}{2}\frac{\textbf{w}^T\hat{\bm{\mu}}}{\hat{\bm{\mu}}^T\hat{\bm{\mu}}}\hat{\bm{\mu}}^T\bm{\mu}+\frac{\textbf{w}^T\hat{\bm{\mu}}}{\hat{\bm{\mu}}^T\hat{\bm{\mu}}}\hat{\bm{\mu}}^T\bm{\Sigma}_i^{1/2}\textbf{z}}_{I_1 (\text{information})}-\frac{1}{2}\underbrace{ \frac{\textbf{w}^T\hat{\bm{\mu}}}{\hat{\bm{\mu}}^T\hat{\bm{\mu}}}\hat{\bm{\mu}}^T\left(\frac{\bm{\Sigma}_0^{1/2}\textbf{Z}_0\textbf{1}}{n_0}+\frac{\bm{\Sigma}_1^{1/2}\textbf{Z}_1\textbf{1}}{n_1}\right)}_{N_1 (\text{noise 1})}
\end{align} 
Note here that the noise is due only to estimation noise from the sample means, since the differing covariances between the two classes are informative.

Substituting \eqref{eq1_b} directly into the second term in \eqref{genDecoUnknown} gives
\begin{align}
   \textbf{w}^T\textbf{P}_{\hat{\bm{\mu}}}\hat{\tilde{\bm{x}}}&=\underbrace{\frac{(-1)^{i+1}}{2}\textbf{w}^T\textbf{P}_{\hat{\bm{\mu}}}{\bm{\mu}}+ \textbf{w}^T\textbf{P}_{\hat{\bm{\mu}}}\bm{\Sigma}_i^{1/2}\textbf{z}}_{I_2\text{ (information)}}-\frac{1}{2}\underbrace{\textbf{w}^T\textbf{P}_{\hat{\bm{\mu}}}\left(\frac{\bm{\Sigma}_0^{1/2}\textbf{Z}_0\textbf{1}}{n_0}+\frac{\bm{\Sigma}_1^{1/2}\textbf{Z}_1\textbf{1}}{n_1}\right)}_{N_2 (\text{noise})}
\end{align}

Alternatively, expressing \eqref{eq1_b} in terms of $\hat{\bm{\mu}}$ reveals the second term in \eqref{genDecoUnknown} to be purely information. By using the fact that $\bm{\mu}=\hat{\bm{\mu}}+\frac{\bm{\Sigma}_0^{1/2}\textbf{Z}_0\textbf{1}}{n_0}-\frac{\bm{\Sigma}_1^{1/2}\textbf{Z}_1\textbf{1}}{n_1}$, 
\begin{equation}
    \hat{\tilde{\bm{x}}}=\frac{(-1)^{i+1}}{2}\hat{\bm{\mu}}+\bm{\Sigma}_i^{1/2}\textbf{z}-\frac{\bm{\Sigma}_i^{1/2}\textbf{Z}_i\textbf{1}}{n_i}\label{metoo}
\end{equation}
and the second term in \eqref{genDecoUnknown} becomes
\begin{align}
   \textbf{w}^T\textbf{P}_{\hat{\bm{\mu}}}\hat{\tilde{\bm{x}}}&=   \textbf{w}^T\textbf{P}_{\hat{\bm{\mu}}}\left(\frac{(-1)^{i+1}}{2}\hat{\bm{\mu}}+\bm{\Sigma}_i^{1/2}\textbf{z}-\frac{\bm{\Sigma}_i^{1/2}\textbf{Z}_i\textbf{1}}{n_i}\right)\\
   &=\underbrace{\textbf{w}^T\textbf{P}_{\hat{\bm{\mu}}}\bm{\Sigma}_i^{1/2}\textbf{z}-\textbf{w}^T\textbf{P}_{\hat{\bm{\mu}}}\left(\frac{\bm{\Sigma}_i^{1/2}\textbf{Z}_i\textbf{1}}{n_i}\right)}_{I_2 (\text{information})}
\end{align}

\section{Derivation of the Deterministic Equivalent of the Probability of Misclassification}\label{DE}

Deriving $\bar{\varepsilon}$ reduces to deriving the deterministic equivalents $\bar{m}_0$, $\bar{m}_1$, $\bar{\sigma}_0^2$, and $\bar{\sigma}_1^2$ reduces to deriving the deterministic equivalents of their constituent quadratic forms. This is the approach taken in what follows.

\subsection{Distinct Covariances}\label{DE_distinct}
The following proofs rely heavily on three main facts. First, under the distinct covariance assumption on the class distributions, $\hat{\bm{\mu}}_i=\bm{\mu}_i+\frac{\bm{\Sigma}_i^{1/2}\textbf{Z}_i\textbf{1}}{n_i}$ for some $\textbf{Z}_i$ with $\mathcal{N}(\textbf{0},\textbf{I})$ columns, $i=0,1$. This follows from expressing the data matrices $\textbf{X}_i, \ i=0,1$ as $\textbf{X}_i=\bm{\mu}_i\textbf{1}_{n_i}^T+\bm{\Sigma}_i^{1/2}\textbf{Z}_i$ for some $\textbf{Z}_i, \ i=0,1$ with columns distributed as $\mathcal{N}\left(\textbf{0}_{p},\textbf{I}_p\right)$. Then $\hat{\bm{\mu}}_i=\frac{1}{n_i}\textbf{X}_i\textbf{1}_{n_i}=\bm{\mu}_i+\frac{\bm{\Sigma}_i^{1/2}\textbf{Z}_i\textbf{1}}{n_i}$.

Second, the sample means $\bm{\mu}_0$ and $\bm{\mu}_1$ are independent of the sample covariance $\bm{\Sigma}$. This can be shown by simply plugging in $\textbf{X}_i=\bm{\mu}_i\textbf{1}_{n_i}^T+\bm{\Sigma}_i^{1/2}\textbf{Z}_i$ and $\hat{\bm{\mu}}_i=\bm{\mu}_i+\frac{\bm{\Sigma}_i^{1/2}\textbf{Z}_i\textbf{1}}{n_i}$ into the corresponding formula for $\hat{\bm{\Sigma}}_i$. This yields
\begin{equation}
    \hat{\bm{\Sigma}}_i=\frac{1}{n_i-1}\bm{\Sigma}_i^{1/2}\textbf{Z}_i\left(\textbf{I}_{n_i}-\frac{\textbf{1}_{n_i}\textbf{1}_{n_i}^T}{n_i}\right)\textbf{Z}_i^T\bm{\Sigma}_i^{1/2}. \label{Sigma_i}
\end{equation}
Since the terms $\textbf{Z}_i\textbf{1}_{n_i}$ and $\textbf{Z}_i\left(\textbf{I}_{n_i}-\frac{\textbf{1}_{n_i}\textbf{1}_{n_i}^T}{n_i}\right)$ are Gaussian and uncorrelated, due to the projection matrix $\left(\textbf{I}_{n_i}-\frac{\textbf{1}_{n_i}\textbf{1}_{n_i}^T}{n_i}\right)$, they are independent. Thus, $\hat{\bm{\mu}}_i$ and $\hat{\bm{\Sigma}}_i$ are independent. Of course, $\hat{\bm{\mu}}_i$ and $\hat{\bm{\Sigma}}_i$ where $i\ne j$ are also independent since $\textbf{Z}_0$ is independent of $\textbf{Z}_1$. It follows that $\hat{\bm{\Sigma}}$ which is a function of $\hat{\bm{\Sigma}}_i, \ i=0,1$, is independent of $\hat{\bm{\mu}}_i, \ i=0,1$.

Lastly, $\hat{\bm{\Sigma}}$ can be expressed as 
\begin{equation}
     \hat{\bm{\Sigma}}=\frac{1}{n-2}\bm{\Sigma}_0^{1/2}\bar{\textbf{Z}}_0\bar{\textbf{Z}}_0^T\bm{\Sigma}_0^{1/2}+\frac{1}{n-2}\bm{\Sigma}_1^{1/2}\bar{\textbf{Z}}_1\bar{\textbf{Z}}_1^T\bm{\Sigma}_1^{1/2} \label{decoSigma}
\end{equation}
for some $\bar{\textbf{Z}}_0\in\mathbb{R}^{p\times (n_0-1)}$ and $\bar{\textbf{Z}}_1\in\mathbb{R}^{p\times (n_1-1)}$, both having columns distributed as $\mathcal{N}\left(\textbf{0}_p,\textbf{I}_p\right)$. Using \eqref{Sigma_i},
\begin{equation}
    \hat{\bm{\Sigma}}=\frac{1}{n-2}\bm{\Sigma}_0^{1/2}\textbf{Z}_0\left(\textbf{I}_{n_0}-\frac{\textbf{1}_{n_0}\textbf{1}_{n_0}^T}{n_0}\right)\textbf{Z}_0^T\bm{\Sigma}_0^{1/2}+\frac{1}{n-2}\bm{\Sigma}_1^{1/2}\textbf{Z}_1\left(\textbf{I}_{n_1}-\frac{\textbf{1}_{n_1}\textbf{1}_{n_1}^T}{n_1}\right)\textbf{Z}_1^T\bm{\Sigma}_1^{1/2}.
\end{equation}
 Since the terms $\frac{\textbf{1}_{n_0}\textbf{1}_{n_0}^T}{n_0}$ and $\frac{\textbf{1}_{n_1}\textbf{1}_{n_1}^T}{n_1}$ each have one eigenvalue which is equal to $1$ in both cases, their eigendecompositions can be represented as
\begin{equation}
\frac{\textbf{1}_{n_0}\textbf{1}_{n_0}^T}{n_0}=\textbf{U}_0\begin{bmatrix}
1 &  &  & \\
 & 0 &  & \\
 &  & \ddots & \\
 &  &  & 0
\end{bmatrix}\textbf{U}_0^T\hspace{15pt}
\text{and} \hspace{15pt}\frac{\textbf{1}_{n_1}\textbf{1}_{n_1}^T}{n_1}=\textbf{U}_1\begin{bmatrix}
1 &  &  & \\
 & 0 &  & \\
 &  & \ddots & \\
 &  &  & 0
\end{bmatrix}\textbf{U}_1^T\label{wow}
\end{equation}
where $\textbf{U}_0$ and $\textbf{U}_1$ have as their first columns the vectors $\frac{\textbf{1}_{n_0}}{\sqrt{n_0}}$ and $\frac{\textbf{1}_{n_1}}{\sqrt{n_1}}$ respectively. By using these same bases to eigendecompose $\textbf{I}_{n_0}$ and $\textbf{I}_{n_1}$ in \eqref{wow}, we obtain
\begin{align}
   \hat{\bm{\Sigma}}&=\frac{1}{n-2}\bm{\Sigma}_0^{1/2}{\textbf{Z}}_0\textbf{U}_0\begin{bmatrix}
0 &  &  & \\
 & 1 &  & \\
 &  & \ddots & \\
 &  &  & 1
\end{bmatrix}\textbf{U}_0^T{\textbf{Z}}_0^T\bm{\Sigma}_0^{1/2}+\frac{1}{n-2}\bm{\Sigma}^{1/2}{\textbf{Z}}_1\textbf{U}_1\begin{bmatrix}
0 &  &  & \\
 & 1 &  & \\
 &  & \ddots & \\
 &  &  & 1
\end{bmatrix}\textbf{U}_1^T{\textbf{Z}}_1^T\bm{\Sigma}_1^{1/2}\\
&\sim\frac{1}{n-2}\bm{\Sigma}_0^{1/2}{\textbf{Z}}_0\begin{bmatrix}
0 &  &  & \\
 & 1 &  & \\
 &  & \ddots & \\
 &  &  & 1
\end{bmatrix}{\textbf{Z}}_0^T\bm{\Sigma}_0^{1/2}+\frac{1}{n-2}\bm{\Sigma}_1^{1/2}{\textbf{Z}}_1\begin{bmatrix}
0 &  &  & \\
 & 1 &  & \\
 &  & \ddots & \\
 &  &  & 1
\end{bmatrix}^T{\textbf{Z}}_1^T\bm{\Sigma}_1^{1/2}\\
&=\frac{1}{n-2}\bm{\Sigma}_0^{1/2}\bar{\textbf{Z}}_0\bar{\textbf{Z}}_0^T\bm{\Sigma}_0^{1/2}+\frac{1}{n-2}\bm{\Sigma}_1^{1/2}\bar{\textbf{Z}}_1\bar{\textbf{Z}}_1^T\bm{\Sigma}_1^{1/2}
\end{align}
where $\bar{\textbf{Z}}_0\in\mathbb{R}^{p\times (n_0-1)}$ is the submatrix of ${\textbf{Z}}_0$ obtained by removing its first column and $\bar{\textbf{Z}}_1\in\mathbb{R}^{p\times (n_1-1)}$ is the submatrix of ${\textbf{Z}}_1$ obtained by removing its first column.

Now we are ready to derive the deterministic equivalents.

\subsubsection{Derivation of $\bar{m}_0$}\label{moore}
The discriminant mean $m_0$ can be expressed as
\begin{align}
    m_0&=(1-\alpha)\rho\hat{\bm{\mu}}^T\left({\bm{\mu}}_0-\frac{\hat{\bm{\mu}}_0+\hat{\bm{\mu}}_1}{2}\right)+\alpha\hat{\bm{\mu}}^T\hat{\bm{\Sigma}}^{-1}\left({\bm{\mu}}_0-\frac{\hat{\bm{\mu}}_0+\hat{\bm{\mu}}_1}{2}\right)
\end{align}
Thus, the problem of deriving this deterministic equivalent can be further decomposed into deriving the following convergence statements
\begin{equation}
    \hat{\bm{\mu}}^T\hat{\bm{\mu}}\asymp\bm{\mu}^T\bm{\mu}+\frac{1}{n_0}\text{tr}\left\{\bm{\Sigma}_0\right\}+\frac{1}{n_1}\text{tr}\left\{\bm{\Sigma}_1\right\}
\end{equation}

\begin{equation}
    \hat{\bm{\mu}}^T\left(\bm{\mu}_0-\frac{\hat{\bm{\mu}}_0+\hat{\bm{\mu}}_1}{2}\right)\asymp-\frac{1}{2}\bm{\mu}^T\bm{\mu}+\frac{1}{2}\left(\frac{1}{n_0}\text{tr}\left\{\bm{\Sigma}_0\right\}-\frac{1}{n_1}\text{tr}\left\{\bm{\Sigma}_1    \right\}\right)
\end{equation}
\begin{equation}
    \hat{\bm{\mu}}^T\hat{\bm{\Sigma}}^{-1}\hat{\bm{\mu}}\asymp {\bm{\mu}}^T\bar{\textbf{Q}}{\bm{\mu}}+\frac{1}{n_0}\text{tr}\left\{\bm{\Sigma}_0\bar{\textbf{Q}}\right\}+\frac{1}{n_1}\text{tr}\left\{\bm{\Sigma}_1\bar{\textbf{Q}}\right\}
\end{equation}
\begin{equation}
    \hat{\bm{\mu}}^T\hat{\bm{\Sigma}}^{-1}\left(\bm{\mu}_0-\frac{\hat{\bm{\mu}}_0+\hat{\bm{\mu}}_1}{2}\right)\asymp -\frac{1}{2}{\bm{\mu}}^T\bar{\textbf{Q}}{\bm{\mu}}+\frac{1}{2}\left(\frac{1}{n_0}\text{tr}\left\{\bm{\Sigma}_0\bar{\textbf{Q}}\right\}-\frac{1}{n_1}\text{tr}\left\{\bm{\Sigma}_1\bar{\textbf{Q}}\right\}\right)
\end{equation}
The first two convergence statements are derived by using the fact that $\hat{\bm{\mu}}_i=\bm{\mu}_i+\frac{\bm{\Sigma}_i^{1/2}\textbf{Z}_i\textbf{1}}{n_i}$ for some $\textbf{Z}_i$ with $\mathcal{N}(\textbf{0},\textbf{I})$ columns, $i=0,1$ and taking the expectation. The terms converge to their respective expectations according to Lemmas 17 and 19 in \cite{muller2016random}. The third and fourth terms involve $\hat{\bm{\Sigma}}$. Since $\hat{\bm{\mu}}$ and $\hat{\bm{\Sigma}}$ are independent, the convergence can be split into stages. 

For the third term, we first have the intermediate convergence result
\begin{equation}
    \hat{\bm{\mu}}^T\hat{\bm{\Sigma}}^{-1}\hat{\bm{\mu}}\asymp {\bm{\mu}}^T\hat{\bm{\Sigma}}^{-1}{\bm{\mu}}+\frac{1}{n_0}\text{tr}\left\{\bm{\Sigma}_0\hat{\bm{\Sigma}}^{-1}\right\}+\frac{1}{n_1}\text{tr}\left\{\bm{\Sigma}_1\hat{\bm{\Sigma}}^{-1}\right\},
\end{equation}
and for the fourth term we have the intermediate convergence result
\begin{equation}
     \hat{\bm{\mu}}^T\hat{\bm{\Sigma}}^{-1}\left(\bm{\mu}_0-\frac{\hat{\bm{\mu}}_0+\hat{\bm{\mu}}_1}{2}\right)\asymp -\frac{1}{2}{\bm{\mu}}^T\hat{\bm{\Sigma}}^{-1}{\bm{\mu}}+\frac{1}{2}\left(\frac{1}{n_0}\text{tr}\left\{\bm{\Sigma}_0\hat{\bm{\Sigma}}^{-1}\right\}-\frac{1}{n_1}\text{tr}\left\{\bm{\Sigma}_1\hat{\bm{\Sigma}}^{-1}\right\}\right)
\end{equation}
each obtained by dealing with $\hat{\bm{\mu}}$ as described above independently of $\hat{\bm{\Sigma}}$.

Next, we express $\hat{\bm{\Sigma}}=\textbf{W}\textbf{W}^T$ where $\textbf{W}\in \mathbb{R}^{p\times (n-2)}$ is defined as 
\begin{equation}
    \textbf{W}=\frac{1}{\sqrt{p}}\left[\sqrt{\frac{p}{n-2}}\bm{\Sigma}_0^{1/2}\bar{\textbf{Z}}_0 \ \sqrt{\frac{p}{n-2}}\bm{\Sigma}_1^{1/2}\bar{\textbf{Z}}_1\right]
\end{equation}

Now define $\textbf{Q}_{\gamma}=\left(\textbf{W}\textbf{W}^T-\gamma\textbf{I}_p\right)^{-1}, \ \gamma<0$. According to \cite{benaych2016spectral},
\begin{equation}
    \textbf{Q}_{\gamma}\leftrightarrow \bar{\textbf{Q}}_{\gamma}
\end{equation}
where 
\begin{equation}
    \bar{\textbf{Q}}_{\gamma}=-\frac{1}{\gamma}\left(\textbf{I}_p+\frac{n_0-1}{n-2}\delta(\gamma)\frac{p}{n-2}\bm{\Sigma}_0+\frac{n_1-1}{n-2}\nu(\gamma)\frac{p}{n-2}\bm{\Sigma}_1\right)^{-1},\label{Q1}
\end{equation}
\begin{equation}
    \frac{p}{n-2}\delta(\gamma)=-\frac{1}{\gamma}\frac{1}{1+\tilde{\delta}(\gamma)},\label{delta}
\end{equation}
\begin{equation}
    \tilde{\delta}(\gamma)=-\frac{1}{\gamma}\frac{1}{p}\text{tr}\left\{\frac{p}{n-2}\bm{\Sigma}_0\left(\textbf{I}_p+\frac{n_0-1}{n-2}\delta(\gamma)\frac{p}{n-2}\bm{\Sigma}_0+\frac{n_1-1}{n-2}\nu(\gamma)\frac{p}{n-2}\bm{\Sigma}_1\right)^{-1}\right\},\label{deltaTilde}
\end{equation}
\begin{equation}
     \frac{p}{n-2}\nu(\gamma)=-\frac{1}{\gamma}\frac{1}{1+\tilde{\nu}(\gamma)},\label{nu}
\end{equation}
and
\begin{equation}
     \tilde{\nu}(\gamma)=-\frac{1}{\gamma}\frac{1}{p}\text{tr}\left\{\frac{p}{n-2}\bm{\Sigma}_1\left(\textbf{I}_p+\frac{n_0-1}{n-2}\delta(\gamma)\frac{p}{n-2}\bm{\Sigma}_0+\frac{n_1-1}{n-2}\nu(\gamma)\frac{p}{n-2}\bm{\Sigma}_1\right)^{-1}\right\}\label{nuTilde}
\end{equation}
The expressions we are working with can be expressed in this notation as
\begin{equation}
     \lim_{\gamma\rightarrow 0}\hat{\bm{\mu}}^T\textbf{Q}_{\gamma}\hat{\bm{\mu}}
\end{equation}
and 
\begin{equation}
      \lim_{\gamma\rightarrow 0}\hat{\bm{\mu}}^T\textbf{Q}_{\gamma}\left(\bm{\mu}_0-\frac{\hat{\bm{\mu}}_0+\hat{\bm{\mu}}_1}{2}\right)
\end{equation}
and we want to derive the corresponding DEs by taking the limits
\begin{equation}
   \lim_{n,p\rightarrow \infty} \lim_{\gamma\rightarrow 0}\hat{\bm{\mu}}^T\textbf{Q}_{\gamma}\hat{\bm{\mu}}
\end{equation}

\begin{equation}
    \lim_{n,p\rightarrow \infty} \lim_{\gamma\rightarrow 0}\hat{\bm{\mu}}^T\textbf{Q}_{\gamma}\left(\bm{\mu}_0-\frac{\hat{\bm{\mu}}_0+\hat{\bm{\mu}}_1}{2}\right).
\end{equation}
The Moore-Osgood theorem allows the interchange of these limits. It is enough to show that the sequences $\hat{\bm{\mu}}^T\textbf{Q}_{\gamma}\hat{\bm{\mu}}$ and $\hat{\bm{\mu}}^T\textbf{Q}_{\gamma}\left(\bm{\mu}_0-\frac{\hat{\bm{\mu}}_0+\hat{\bm{\mu}}_1}{2}\right)$ converge uniformly. Since these sequences converge pointwise (this follows from convergence in probability), this can be shown by uniformly bounding their first derivative \cite{875205}. 

We have 
\begin{align}
    \frac{d\left[\hat{\bm{\mu}}^T\textbf{Q}_{\gamma}\hat{\bm{\mu}}\right]}{d\gamma}&=\hat{\bm{\mu}}^T\left(\hat{\bm{\Sigma}}-\gamma\textbf{I}_p\right)^{-2}\hat{\bm{\mu}}^T\\
    &\leq \lVert\hat{\bm{\mu}}\rVert_2^2\left\lVert\left(\hat{\bm{\Sigma}}-\gamma\textbf{I}_p\right)^{-1}\right\rVert_2^2\\
    &=\frac{\lVert\hat{\bm{\mu}}\rVert_2^2}{\lambda_{\text{min}}\{\hat{\bm{\Sigma}}\}-\gamma}\\
    &\leq \frac{\lVert\hat{\bm{\mu}}\rVert_2^2}{ C}
\end{align}
where the last line follows from the result in \cite{kammoun2014smallest} which shows that $\lambda_{\text{min}}\{\hat{\bm{\Sigma}}\}>C$ for some constant $C$ almost surely. Using the growth regime assumption (c) it can be shown that $\lVert\hat{\bm{\mu}}\rVert_2^2$ is bounded. This completes the proof. The other term can be handled in a similar way.

We can now apply the result in \cite{benaych2016spectral} which manifests in equations \eqref{Q1}, \eqref{delta}, \eqref{deltaTilde}, \eqref{nu}, and \eqref{nuTilde}. We can then take the limit as $\gamma\rightarrow 0$.

Combining \eqref{delta} and \eqref{deltaTilde}, we have
\begin{equation}
    \frac{p}{n-2}\delta(\gamma)=-\frac{1}{\gamma}\frac{1}{1-\frac{1}{\gamma}\frac{1}{p}\text{tr}\left\{\frac{p}{n-2}\bm{\Sigma}_0\left(\textbf{I}_p+\frac{n_0-1}{n-2}\delta(\gamma)\frac{p}{n-2}\bm{\Sigma}_0+\frac{n_1-1}{n-2}\nu(\gamma)\frac{p}{n-2}\bm{\Sigma}_1\right)^{-1}\right\}}.
\end{equation}

Combining \eqref{nu} and \eqref{nuTilde}, we have
\begin{equation}
    \frac{p}{n-2}\nu(\gamma)=-\frac{1}{\gamma}\frac{1}{1-\frac{1}{\gamma}\frac{1}{p}\text{tr}\left\{\frac{p}{n-2}\bm{\Sigma}_1\left(\textbf{I}_p+\frac{n_0-1}{n-2}\delta(\gamma)\frac{p}{n-2}\bm{\Sigma}_0+\frac{n_1-1}{n-2}\nu(\gamma)\frac{p}{n-2}\bm{\Sigma}_1\right)^{-1}\right\}}.
\end{equation}

These equations shows that $\delta(\gamma)$ and $\nu(\gamma)$ vary as $\frac{1}{\gamma}$, and so they diverge as $\gamma\rightarrow 0$

Combining \eqref{deltaTilde}, \eqref{delta}, and \eqref{nu}, we have
\begin{equation}
    \tilde{\delta}(\gamma)=\frac{1}{n-2}\text{tr}\left\{\bm{\Sigma}_0\left(-\gamma\textbf{I}_p+\frac{n_0-1}{n-2}\frac{1}{1+\tilde{\delta}(\gamma)}\bm{\Sigma}_0+\frac{n_1-1}{n-2}\frac{1}{1+\tilde{\nu}(\gamma)}\bm{\Sigma}_1\right)^{-1}\right\},\label{algeq1}
\end{equation}
Combining \eqref{nuTilde}, \eqref{delta}, and \eqref{nu}, we have
\begin{equation}
    \tilde{\nu}(\gamma)=\frac{1}{n-2}\text{tr}\left\{\bm{\Sigma}_1\left(-\gamma\textbf{I}_p+\frac{n_0-1}{n-2}\frac{1}{1+\tilde{\delta}(\gamma)}\bm{\Sigma}_0+\frac{n_1-1}{n-2}\frac{1}{1+\tilde{\nu}(\gamma)}\bm{\Sigma}_1\right)^{-1}\right\},\label{algeq2}
\end{equation}
This pair of equations does not pose problems as $\gamma\rightarrow 0$, therefore we work with $\tilde{\delta}(\gamma)$ and $\tilde{\nu}(\gamma)$. Taking the limit as $\gamma\rightarrow 0$, \eqref{algeq1} becomes
\begin{equation}
    \tilde{\delta}(0)=\frac{1}{n-2}\text{tr}\left\{\bm{\Sigma}_0\left(\frac{n_0-1}{n-2}\frac{1}{1+\tilde{\delta}(0)}\bm{\Sigma}_0+\frac{n_1-1}{n-2}\frac{1}{1+\tilde{\nu}(0)}\bm{\Sigma}_1\right)^{-1}\right\},\label{algeq1_0}
\end{equation}
and \eqref{algeq2} becomes
\begin{equation}
    \tilde{\nu}(0)=\frac{1}{n-2}\text{tr}\left\{\bm{\Sigma}_1\left(\frac{n_0-1}{n-2}\frac{1}{1+\tilde{\delta}(0)}\bm{\Sigma}_0+\frac{n_1-1}{n-2}\frac{1}{1+\tilde{\nu}(0)}\bm{\Sigma}_1\right)^{-1}\right\},\label{algeq2_0}
\end{equation}
Although there are no closed-form solutions for $    \tilde{\delta}(0)$ and $\tilde{\nu}(0)$, these equations fit under the framework of a standard inference problem (see Definition 6.2 \cite{couillet2011random}). The fixed point iteration algorithm stated in Theorem 2 is guaranteed to converge to a unique solution $(\tilde{\delta}(0),\tilde{\nu}(0))$ (see Theorem 6.18 in \cite{couillet2011random}), denoted $(\tilde{\delta},\tilde{\nu})$ in the equations leading up to Theorem $2$. So, overall we have
\begin{equation}
    \hat{\bm{\mu}}^T\hat{\bm{\Sigma}}^{-1}\hat{\bm{\mu}}\asymp {\bm{\mu}}^T\bar{\textbf{Q}}{\bm{\mu}}+\frac{1}{n_0}\text{tr}\left\{\textbf{A}_0\right\}+\frac{1}{n_1}\text{tr}\left\{\textbf{A}_1\right\},
\end{equation}
and 
\begin{equation}
     \hat{\bm{\mu}}^T\hat{\bm{\Sigma}}^{-1}\left(\bm{\mu}_0-\frac{\hat{\bm{\mu}}_0+\hat{\bm{\mu}}_1}{2}\right)\asymp -\frac{1}{2}{\bm{\mu}}^T\bar{\textbf{Q}}{\bm{\mu}}+\frac{1}{2}\left(\frac{1}{n_0}\text{tr}\left\{\textbf{A}_0\right\}-\frac{1}{n_1}\text{tr}\left\{\textbf{A}_1\right\}\right),
\end{equation}
where
\begin{align}
    \bar{\textbf{Q}}&:=\lim_{\gamma\rightarrow 0}\bar{\textbf{Q}}_{\gamma}\\
    &=\left(\frac{n_0-1}{n-2}\frac{1}{1+\tilde{\delta}(0)}\bm{\Sigma}_0+\frac{n_1-1}{n-2}\frac{1}{1+\tilde{\nu}(0)}\bm{\Sigma}_1\right)^{-1}.
\end{align}

\subsubsection{Derivation of $\bar{m}_1$}
Similarly, the problem of deriving this deterministic equivalent can be further decomposed into deriving the following additional convergence statements 
\begin{equation}
    \hat{\bm{\mu}}^T\left(\bm{\mu}_1-\frac{\hat{\bm{\mu}}_0+\hat{\bm{\mu}}_1}{2}\right)\asymp\frac{1}{2}\bm{\mu}^T\bm{\mu}+\frac{1}{2}\left(\frac{1}{n_0}\text{tr}\left\{\bm{\Sigma}_0\right\}-\frac{1}{n_1}\text{tr}\left\{\bm{\Sigma}_1    \right\}\right)
\end{equation}

\begin{equation}
    \hat{\bm{\mu}}^T\hat{\bm{\Sigma}}^{-1}\left(\bm{\mu}_1-\frac{\hat{\bm{\mu}}_0+\hat{\bm{\mu}}_1}{2}\right)\asymp \frac{1}{2}{\bm{\mu}}^T\bar{\textbf{Q}}{\bm{\mu}}+\frac{1}{2}\left(\frac{1}{n_0}\text{tr}\left\{\textbf{A}_0\right\}-\frac{1}{n_1}\text{tr}\left\{\textbf{A}_1\right\}\right)
\end{equation}
which can be proven in a similar way to the terms composing $m_0$.

\subsubsection{Derivation of $\bar{\sigma}_0^2$}
The discriminant variance $\sigma_0^2$ can be expressed as
\begin{align}
    \sigma_0^2
    &=(1-\alpha)^2\rho^2\hat{\bm{\mu}}^T{\bm{\Sigma}}_0\hat{\bm{\mu}}+\alpha^2\hat{\bm{\mu}}^T\hat{\bm{\Sigma}}^{-1}\bm{\Sigma}_0\hat{\bm{\Sigma}}^{-1}\hat{\bm{\mu}}+2\alpha(1-\alpha)\rho\hat{\bm{\mu}}^T\bm{\Sigma}_0\hat{\bm{\Sigma}}^{-1}\hat{\bm{\mu}}
\end{align}
The problem of deriving this deterministic equivalent can be further decomposed into deriving the following additional convergence statements
\begin{equation}
    \hat{\bm{\mu}}^T\bm{\Sigma}_0 \hat{\bm{\mu}}\asymp \bm{\mu}^T\bm{\Sigma}_0\bm{\mu}+\frac{1}{n_0}\text{tr}\left\{\bm{\Sigma}_0^2\right\}+\frac{1}{n_1}\text{tr}\left\{\bm{\Sigma}_0\bm{\Sigma}_1\right\}
\end{equation}
\begin{equation}
\hat{\bm{\mu}}^T{\bm{\Sigma}_0}{\hat{\bm{\Sigma}}}^{-1}\hat{\bm{\mu}}\asymp\bm{\mu}^T\textbf{A}_0\bm{\mu}+\frac{1}{n_0}\text{tr}\left\{\bm{\Sigma}_0\textbf{A}_0\right\}+\frac{1}{n_1}\text{tr}\left\{\bm{\Sigma}_0\textbf{A}_1\right\}
\end{equation}
\begin{align}
  &\hat{\bm{\mu}}^T{\hat{\bm{\Sigma}}}^{-1}{\bm{\Sigma}_0}{\hat{\bm{\Sigma}}}^{-1}\hat{\bm{\mu}}\asymp \bm{\mu}^T\tilde{\textbf{Q}}_0\bm{\mu}+\frac{1}{n_0}\text{tr}\left\{\bm{\Sigma}_0\tilde{\textbf{Q}}_0\right\}+\frac{1}{n_1}\text{tr}\left\{\bm{\Sigma}_1\tilde{\textbf{Q}}_0\right\}
\end{align}

The first two results can be shown using the same techniques as above. The third result needs special treatment, as it involves a double resolvent. Using the result for double resolvents in \cite{benaych2016spectral}, in conjunction with taking $\gamma\rightarrow 0$, we have
\begin{equation}
    \hat{\bm{\Sigma}}^{-1}\bm{\Sigma}_0\hat{\bm{\Sigma}}^{-1}\leftrightarrow \tilde{\textbf{Q}}_0.
\end{equation}

\subsubsection{Derivation of $\bar{\sigma}_1^2$}
Similarly, the problem of deriving this deterministic equivalent can be further decomposed into deriving the following additional convergence statements

\begin{equation}
    \hat{\bm{\mu}}^T\bm{\Sigma}_1 \hat{\bm{\mu}}\asymp \bm{\mu}^T\bm{\Sigma}_1\bm{\mu}+\frac{1}{n_0}\text{tr}\left\{\bm{\Sigma}_0\bm{\Sigma}_1\right\}+\frac{1}{n_1}\text{tr}\left\{\bm{\Sigma}_1^2\right\}
\end{equation}
\begin{equation}
\hat{\bm{\mu}}^T{\bm{\Sigma}_1}{\hat{\bm{\Sigma}}}^{-1}\hat{\bm{\mu}}\asymp\bm{\mu}^T\textbf{A}_1\bm{\mu}+\frac{1}{n_0}\text{tr}\left\{\bm{\Sigma}_1\textbf{A}_0\right\}+\frac{1}{n_1}\text{tr}\left\{\bm{\Sigma}_1\textbf{A}_1\right\}
\end{equation}
\begin{align}
  &\hat{\bm{\mu}}^T{\hat{\bm{\Sigma}}}^{-1}{\bm{\Sigma}_1}{\hat{\bm{\Sigma}}}^{-1}\hat{\bm{\mu}}\asymp\bm{\mu}^T\tilde{\textbf{Q}}_1\bm{\mu}+\frac{1}{n_0}\text{tr}\left\{\bm{\Sigma}_0\tilde{\textbf{Q}}_1\right\}+\frac{1}{n_1}\text{tr}\left\{\bm{\Sigma}_1\tilde{\textbf{Q}}_1\right\}
\end{align}
The third convergence statement uses the result 
\begin{equation}
    \hat{\bm{\Sigma}}^{-1}\bm{\Sigma}_1\hat{\bm{\Sigma}}^{-1}\leftrightarrow \tilde{\textbf{Q}}_1
\end{equation}
from \cite{benaych2016spectral}.
\subsection{Common Covariances}\label{DE_common}
The following proofs rely heavily on three main facts. Firstly, under the assumption that the two classes have common covariance $\bm{\Sigma}$, $\hat{\bm{\mu}}_i=\bm{\mu}_i+\frac{\bm{\Sigma}^{1/2}\textbf{Z}_i\textbf{1}}{n_i}$ for some $\textbf{Z}_i$ with $\mathcal{N}(\textbf{0},\textbf{I})$ columns, $i=0,1$. Secondly, $\hat{\bm{\mu}}_i, \ i=0,1$ are independent of $\hat{\bm{\Sigma}}$. Finally, $\hat{\bm{\Sigma}}$ can be expressed as 
\begin{equation}
    \hat{\bm{\Sigma}}= \frac{1}{n-2}\bm{\Sigma}^{1/2}{\bar{\textbf{Z}}}{\bar{\textbf{Z}}}^T\bm{\Sigma}^{1/2}
\end{equation}
for some ${\textbf{Z}}\in\mathbb{R}^{p\times (n-2)}$ which has i.i.d. entries distributed as $\mathcal{N}({0},1)$. The proofs follow the same line of reasoning as those at the beginning of Section \ref{DE_distinct}.

\subsubsection{Derivation of $\bar{m}_0$}\label{barm0}
The problem of deriving this deterministic equivalent can be further decomposed into deriving the following convergence statements
\begin{equation}
    \hat{\bm{\mu}}^T\hat{\bm{\mu}}\asymp\bm{\mu}^T\bm{\mu}+\left(\frac{1}{n_0}+\frac{1}{n_1}\right)\text{tr}\left\{\bm{\Sigma}\right\}
\end{equation}
\begin{equation}
    \hat{\bm{\mu}}^T\hat{\bm{\Sigma}}^{-1}\hat{\bm{\mu}}\asymp\tau\left[\bm{\mu}^T\bm{\Sigma}^{-1}\bm{\mu}+\frac{p}{n_0}+\frac{p}{n_1}\right]
\end{equation}
\begin{equation}
    \hat{\bm{\mu}}^T\left(\bm{\mu}_0-\frac{\hat{\bm{\mu}}_0+\hat{\bm{\mu}}_1}{2}\right)\asymp-\frac{1}{2}\bm{\mu}^T\bm{\mu}+\frac{1}{2}\left(\frac{1}{n_0}-\frac{1}{n_1}\right)\text{tr}\left\{\bm{\Sigma}\right\}
\end{equation}

\begin{equation}
    \hat{\bm{\mu}}^T\hat{\bm{\Sigma}}^{-1}\left(\bm{\mu}_0-\frac{\hat{\bm{\mu}}_0+\hat{\bm{\mu}}_1}{2}\right)\asymp-\frac{\tau}{2}\left[\bm{\mu}^T\bm{\Sigma}^{-1}\bm{\mu}-\frac{p}{n_0}+\frac{p}{n_1}\right]
\end{equation}
We now derive the second convergence statement in detail. It is mostly representative of the rest of the derivations. The term $\hat{\bm{\mu}}^T\hat{\bm{\Sigma}}^{-1}\hat{\bm{\mu}}$ can be expressed as
\begin{align}
    \hat{\bm{\mu}}^T\hat{\bm{\Sigma}}^{-1}\hat{\bm{\mu}}&=\left({\bm{\mu}}+\frac{\bm{\Sigma}^{1/2}\textbf{Z}_1\textbf{1}}{n_1}-\frac{\bm{\Sigma}^{1/2}\textbf{Z}_0\textbf{1}}{n_0}\right)^T\hat{\bm{\Sigma}}^{-1}\left({\bm{\mu}}+\frac{\bm{\Sigma}^{1/2}\textbf{Z}_1\textbf{1}}{n_1}-\frac{\bm{\Sigma}^{1/2}\textbf{Z}_1\textbf{0}}{n_0}\right)
\end{align}
where $\textbf{Z}_i\in\mathbb{R}^{p\times n_i}, \ i=0,1$ has i.i.d. $\mathcal{N}\left(0,1\right)$ entries. Taking the expectation over $\textbf{Z}_i\textbf{1},\ i=0,1$, while making use of the fact that $\hat{\bm{\Sigma}}$ is independent of $\hat{\bm{\mu}}$, and that $\frac{\textbf{Z}_i\textbf{1}}{n_i}\sim\mathcal{N}\left(\textbf{0}_p,\frac{1}{n_i}\textbf{I}_p\right), \ i=0,1$, we have the following intermediate convergence result
\begin{align}
     \hat{\bm{\mu}}^T\hat{\bm{\Sigma}}^{-1}\hat{\bm{\mu}}&\asymp \bm{\mu}^T\hat{\bm{\Sigma}}^{-1}\bm{\mu}+\left(\frac{1}{n_0}+\frac{1}{n_1}\right)\text{tr}\left\{\bm{\Sigma}\hat{\bm{\Sigma}}^{-1}\right\}
\end{align}
 We have
\begin{align}
    \bm{\mu}^T\hat{\bm{\Sigma}}^{-1}\bm{\mu}+\left(\frac{1}{n_0}+\frac{1}{n_1}\right)\text{tr}\left\{\bm{\Sigma}\hat{\bm{\Sigma}}^{-1}\right\}&\asymp\bm{\mu}^T\left(\frac{1}{n-2}\bm{\Sigma}^{1/2}{\bar{\textbf{Z}}}{\bar{\textbf{Z}}}^T\bm{\Sigma}^{1/2}\right)^{-1}\bm{\mu}\\
    &\hspace{40px}+\left(\frac{1}{n_0}+\frac{1}{n_1}\right)\text{tr}\left\{\bm{\Sigma}\left(\frac{1}{n-2}\bm{\Sigma}^{1/2}{\bar{\textbf{Z}}}{\bar{\textbf{Z}}}^T\bm{\Sigma}^{1/2}\right)^{-1}\right\}\\
    &=\lim\limits_{\gamma\rightarrow 0}\left[\bm{\mu}^T\textbf{V}\textbf{Q}\textbf{V}^T\bm{\mu}+\left(\frac{1}{n_0}+\frac{1}{n_1}\right)\text{tr}\left\{\textbf{D}_{\bm{\Sigma}}\textbf{Q}\right\}\right]
\end{align}
where $\textbf{Q}=\left(\frac{1}{n-2}\textbf{D}_{\bm{\Sigma}}^{1/2}{\textbf{W}}{\textbf{W}}^T\textbf{D}_{\bm{\Sigma}}^{1/2}+\gamma\textbf{I}_p\right)^{-1}$ and $\textbf{W}=\textbf{V}^T\bar{\textbf{Z}}\in\mathbb{R}^{p\times n-2}$ also has i.i.d entries distributed as $\mathcal{N}({0},1)$ due to invariance of the Gaussian distribution to orthogonal transformations. Using the results in \cite{hachem2013bilinear}, we have 
\begin{align}
  \bm{\mu}^T\textbf{V}\textbf{Q}\textbf{V}^T\bm{\mu}+\left(\frac{1}{n_0}+\frac{1}{n_1}\right)\text{tr}\left\{\textbf{D}_{\bm{\Sigma}}\textbf{Q}\right\}\asymp \bm{\mu}^T\textbf{V}\textbf{T}\textbf{V}^T\bm{\mu}+\left(\frac{1}{n_0}+\frac{1}{n_1}\right)\text{tr}\left\{\textbf{D}_{\bm{\Sigma}}\textbf{T}\right\}
\end{align}
where
\begin{equation}
    \textbf{T}=-\frac{1}{\gamma}\left(\textbf{I}_p+\tilde{\delta}\textbf{D}_{\bm{\Sigma}}\right)^{-1}
\end{equation}
and
\begin{align}
    \delta&=\frac{1}{n}\textbf{tr}\left\{\textbf{D}_{\bm{\Sigma}}\left(-\gamma\left(\textbf{I}_p+\tilde{\delta}\textbf{D}_{\bm{\Sigma}}\right)\right)^{-1}\right\}\\
    \tilde{\delta}&=-\frac{1}{\gamma(1+\delta)}
\end{align}
The desired DE is
\begin{equation}
   \lim\limits_{n,p\rightarrow\infty} \lim\limits_{\gamma\rightarrow 0}\left[\bm{\mu}^T\textbf{V}\textbf{Q}\textbf{V}^T\bm{\mu}+\left(\frac{1}{n_0}+\frac{1}{n_1}\right)\text{tr}\left\{\textbf{D}_{\bm{\Sigma}}\textbf{Q}\right\}\right] \label{hey}
\end{equation}
To be able to apply the above asymptotic result to this expression, we first need to justify the interchange of the limits in \eqref{hey}. This can be done using the Moore-Osgood theorem in a similar way to that shown in Section \ref{moore}.
\begin{align}
    \lim\limits_{\gamma\rightarrow 0}\left[\bm{\mu}^T\textbf{V}\textbf{Q}\textbf{V}^T\bm{\mu}+\left(\frac{1}{n_0}+\frac{1}{n_1}\right)\text{tr}\left\{\textbf{D}_{\bm{\Sigma}}\textbf{Q}\right\}\right]&\asymp \lim\limits_{n,p\rightarrow\infty}\lim\limits_{\gamma\rightarrow 0}\left[\bm{\mu}^T\textbf{V}\textbf{Q}\textbf{V}^T\bm{\mu}+\left(\frac{1}{n_0}+\frac{1}{n_1}\right)\text{tr}\left\{\textbf{D}_{\bm{\Sigma}}\textbf{Q}\right\}\right] \\
    &=\lim\limits_{\gamma\rightarrow 0}\lim\limits_{n,p\rightarrow\infty}\left[\bm{\mu}^T\textbf{V}\textbf{Q}\textbf{V}^T\bm{\mu}+\left(\frac{1}{n_0}+\frac{1}{n_1}\right)\text{tr}\left\{\textbf{D}_{\bm{\Sigma}}\textbf{Q}\right\}\right]\\
    &=\lim\limits_{\gamma\rightarrow 0}\left[\bm{\mu}^T\textbf{V}\textbf{T}\textbf{V}^T\bm{\mu}+\left(\frac{1}{n_0}+\frac{1}{n_1}\right)\text{tr}\left\{\textbf{D}_{\bm{\Sigma}}\textbf{T}\right\}\right]
\end{align}
By making appropriate substitutions in $\textbf{T}$ and taking the limit, it can be shown that
\begin{equation}
    \lim\limits_{\gamma\rightarrow 0}\textbf{T}=\tau\textbf{D}_{\bm{\Sigma}}^{-1}
\end{equation}
under growth condition (d). So overall we obtain
\begin{equation}
    \hat{\bm{\mu}}^T\hat{\bm{\Sigma}}^{-1}\hat{\bm{\mu}}\asymp\tau\left[\bm{\mu}^T\bm{\Sigma}^{-1}\bm{\mu}+\frac{p}{n_0}+\frac{p}{n_1}\right]
\end{equation}
\subsubsection{Derivation of $\bar{m}_1$}
The problem of deriving this deterministic equivalent can be reduced to deriving the following additional convergence statements

\begin{equation}
    \hat{\bm{\mu}}^T\left(\bm{\mu}_1-\frac{\hat{\bm{\mu}}_0+\hat{\bm{\mu}}_1}{2}\right)\asymp\frac{1}{2}\bm{\mu}^T\bm{\mu}+\frac{1}{2}\left(\frac{1}{n_0}-\frac{1}{n_1}\right)\text{tr}\left\{\bm{\Sigma}\right\}
\end{equation}
\begin{equation}
  \hat{\bm{\mu}}^T\hat{\bm{\Sigma}}^{-1}\left(\bm{\mu}_1-\frac{\hat{\bm{\mu}}_0+\hat{\bm{\mu}}_1}{2}\right)\asymp\frac{\tau}{2}\left[\bm{\mu}^T\bm{\Sigma}^{-1}\bm{\mu}+\frac{p}{n_0}-\frac{p}{n_1}\right]
\end{equation}
The proofs are similar to those in Section \ref{barm0}

\subsubsection{Derivation of $\bar{\sigma}_0^2=\bar{\sigma}_1^2$}
The discriminant variance can be expressed as
\begin{align}
    \sigma_0^2=\sigma_1^2&=\left(\rho\hat{\bm{\mu}}^T+\alpha\hat{\bm{\mu}}^T\hat{\bm{\Sigma}}^{-1}\textbf{P}_{\hat{\bm{\mu}}}\right)\bm{\Sigma}\left(\rho\hat{\bm{\mu}}^T+\alpha\hat{\bm{\mu}}^T\hat{\bm{\Sigma}}^{-1}\textbf{P}_{\hat{\bm{\mu}}}\right)^T\\
    &=(1-\alpha)^2\rho^2\hat{\bm{\mu}}^T{\bm{\Sigma}}\hat{\bm{\mu}}+\alpha^2\hat{\bm{\mu}}^T\hat{\bm{\Sigma}}^{-1}\bm{\Sigma}\hat{\bm{\Sigma}}^{-1}\hat{\bm{\mu}}+2\alpha(1-\alpha)\rho\hat{\bm{\mu}}^T\bm{\Sigma}\hat{\bm{\Sigma}}^{-1}\hat{\bm{\mu}}
\end{align}
The problem of deriving this deterministic equivalent can be reduced to deriving the following additional convergence statements

\begin{equation}
    \hat{\bm{\mu}}^T{\bm{\Sigma}}\hat{\bm{\mu}}\asymp\bm{\mu}^T\bm{\Sigma}\bm{\mu}+\left(\frac{1}{n_0}+\frac{1}{n_1}\right)\text{tr}\left\{\bm{\Sigma}^2\right\}
\end{equation}

\begin{equation}
\hat{\bm{\mu}}^T{\bm{\Sigma}}{\hat{\bm{\Sigma}}}^{-1}\hat{\bm{\mu}}\asymp\tau\left[\bm{\mu}^T\bm{\mu}+\left(\frac{1}{n_0}+\frac{1}{n_1}\right)\text{tr}\left\{\bm{\Sigma}\right\}\right]
\end{equation}
\begin{equation}
  \hat{\bm{\mu}}^T{\hat{\bm{\Sigma}}}^{-1}{\bm{\Sigma}}{\hat{\bm{\Sigma}}}^{-1}\hat{\bm{\mu}}\asymp\tau^3\left[\bm{\mu}^T\bm{\Sigma}^{-1}\bm{\mu}+\frac{p}{n_0}+\frac{p}{n_1}\right]
\end{equation}
The last convergence claim involves a double resolvent and therefore we include its derivation here. Using the same technique as before to remove the randomness coming from the sample means, we can show that
\begin{align}
    \hat{\bm{\mu}}^T{\hat{\bm{\Sigma}}}^{-1}{\bm{\Sigma}}{\hat{\bm{\Sigma}}}^{-1}\hat{\bm{\mu}}&\asymp {\bm{\mu}}^T{\hat{\bm{\Sigma}}}^{-1}{\bm{\Sigma}}{\hat{\bm{\Sigma}}}^{-1}{\bm{\mu}}+\left(\frac{1}{n_0}+\frac{1}{n_1}\right)\text{tr}\left\{\bm{\Sigma}\hat{\bm{\Sigma}}^{-1}\bm{\Sigma}\hat{\bm{\Sigma}}^{-1}\right\}\\
    &\asymp\lim\limits_{\gamma\rightarrow 0}\left[{\bm{\mu}}^T\textbf{V}\textbf{Q}\textbf{D}_{\bm{\Sigma}}\textbf{Q}\textbf{V}^T{\bm{\mu}}+\left(\frac{1}{n_0}+\frac{1}{n_1}\right)\text{tr}\left\{\textbf{D}_{\bm{\Sigma}}\textbf{Q}\textbf{D}_{\bm{\Sigma}}\textbf{Q}\right\}\right]
\end{align}
Using the result in \cite{kammoun2019asymptotic} for double resolvents and by interchanging limits as before, we can show that the double resolvent introduces a correction factor of $\frac{1}{1-\frac{p}{n}}$ (in addition to the $\frac{1}{1-\frac{p}{n}}$ introduced by each of the sample covariance matrices) and thus we have 
\begin{equation}
  \hat{\bm{\mu}}^T{\hat{\bm{\Sigma}}}^{-1}{\bm{\Sigma}}{\hat{\bm{\Sigma}}}^{-1}\hat{\bm{\mu}}\asymp\tau^3\left[\bm{\mu}^T\bm{\Sigma}^{-1}\bm{\mu}+\frac{p}{n_0}+\frac{p}{n_1}\right]
\end{equation}

\section{Derivation of the G-estimator of the Probability of Misclassification}\label{GE}
Deriving the G-estimators $\hat{m}_0$, $\hat{m}_1$, $\hat{\sigma}_0^2$, and $\hat{\sigma}_1^2$ reduces to deriving the G-estimators of the constituent quadratic forms that are functions of true statistics. This is the approach taken in what follows.
\subsection{Distinct Covariances}\label{GE_distinct}
\subsubsection{Derivation of $\hat{m}_0$}
Deriving the G-estimator for $m_0$ decomposes into deriving G-estimators of the following terms
\begin{equation}
    \hat{\bm{\mu}}^T\left(\bm{\mu}_0-\frac{\hat{\bm{\mu}}_0+\hat{\bm{\mu}}_1}{2}\right)
\end{equation}

\begin{equation}
    \hat{\bm{\mu}}^T\hat{\bm{\Sigma}}^{-1}\left(\bm{\mu}_0-\frac{\hat{\bm{\mu}}_0+\hat{\bm{\mu}}_1}{2}\right)
\end{equation}

Comparing the DE of the plugin estimator $ \hat{\bm{\mu}}^T\left(\hat{\bm{\mu}}_0-\frac{\hat{\bm{\mu}}_0+\hat{\bm{\mu}}_1}{2}\right)$ to that of $ \hat{\bm{\mu}}^T\left(\bm{\mu}_0-\frac{\hat{\bm{\mu}}_0+\hat{\bm{\mu}}_1}{2}\right)$, we see that we need to add a correction of $\frac{1}{n_0}\text{tr}\left\{\bm{\Sigma}_0\right\}$ to the plugin estimator. It is easy to show that 
\begin{equation}
    \frac{1}{n_0}\text{tr}\left\{\hat{\bm{\Sigma}}_0\right\}\asymp\frac{1}{n_0}\text{tr}\left\{{\bm{\Sigma}}_0\right\}
\end{equation}
from which it follows that
\begin{align} 
      \hat{\bm{\mu}}^T\left(\hat{\bm{\mu}}_0-\frac{\hat{\bm{\mu}}_0+\hat{\bm{\mu}}_1}{2}\right)+\frac{1}{n_0}\text{tr}\left\{\hat{\bm{\Sigma}}_0\right\}\asymp \hat{\bm{\mu}}^T\left(\bm{\mu}_0-\frac{\hat{\bm{\mu}}_0+\hat{\bm{\mu}}_1}{2}\right)
\end{align}

By comparing the DE of the plugin estimator $ \hat{\bm{\mu}}^T\hat{\bm{\Sigma}}^{-1}\left(\hat{\bm{\mu}}_0-\frac{\hat{\bm{\mu}}_0+\hat{\bm{\mu}}_1}{2}\right)$ to that of $ \hat{\bm{\mu}}^T\hat{\bm{\Sigma}}^{-1}\left(\bm{\mu}_0-\frac{\hat{\bm{\mu}}_0+\hat{\bm{\mu}}_1}{2}\right)$, we observe that we must add a correction of $\frac{1}{n_0}\text{tr}\left\{\bm{\Sigma}_0\hat{\bm{\Sigma}}^{-1}\right\}$ to the plugin estimator. A G-estimator for $\frac{1}{n_0}\text{tr}\left\{\bm{\Sigma}_0\hat{\bm{\Sigma}}^{-1}\right\}$ is derived as follows. Expressing $\hat{\bm{\Sigma}}_0$ and $\hat{\bm{\Sigma}}_1$ as 
\begin{align}
   \hat{\bm{\Sigma}}_0&=\frac{1}{n-2}\sum_{i=1}^{n-2}\tilde{\textbf{y}}_{i0}\tilde{\textbf{y}}_{i0}^T
\end{align}
where $\tilde{\textbf{y}}_{i0}\sim\mathcal{N}\left(\textbf{0},\bm{\Sigma}_0\right), \ i=1,2,\ldots,n-2$, and
\begin{align}
   \hat{\bm{\Sigma}}_1&=\frac{1}{n-2}\sum_{i=1}^{n-2}\tilde{\textbf{y}}_{i1}\tilde{\textbf{y}}_{i1}^T
\end{align}
where $\tilde{\textbf{y}}_{i1}\sim\mathcal{N}\left(\textbf{0},\bm{\Sigma}_1\right), \ i=1,2,\ldots,n-2$, then

\begin{align}
    \frac{1}{n_0}\text{tr}\left\{\hat{\bm{\Sigma}}_0\hat{\bm{\Sigma}}^{-1}\right\}&=\frac{1}{n-2}\sum_{i=1}^{n-2}\frac{1}{n_0}\text{tr}\left\{\tilde{\textbf{y}}_{0i}^T\left(\frac{1}{n-2}\sum_j\tilde{\textbf{y}}_{0j}\tilde{\textbf{y}}_{0j}^T+\frac{1}{n-2}\sum_k\tilde{\textbf{y}}_{1k}\tilde{\textbf{y}}_{1k}^T\right)^{-1}\tilde{\textbf{y}}_{0i}\right\}\\
    &=\frac{1}{n-2}\sum_{i=1}^{n-2}\frac{\frac{1}{n_0}\tilde{\textbf{y}}_{0i}\textbf{Q}_i\tilde{\textbf{y}}_{0i}}{1+\frac{1}{n-2}\tilde{\textbf{y}}_{0i}^T\textbf{Q}_i\tilde{\textbf{y}}_{0i}}\\
    &\asymp \frac{1}{n-2}\sum_{i=1}^{n-2}\frac{\frac{1}{n_0}\text{tr}\left\{\textbf{Q}_i\bm{\Sigma}_0\right\}}{1+\frac{1}{n-2}\text{tr}\left\{\textbf{Q}_i\bm{\Sigma}_0\right\}}\\
    &\asymp \frac{\frac{1}{n_0}\text{tr}\left\{\bm{\Sigma}_0\hat{\bm{\Sigma}}^{-1}\right\}}{1+\frac{1}{n-2}\text{tr}\left\{\bm{\Sigma}_0\hat{\bm{\Sigma}}^{-1}\right\}}
\end{align}
where $\textbf{Q}_i=\left(\frac{1}{n-2}\sum_{j\ne i}\tilde{\textbf{y}}_{0j}\tilde{\textbf{y}}_{0j}^T+\frac{1}{n-2}\sum_k\tilde{\textbf{y}}_{1k}\tilde{\textbf{y}}_{1k}^T\right)^{-1}$.
Rearranging, we have
\begin{align}
   \frac{n-2}{n_0}\lambda_0 \asymp\frac{1}{n_0}\text{tr}\left\{\bm{\Sigma}_0\hat{\bm{\Sigma}}^{-1}\right\}
\end{align}
 and so overall,
 \begin{align}
     \hat{\bm{\mu}}^T\hat{\bm{\Sigma}}^{-1}\left(\hat{\bm{\mu}}_0-\frac{\hat{\bm{\mu}}_0+\hat{\bm{\mu}}_1}{2}\right)+\frac{n-2}{n_0}\lambda_0\asymp \hat{\bm{\mu}}^T\left(\bm{\mu}_0-\frac{\hat{\bm{\mu}}_0+\hat{\bm{\mu}}_1}{2}\right)
 \end{align}
\subsubsection{Derivation of $\hat{m}_1$}
Using the same approach as is used for deriving $\hat{m}_0$, it can be shown that
\begin{equation}
     \hat{\bm{\mu}}^T\left(\hat{\bm{\mu}}_1-\frac{\hat{\bm{\mu}}_0+\hat{\bm{\mu}}_1}{2}\right)-\frac{1}{n_1}\text{tr}\left\{\hat{\bm{\Sigma}}_1\right\}\asymp\hat{\bm{\mu}}^T\left(\bm{\mu}_1-\frac{\hat{\bm{\mu}}_0+\hat{\bm{\mu}}_1}{2}\right)
\end{equation}

\begin{equation}
     \hat{\bm{\mu}}^T\hat{\bm{\Sigma}}^{-1}\left(\hat{\bm{\mu}}_1-\frac{\hat{\bm{\mu}}_0+\hat{\bm{\mu}}_1}{2}\right)-\frac{n-2}{n_1}\lambda_1\asymp\hat{\bm{\mu}}^T\hat{\bm{\Sigma}}^{-1}\left(\bm{\mu}_1-\frac{\hat{\bm{\mu}}_0+\hat{\bm{\mu}}_1}{2}\right)
\end{equation}

\subsubsection{Derivation of $\hat{\sigma}_0^2$}
Deriving the G-estimator for $\sigma_0^2$ decomposes into deriving G-estimators of $\hat{\bm{\mu}}^T\bm{\Sigma}_0\hat{\bm{\mu}}$, $    \hat{\bm{\mu}}^T\bm{\Sigma}_0\hat{\bm{\Sigma}}^{-1}\hat{\bm{\mu}}$, and $ \hat{\bm{\mu}}^T\hat{\bm{\Sigma}}^{-1}\bm{\Sigma}_0\hat{\bm{\Sigma}}^{-1}\hat{\bm{\mu}}$. We can easily show 
\begin{align}
    \hat{\bm{\mu}}^T\hat{\bm{\Sigma}}_0\hat{\bm{\mu}}\asymp \hat{\bm{\mu}}^T\bm{\Sigma}_0\hat{\bm{\mu}}
\end{align}
which takes care of the first term. We will now show that
\begin{align}
     \left(1+\lambda_0\right)\hat{\bm{\mu}}^T\hat{\bm{\Sigma}}_0\hat{\bm{\Sigma}}^{-1}\hat{\bm{\mu}} \asymp\hat{\bm{\mu}}^T\bm{\Sigma}_0\hat{\bm{\Sigma}}^{-1}\hat{\bm{\mu}}
\end{align}
Firstly,
\begin{align}
   \hat{\bm{\mu}}^T\hat{\bm{\Sigma}}_0\hat{\bm{\Sigma}}^{-1}\hat{\bm{\mu}}&=\frac{1}{n-2}\sum_{i=1}^{n-2}\hat{\bm{\mu}}^T\tilde{\textbf{y}}_{i0}\tilde{\textbf{y}}_{i0}^T\hat{\bm{\Sigma}}^{-1}\hat{\bm{\mu}} \\
   &=\frac{1}{n-2}\sum_{i=1}^{n-2}\frac{\hat{\bm{\mu}}^T\tilde{\textbf{y}}_{i0}\tilde{\textbf{y}}_{i0}^T\left(\frac{1}{n-2}\sum_{j\neq i}\tilde{\textbf{y}}_{j0}\tilde{\textbf{y}}_{j0}^T+\frac{1}{n-2}\sum_{k=1}^{n-2}\tilde{\textbf{y}}_{k1}\tilde{\textbf{y}}_{k1}^T\right)^{-1}\hat{\bm{\mu}}}{1+\frac{1}{n-2}\tilde{\textbf{y}}_{i0}^T\left(\frac{1}{n-2}\sum_{j\neq i}\tilde{\textbf{y}}_{j0}\tilde{\textbf{y}}_{j0}^T+\frac{1}{n-2}\sum_{k=1}^{n-2}\tilde{\textbf{y}}_{k1}\tilde{\textbf{y}}_{k1}^T\right)^{-1}\tilde{\textbf{y}}_{i0}}\\
   &\asymp \frac{\hat{\bm{\mu}}^T\bm{\Sigma}_0\hat{\bm{\Sigma}}^{-1}\hat{\bm{\mu}}}{1+\frac{1}{n-2}\text{tr}\left\{\bm{\Sigma}_0\hat{\bm{\Sigma}}^{-1}\right\}}
\end{align}
which means that
\begin{align}
    \left(1+\frac{1}{n-2}\text{tr}\left\{\bm{\Sigma}_0\hat{\bm{\Sigma}}^{-1}\right\}\right) \hat{\bm{\mu}}^T\hat{\bm{\Sigma}}_0\hat{\bm{\Sigma}}^{-1}\hat{\bm{\mu}}\asymp  \hat{\bm{\mu}}^T{\bm{\Sigma}}_0\hat{\bm{\Sigma}}^{-1}\hat{\bm{\mu}}
\end{align}
The final expression is obtained by substituting the G-estimator of $\frac{1}{n-2}\text{tr}\left\{\bm{\Sigma}_0\hat{\bm{\Sigma}}^{-1}\right\}$ derived previously.

In a similar way, it can be shown that
\begin{align}
     \left(1+\lambda_0\right)^2\hat{\bm{\mu}}^T\hat{\bm{\Sigma}}^{-1}\hat{\bm{\Sigma}}_0\hat{\bm{\Sigma}}^{-1}\hat{\bm{\mu}} \asymp\hat{\bm{\mu}}^T\hat{\bm{\Sigma}}^{-1}\bm{\Sigma}_0\hat{\bm{\Sigma}}^{-1}\hat{\bm{\mu}}
\end{align}

\subsubsection{Derivation of $\hat{\sigma}_1^2$}
In a similar manner, we derive the following convergence relations for the constituent terms of $\sigma_1^2$ 
\begin{align}
    \hat{\bm{\mu}}^T\hat{\bm{\Sigma}}_1\hat{\bm{\mu}}\asymp\hat{\bm{\mu}}^T\bm{\Sigma}_1\hat{\bm{\mu}}
\end{align}

\begin{align}
     \left(1+\lambda_1\right)\hat{\bm{\mu}}^T\hat{\bm{\Sigma}}_1\hat{\bm{\Sigma}}^{-1}\hat{\bm{\mu}}\asymp\hat{\bm{\mu}}^T\bm{\Sigma}_1\hat{\bm{\Sigma}}^{-1}\hat{\bm{\mu}}
\end{align}

\begin{align}
      \left(1+\lambda_1\right)^2\hat{\bm{\mu}}^T\hat{\bm{\Sigma}}^{-1}\hat{\bm{\Sigma}}_1\hat{\bm{\Sigma}}^{-1}\hat{\bm{\mu}}\asymp\hat{\bm{\mu}}^T\hat{\bm{\Sigma}}^{-1}\bm{\Sigma}_1\hat{\bm{\Sigma}}^{-1}\hat{\bm{\mu}}
\end{align}

\subsection{Common Covariances}\label{GE_common}
\subsubsection{Derivation of $\hat{m}_0$}
Expressing $m_0$ as
\begin{align}
    m_0
     &=\left(\rho\hat{\bm{\mu}}^T+\alpha\hat{\bm{\mu}}^T\hat{\bm{\Sigma}}^{-1}\textbf{P}_{\hat{\bm{\mu}}}\right)\left(\hat{\bm{\mu}}_0-\frac{\hat{\bm{\mu}}_0+\hat{\bm{\mu}}_1}{2}+{\bm{\mu}}_0-\hat{\bm{\mu}}_0\right)\\
     &=(1-\alpha)\rho\hat{\bm{\mu}}^T\left(\hat{\bm{\mu}}_0-\frac{\hat{\bm{\mu}}_0+\hat{\bm{\mu}}_1}{2}\right)+\alpha\hat{\bm{\mu}}^T\hat{\bm{\Sigma}}^{-1}\left(\hat{\bm{\mu}}_0-\frac{\hat{\bm{\mu}}_0+\hat{\bm{\mu}}_1}{2}\right)\\
     &\hspace{15px}+(1-\alpha)\rho\hat{\bm{\mu}}^T\left({\bm{\mu}}_0-\hat{\bm{\mu}}_0\right)+\alpha\hat{\bm{\mu}}^T\hat{\bm{\Sigma}}^{-1}\left({\bm{\mu}}_0-\hat{\bm{\mu}}_0\right). \label{m0_form}
\end{align}
we see that G-estimators for $\hat{\bm{\mu}}^T\left({\bm{\mu}}_0-\hat{\bm{\mu}}_0\right)$ and $\hat{\bm{\mu}}^T\hat{\bm{\Sigma}}^{-1}\left({\bm{\mu}}_0-\hat{\bm{\mu}}_0\right)$ are needed.
By substituting 
 \begin{equation}
    \hat{\bm{\mu}}_0={\bm{\mu}}_0+\frac{\bm{\Sigma}^{1/2}\textbf{Z}_0\textbf{1}}{n_0}
\end{equation}
\begin{equation}
    \hat{\bm{\mu}}_1={\bm{\mu}}_1+\frac{\bm{\Sigma}^{1/2}\textbf{Z}_1\textbf{1}}{n_1}
\end{equation}
and taking the expectation over $\textbf{Z}_0\textbf{1}$ and $\textbf{Z}_1\textbf{1}$ in $\hat{\bm{\mu}}^T\left({\bm{\mu}}_0-\hat{\bm{\mu}}_0\right)$, we obtain 
\begin{align}
   \hat{\bm{\mu}}^T\left({\bm{\mu}}_0-\hat{\bm{\mu}}_0\right)\asymp \frac{1}{n_0}\text{tr}\left\{\bm{\Sigma}\right\}
\end{align}
We can easily show that
\begin{equation}
    \frac{1}{n_0}\text{tr}\left\{\hat{\bm{\Sigma}}\right\}\asymp \frac{1}{n_0}\text{tr}\left\{\bm{\Sigma}\right\}
\end{equation}
by substituting $\frac{1}{n-2}\bm{\Sigma}^{1/2}{\bar{\textbf{Z}}}{\bar{\textbf{Z}}}^T\bm{\Sigma}^{1/2}$ for $\hat{\bm{\Sigma}}$ and taking the expectation. Thus, we have
\begin{equation}
     \frac{1}{n_0}\text{tr}\left\{\hat{\bm{\Sigma}}\right\}\asymp  \hat{\bm{\mu}}^T\left({\bm{\mu}}_0-\hat{\bm{\mu}}_0\right)
\end{equation}

Through a similar derivation, we obtain
\begin{align}
    \hat{\bm{\mu}}^T\hat{\bm{\Sigma}}^{-1}\left({\bm{\mu}}_0-\hat{\bm{\mu}}_0\right)\asymp \frac{1}{n_0}\text{tr}\left\{\bm{\Sigma}\hat{\bm{\Sigma}}^{-1}\right\}
\end{align}
To find the G-estimator of this quantity, replace $\bm{\Sigma}$ with its estimate and then express this as a function of the original quantity as follows. First express $\hat{\bm{\Sigma}}$ as
\begin{align}
    \hat{\bm{\Sigma}}
    &=\frac{1}{n-2}\sum_{i=1}^{n-2}\tilde{\textbf{y}}_i\tilde{\textbf{y}}_i^T
\end{align}
where  $\tilde{\textbf{y}}_i\sim \mathcal{N}\left(\textbf{0},\bm{\Sigma}\right), \ i=1,2,\ldots, n-2$.
So we have
\begin{align}
    \frac{1}{n_0}\text{tr}\left\{\bm{\Sigma}\hat{\bm{\Sigma}}^{-1}\right\}&=\frac{1}{n-2}\sum_{i=1}^{n-2}\frac{1}{n_0}\text{tr}\left\{\tilde{\textbf{y}}_i\tilde{\textbf{y}}_i^T\left(\frac{1}{n-2}\sum_{j=1}^{n-2}\tilde{\textbf{y}}_j\tilde{\textbf{y}}_j^T\right)^{-1}\right\}\\
    &=\frac{1}{n-2}\sum_{i=1}^{n-2}\frac{1}{n_0}\text{tr}\left\{\tilde{\textbf{y}}_i^T\left(\frac{1}{n-2}\sum_{j=1}^{n-2}\tilde{\textbf{y}}_j\tilde{\textbf{y}}_j^T\right)^{-1}\tilde{\textbf{y}}_i\right\}\\
    &=\frac{1}{n-2}\sum_{i=1}^{n-2}\frac{1}{n_0}\text{tr}\left\{\tilde{\textbf{y}}_i^T\left(\frac{1}{n-2}\sum_{j\ne i}\tilde{\textbf{y}}_j\tilde{\textbf{y}}_j^T+\tilde{\textbf{y}}_i\tilde{\textbf{y}}_i^T\right)^{-1}\tilde{\textbf{y}}_i\right\}\\
    &=\frac{\frac{1}{n_0}\tilde{\textbf{y}}_i^T\textbf{Q}_i\tilde{\textbf{y}}_i}{1+\frac{1}{n-2}\tilde{\textbf{y}}_i^T\textbf{Q}_i\tilde{\textbf{y}}_i}
\end{align}
where $\textbf{Q}_i=\left(\frac{1}{n-2}\sum_{j\ne i}\tilde{\textbf{y}}_j\tilde{\textbf{y}}_j^T\right)^{-1}$ and the last line follows from applying the matrix inversion lemma in \cite{muller2016random}.
It can be shown that
\begin{align}
    \frac{\frac{1}{n_0}\tilde{\textbf{y}}_i^T\textbf{Q}_i\tilde{\textbf{y}}_i}{1+\frac{1}{n-2}\tilde{\textbf{y}}_i^T\textbf{Q}_i\tilde{\textbf{y}}_i}\asymp
    \frac{\frac{1}{n_0}\text{tr}\{\bm{\Sigma}\hat{\bm{\Sigma}}^{-1}\}}{1+\frac{1}{n-2}\text{tr}\{\bm{\Sigma}\hat{\bm{\Sigma}}^{-1}\}}
\end{align}
Therefore,
\begin{equation}
    \frac{1}{n_0}\text{tr}\{\hat{\bm{\Sigma}}\hat{\bm{\Sigma}}^{-1}\}=\frac{p}{n_0}\asymp     \frac{\frac{1}{n_0}\text{tr}\{\bm{\Sigma}\hat{\bm{\Sigma}}^{-1}\}}{1+\frac{1}{n-2}\text{tr}\{\bm{\Sigma}\hat{\bm{\Sigma}}^{-1}\}}
\end{equation}
and solving for the original quantity, we have
\begin{equation}
    \frac{\frac{p}{n_0}}{1-\frac{p}{n-2}}\asymp \frac{1}{n_0}\text{tr}\left\{\bm{\Sigma}\hat{\bm{\Sigma}}^{-1}\right\}
\end{equation}
% Overall, the G-estimator of $m_0$ is given by 
% \begin{align}
%   \hat{m}_0&=\left(\frac{\hat{\bm{\mu}}^T\hat{\bm{\Sigma}}^{-1}\hat{\bm{\mu}}}{\hat{\bm{\mu}}^T\hat{\bm{\mu}}}\hat{\bm{\mu}}^T+\alpha\hat{\bm{\mu}}^T\hat{\bm{\Sigma}}^{-1}\left(\textbf{I}-\frac{\hat{\bm{\mu}}\hat{\bm{\mu}}^T}{\hat{\bm{\mu}}^T\hat{\bm{\mu}}}\right)\right)\left(\hat{\bm{\mu}}_0-\frac{\hat{\bm{\mu}}_0+\hat{\bm{\mu}}_1}{2}\right)\\
%     & \hspace{30px}+(1-\alpha)\frac{\hat{\bm{\mu}}^T\hat{\bm{\Sigma}}^{-1}\hat{\bm{\mu}}}{\hat{\bm{\mu}}^T\hat{\bm{\mu}}} \frac{1}{n_0}\text{tr}\left\{\hat{\bm{\Sigma}}\right\}+\alpha\frac{\frac{p}{n_0}}{1-\frac{p}{n-2}}
% \end{align}
\subsubsection{Derivation of $\hat{m}_1$}
The G-estimators for $\hat{\bm{\mu}}^T\left({\bm{\mu}}_1-\hat{\bm{\mu}}_1\right)$ and $\hat{\bm{\mu}}^T\hat{\bm{\Sigma}}^{-1}\left({\bm{\mu}}_1-\hat{\bm{\mu}}_1\right)$ are derived in a similar fashion to the previous section.

% Overall, the G-estimator of $m_1$ is given by 
% \begin{align}
%     \hat{m}_1&=\left(\frac{\hat{\bm{\mu}}^T\hat{\bm{\Sigma}}^{-1}\hat{\bm{\mu}}}{\hat{\bm{\mu}}^T\hat{\bm{\mu}}}\hat{\bm{\mu}}^T+\alpha\hat{\bm{\mu}}^T\hat{\bm{\Sigma}}^{-1}\left(\textbf{I}-\frac{\hat{\bm{\mu}}\hat{\bm{\mu}}^T}{\hat{\bm{\mu}}^T\hat{\bm{\mu}}}\right)\right)\left(\hat{\bm{\mu}}_1-\frac{\hat{\bm{\mu}}_0+\hat{\bm{\mu}}_1}{2}\right)\\
%     &\hspace{30px}-(1-\alpha)\frac{\hat{\bm{\mu}}^T\hat{\bm{\Sigma}}^{-1}\hat{\bm{\mu}}}{\hat{\bm{\mu}}^T\hat{\bm{\mu}}} \frac{1}{n_1}\text{tr}\left\{\hat{\bm{\Sigma}}\right\}-\alpha\frac{\frac{p}{n_1}}{1-\frac{p}{n-2}}
% \end{align}
\subsection{Derivation of $\hat{\sigma}_0^2=\hat{\sigma}_1^2$}
We need G-estimators for the terms $\hat{\bm{\mu}}^T{\bm{\Sigma}}\hat{\bm{\mu}}$, $\hat{\bm{\mu}}^T\hat{\bm{\Sigma}}^{-1}\bm{\Sigma}\hat{\bm{\Sigma}}^{-1}\hat{\bm{\mu}}$, and $\hat{\bm{\mu}}^T\bm{\Sigma}\hat{\bm{\Sigma}}^{-1}\hat{\bm{\mu}}$. It can be easily shown that
\begin{equation}
    \hat{\bm{\mu}}^T{\hat{\bm{\Sigma}}}\hat{\bm{\mu}}\asymp\hat{\bm{\mu}}^T{\bm{\Sigma}}\hat{\bm{\mu}}
\end{equation}

From Appendix \ref{DE_common}, we have
\begin{equation}
    \hat{\bm{\mu}}^T\hat{\bm{\Sigma}}^{-1}\bm{\Sigma}\hat{\bm{\Sigma}}^{-1}\hat{\bm{\mu}}\asymp \tau^3\left[\bm{\mu}^T\bm{\Sigma}^{-1}\bm{\mu}+\frac{p}{n_0}+\frac{p}{n_1}\right]
\end{equation}
If we replace $\bm{\Sigma}$ by its estimator, we have
\begin{align}
   \hat{\bm{\mu}}^T\hat{\bm{\Sigma}}^{-1}\hat{\bm{\Sigma}}\hat{\bm{\Sigma}}^{-1}\hat{\bm{\mu}}&= \hat{\bm{\mu}}^T\hat{\bm{\Sigma}}^{-1}\hat{\bm{\mu}}\\
    &\asymp \tau\left[\bm{\mu}^T\bm{\Sigma}^{-1}\bm{\mu}+\frac{p}{n_0}+\frac{p}{n_1}\right]
\end{align}
therefore,
\begin{equation}
    \tau^2 \hat{\bm{\mu}}^T\hat{\bm{\Sigma}}^{-1}\hat{\bm{\mu}}\asymp  \hat{\bm{\mu}}^T\hat{\bm{\Sigma}}^{-1}\bm{\Sigma}\hat{\bm{\Sigma}}^{-1}\hat{\bm{\mu}}
\end{equation}

From Appendix \ref{DE_common}, we know that 
\begin{equation}
 \hat{\bm{\mu}}^T\bm{\Sigma}\hat{\bm{\Sigma}}^{-1}\hat{\bm{\mu}}\asymp \tau\left[\bm{\mu}^T\bm{\mu}+\left(\frac{1}{n_0}+\frac{1}{n_1}\right)\text{tr}\left\{\bm{\Sigma}\right\}\right]
\end{equation}
If we replace $\bm{\Sigma}$ by its estimator, we have
\begin{align}
    \hat{\bm{\mu}}^T\hat{\bm{\Sigma}}\hat{\bm{\Sigma}}^{-1}\hat{\bm{\mu}}&=\hat{\bm{\mu}}^T\hat{\bm{\mu}}\asymp \bm{\mu}^T\bm{\mu}+\left(\frac{1}{n_0}+\frac{1}{n_1}\right)\text{tr}\left\{\bm{\Sigma}\right\}
\end{align}
where the last line is from Appendix \ref{DE_common}. Therefore,
\begin{equation}
    \tau\hat{\bm{\mu}}^T\hat{\bm{\mu}}\asymp  \hat{\bm{\mu}}^T\bm{\Sigma}\hat{\bm{\Sigma}}^{-1}\hat{\bm{\mu}}
\end{equation}

\bibliographystyle{IEEEtran}
\bibliography{citations}

\end{document}